\documentclass{article}

\usepackage{arxiv}

\usepackage[utf8]{inputenc} 
\usepackage[T1]{fontenc}    
\usepackage{hyperref}       
\usepackage{url}            
\usepackage{booktabs}       
\usepackage{amsfonts}       
\usepackage{nicefrac}       
\usepackage{microtype}      

\usepackage{comment}
\usepackage{amssymb}
\usepackage{amsmath}
\usepackage{amsthm}
\DeclareMathOperator*{\argmax}{\arg\!\max}
\DeclareMathOperator*{\argmin}{\arg\!\min}
\usepackage{graphicx}
\usepackage{mathtools}
\usepackage{dsfont}
\usepackage[algoruled,boxed,linesnumbered]{algorithm2e}
\usepackage{parskip}
\usepackage{textcomp}
\usepackage{subcaption}
\usepackage{physics}
\usepackage[tableposition=top]{caption}
\usepackage[table,dvipsnames]{xcolor}
\usepackage{float}
\usepackage{breakcites}

\hypersetup{bookmarksnumbered}

\definecolor{MyLightGray}{gray}{0.9}
\definecolor{MyDarkGray}{gray}{0.6}

\SetCommentSty{mycommfont}

\date{}

\title{Optimality Inductive Biases and
Agnostic Guidelines for Offline Reinforcement Learning}

\author{ 
Lionel~Blond\'e\thanks{Correspondence to Lionel Blond\'e: \texttt{lionel.blonde@gmail.com}.} \\
University of Geneva, \\
HES-SO, Switzerland \\
\And
Alexandros~Kalousis \\
University of Geneva, \\
HES-SO, Switzerland \\
\And
Stéphane Marchand-Maillet \\
University of Geneva \\
Switzerland
}

\begin{document}

\maketitle

\begin{abstract}

The performance of state-of-the-art offline RL methods varies widely over the
spectrum of dataset qualities, ranging from far-from-optimal random data
to close-to-optimal expert demonstrations.
We re-implement these methods to test their reproducibility,
and show that when a given method outperforms the others
on one end of the spectrum, it never does on the other end.
This prevents us from naming a victor across the board.
We attribute the asymmetry
to the amount of inductive bias injected into the agent to entice it to posit that the behavior underlying
the offline dataset is optimal for the task.
Our investigations confirm that careless
injections of such optimality inductive biases
make dominant agents subpar as soon as the offline policy is sub-optimal.
To bridge this gap, we generalize importance-weighted regression methods that have proved
the most versatile across the spectrum of dataset grades
into a modular framework that allows for the design of methods that align with how much we know
about the dataset.
This modularity
enables qualitatively different injections of optimality inductive biases.
We show that certain orchestrations strike the right balance, improving the return on one end of the spectrum
without harming it on the other end.
While the formulation of guidelines for the design of an offline method
reduces to aligning the amount of optimality bias to inject with what we know about the quality of the data,
the design of an agnostic method for which we need not know the quality of the data beforehand is more nuanced.
Only our framework allowed us to design a method that performed well across the spectrum while
remaining modular if more information about the quality of the data ever becomes available.
\end{abstract}

\section{Introduction}
\label{introoffline}

Reinforcement learning (RL) \cite{Sutton1998-ow}
is the branch of interactive machine learning
that has received the most attention in recent years, due to
its instrumental role in tackling a number of grand AI challenges
(\textit{e.g.} going beyond human performance in
board and video games \cite{Mnih2013-rb, Mnih2015-iy, Silver2017-xo, Silver2016-my, Vinyals2019-vx, OpenAI2019-xy},
hitting a new milestone in AI-operated hand dexterity leading
to the resolution of a Rubik's cube \cite{OpenAI2019-vy}).
The online RL agents learn by acting in the world (either real or simulated),
and update their intrinsic decision-making process by
internalizing the feedback returned by the world upon interaction.
The feedback takes the form of a reward signal,
which scores the agent according to how appropriate its \emph{own} executed actions were for the task at hand
(how rewards are designed or come from is out of the scope of this paper,
\textit{cf.}~\cite{Singh2009-cm}).
Nonetheless, while learning from our mistakes is undeniably valuable, it is of far greater
value to be able to learn from the mistakes of others,
a sub-branch of machine learning sometimes referred to as \emph{counterfactual} learning
\cite{Bottou2013-so}.
In the context of RL, pure
counterfactual learning crystallizes as \emph{offline} RL \cite{Levine2020-hz}
(alternatively, \emph{batch} RL \cite{Lange2012-cc}).
The agent is only allowed to learn from an offline dataset.
This dataset contains logged interactions
that were experienced and collected by another policy as it interacted with the world.
The offline RL agent is not allowed to interact with the world,
and is consequently unable to learn interactively from her own mistakes.

Offline RL shines
\textit{a)} when interaction data (albeit from multiple non-egocentric sources) are abundant and diverse
and
\textit{b)} when simulator-in-the-loop approaches
are impractical, due to deterringly high costs or safety concerns.
In practice, it is incredibly tedious and challenging to design and implement
a data collection strategy able to capture the
diversity of the real world in a dataset.
Besides, crafting a simulator able to model how the world reacts to the agent with high fidelity
is an engineering feat too
(\textit{e.g.} the elaborate system of automatic domain randomization used in \cite{OpenAI2019-vy}
to assist in the resolution of of a Rubik's cube, which required the aid of the purposely-developed
asset randomization engine reported in \cite{Chociej2019-ot}).
Iteratively integrating hitherto-omitted edge cases is a long-term endeavor.
Even when one manages to train agents yielding high return in simulation, they are not guaranteed to
transfer to the real world, and often require additional model surgery
(\textit{cf.}~\textit{``sim-to-real''} research
\cite{Rusu2016-dc},
a sub-domain of transfer RL
\cite{Taylor2009-pc}).
Designing offline RL methods able to squeeze the most juice out of potentially imperfect
offline real-world interaction data is both a crucial and timely problem to solve.
The crying need for such methods in popular consumer-facing
real-world applications (\textit{e.g.} robotics, autonomous driving, healthcare, conversational agents)
explains the resurgence of offline RL \cite{Levine2020-hz}.

Online off-policy actor-critic methods
require the decision-maker to learn an estimate of the state-action value
through bootstrapping with next actions generated by the agent,
and as such implement the \textit{SARSA} update rule
\cite{Rummery1994-qp, Thrun1995-sz, Sutton1996-ky, Van_Seijen2009-yw}.
These methods suffer from a \emph{distributional shift} \cite{Fu2019-kb}
as soon as the experience replay \cite{Lin1993-qe}
mechanism comes into play, since an action generated by the current policy
is used to update the value at an action randomly picked from the replay memory --- effectively
distributed as a mixture of past policy updates.
This covariate shift phenomenon is exacerbated in the \emph{offline} setting,
where the value is \emph{only} subject to updates at the fixed set of actions present on the offline dataset.
Since the distribution over actions underlying the offline dataset neither evolves nor need be tied to the
agent's policy in any way, the distributional shift can grow arbitrarily large. By contrast,
the mixture of past policy updates underlying a replay memory will by design
always track (yet not necessarily closely) the current policy update, effectively limiting the
gap between the action distributions.
Since
in the offline setting
the value is solely updated over a \emph{frozen} offline dataset detached from the policy
followed by the agent, the value is all the more so exposed to being evaluated at \emph{out-of-distribution} actions
than its counterpart in the online (off-policy) setting.
As such, the value can be arbitrarily erroneous at actions that are
too far off the offline distribution underlying the dataset.
This motivates the design of offline RL methods that
deter the agent from straying onto out-of-distribution action
by urging the agent to \textit{``stay close to the offline dataset''} (\textit{cf.}~\textsc{Figure}~\ref{gpidiag}),
either explicitly or implicitly.
We lay these out in \textsc{Section}~\ref{relatedwork}.

\section{Contributions}
\label{contriblist}

\paragraph{Contribution \#1: critical evaluation of the landscape.}

We open-source\footnote{Code made available at the URL:
\texttt{https://github.com/lionelblonde/giwr-pytorch}.}
fair re-implementations of state-of-the-art offline RL methods
under a proposed unified framework.
In \textsc{Section}~\ref{baselines}, we discuss the methods included in this release,
before assessing and comparing
these competing methods empirically in \textsc{Section}~\ref{experimentalassessment}.
This study, along with all the other analyses reported in this work, are conducted under the experimental
setting laid out in \textsc{Appendix}~\ref{experimentalsetting}.

\paragraph{Contribution \#2: optimality inductive biases.}

In \textsc{Section}~\ref{conceptualization},
we explicitly formalize the notion of \emph{optimality inductive bias}.
When injected into the decision-maker, it conceptually quantifies how much the latter
posits that the policy underlying the offline dataset is optimally solving the task at hand.
We investigate empirically to what extent the amount of inductive bias injected in the agent impacts
its performance across the spectrum of considered datasets and environments.
The performance comparison depicted in \textsc{Section}~\ref{experimentalassessment} shows how most baselines
inject far too much optimality inductive bias to achieve reasonable levels of performance when the
dataset is sub-optimal.
Besides, these baselines have proven hard to tune \cite{Wu2019-nl, Le_Paine2020-sb, Monier2020-tq},
and the hyper-parameters displaying the largest swings in performance are the ones
controlling the amount of bias injected.
From \textsc{Section}~\ref{inductivebiases} forward, we use the advantage-weighted regression
template (\textit{cf.}~\textsc{Algorithm}~\ref{algobase})
adopted identically in \cite{Wang2020-sr}
and \cite{Nair2020-gd},
given that it yields the most favorable empirical results in \textsc{Section}~\ref{experimentalassessment}.
In order to verify that its competitive superiority over the spectrum of dataset qualities is due to the
\emph{implicit} nature of how such agents are injected with
optimality inductive bias,
we construct and diagnose an extension of the base advantage-weighted regression method in
\textsc{Section}~\ref{evidence}.
We handcraft this extension by regularizing the optimized objective
such that the extension injects more optimality inductive bias into the agents trained with it
than the base method does.
We push the diagnostic further by evaluating the return of both methods, for a given environment,
\textbf{on a series of mixed datasets each simulating a scenario
in which the expert dataset has been corrupted
with transitions from the random dataset}, to various degrees of severity.
These reveal just how strikingly
sensitive to the means of bias injection offline RL methods are,
especially considering how often real-world data is corrupted.

\paragraph{Contribution \#3: Generalized Policy Iteration (GPI) revisitation for offline RL,
leading to the novel Generalized Importance-Weighted Regression (GIWR) framework.}

\begin{figure}[!h]
  \center\scalebox{0.4}[0.4]{\includegraphics{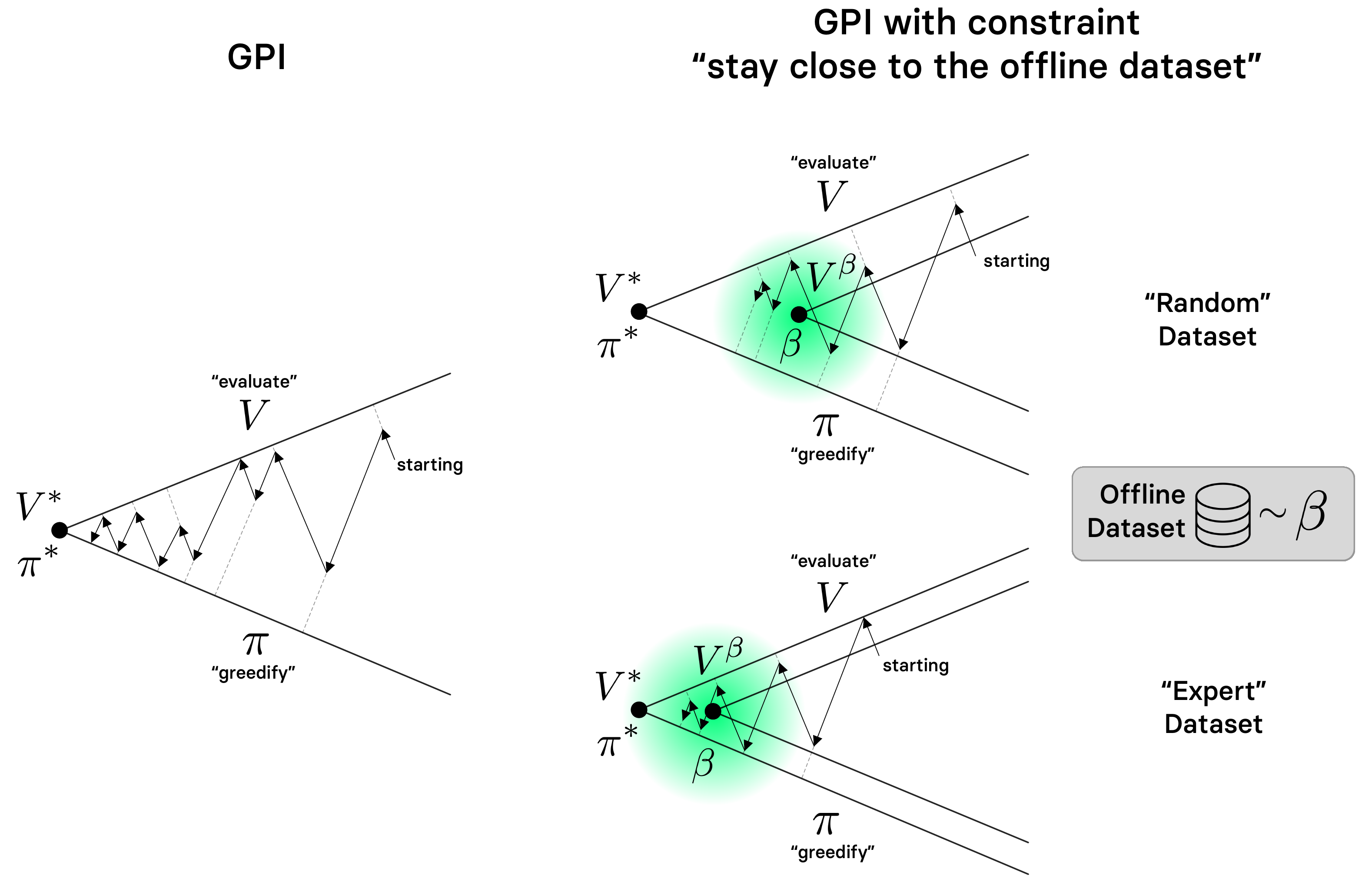}}
  \caption{Sequence diagram representation of generalized policy iteration (GPI):
  as depicted in \cite{Sutton1998-ow} for the online RL scenario on the left-hand side,
  and augmented for the offline RL case on the right-hand side.
  As reminded in \cite{Sutton1998-ow}, the real geometry is considerably more complex that this.
  In contrast with the diagram of \cite{Sutton1998-ow} however,
  the consecutive policy \textit{evaluation} and \textit{improvement}
  sub-goals are not performed to completion at each step.
  The green area depicts the subspace of the policy-value joint parameter space where the constraint
  encoding the desideratum \textit{``stay close to the offline dataset''} is satisfied
  (with the color density depicting the degree of satisfaction).
  The shape of the green area is determined by the metric used to enforce the constraint
  (although chosen here for legibility, we would have disks if we were employing an $\ell_2$-loss over parameters).
  Crucially, this diagram illustrates
  \textit{a)} that the gap between $(\pi^*, V^*)$ and $(\beta, V^\beta)$ is greater when
  the dataset is of poor quality,
  and \textit{b)} how this gap would prevent an agent eager to remain close to the offline dataset (green area)
  from ever reaching optimal performance.}
  \label{gpidiag}
\end{figure}

In \textsc{Sections}~\ref{gpirevisitationintro}, \ref{pe} and \ref{pi},
we propose generalizations of the value and policy objectives involved in the considered base actor-critic
method (\textit{cf.}~\textsc{Algorithm}~\ref{algobase}),
where \textsc{Section}~\ref{gpirevisitationintro} primarily aims at setting the stage
for both of the subsequent remaining sections.
\textsc{Section}~\ref{pe} is dedicated to policy evaluation,
and \textsc{Section}~\ref{pi} is dedicated to policy improvement.
These generalizations involve \emph{proposal} policies,
effectively acting as placeholders for a slew of different
action distributions that we introduce (9 in total) and evaluate,
in both contexts (\emph{evaluation} in \textsc{Section}~\ref{pe}, \emph{improvement} in \textsc{Section}~\ref{pi}).
Each proposal strategy tackles, with its own distinct flavor, the \emph{balancing act} that both the policy
and value entangled in a GPI scheme must address in offline RL:
\textbf{\emph{how to come close to behaving optimally for the given task
while avoiding being hindered by out-of-distribution actions?}}
Under a notion of safety purposely derived from this desideratum, we can equivalently ask:
how to teach offline agents to approach optimality \emph{safely}?
As illustrated in \textsc{Figure}~\ref{gpidiag},
striking the right balance is far from obvious and highly dependent on the quality of the dataset.
Nonetheless, we show that our new, highly modular,
generalized framework
(we call it GIWR, which stands for \textbf{Generalized Importance-Weighted Regression},
\textit{cf.}~\textsc{Section}~\ref{multisteams})
enables us to \textbf{safely inject optimality inductive bias(es)
in the method in such a way that the results are improved when the policy
underlying the dataset is optimal, while not being harmed when it is not}.

\paragraph{N.B.} Our agents are \emph{never} made aware of the quality of the offline dataset they are provided with.

\section{Related Work}
\label{relatedwork}

The subfield of offline RL has been outlined in two surveys, under slightly different scopes.
In \cite{Lange2012-cc},
the authors give an overview of the \textit{batch} setting, and a thorough picture of how it was tackled from
its inception by Gordon in 1995 (\textit{cf.}~\cite{Gordon1995-er})
to the release date of the survey.
Eight years later,
in \cite{Levine2020-hz},
the offline RL landscape is painted by the authors through a more practice-oriented lens.
While \cite{Lange2012-cc}
identified \emph{stability} as the main culprit hindering the wide adoption of offline RL in real-world systems
(defending that a genuine expertise was needed to pull it off),
\cite{Levine2020-hz}
can depict a more optimistic view of the domain since it has since received attention and been leveraged in
various applications (\textit{e.g.} such as dialog systems).
Despite having been addressed numerous times in recent years, and approached via different angles,
the practical stability issues discussed in \cite{Lange2012-cc}
have yet to be solved, and are barely mitigated in the current state-of-the-art methods.
Going back to the inception of batch RL,
the work of \cite{Gordon1995-er}
proposes a model-based method called \textsc{Averagers} that estimates the action-value exactly at
\textit{supports}. It does so by leveraging the exact dynamic programming operation,
requiring the availability of a model of the transition function
to be able to solve the subsequent \emph{planning} task with it.
Besides, the approach of \cite{Gordon1995-er}
identifies as a \textit{``fitted''} one,
since it solves the action-value estimation problem before deriving a control policy from it.
Note, the sequence diagrams depicted in \textsc{Figure}~\ref{gpidiag} do \emph{not} temporally represent
the learning process of agents learned
under said \textit{fitted} paradigm --- otherwise, the dotted arrows would reach
the solid lines every single time.
Albeit raised and discussed, \cite{Gordon1995-er}
did not propose a model-free counterpart of the method,
due to convergence issues for estimating the action-values at the supports
from the action-values estimated at samples from the respective neighborhoods of the supports.
Ormoneit and Sen \cite{Ormoneit2002-wk}
address this issue, though not in the \textsc{Averagers} framework.
Instead of attempting to estimate the values from neighboring samples
at supports, they
estimate the value directly via a \emph{sample-based} approximator (which is kernel-based).
In other words, instead of relying on a transition model and carrying out
the exact dynamic programming operation with it,
their Kernel-based Approximate Dynamic Programming (KADP) method
detaches itself from the need to possess a model of the world:
it approximates the exact dynamic programming operator
with the implicit sample-based transition model observable via the
collected transitions in the batch
(that we have access to, yet ignore the underlying dynamics and behavior models of).
Our work is closer to the Least-Squares Policy Iteration (LSPI) method
from \cite{Lagoudakis2003-gp}
--- whose first release was concurrent with KADP,
since we set out to revisit the generalized policy iteration learning template in the offline regime
(\textit{cf.}~\textsc{Section}~\ref{introoffline} for what motivates our study,
\textit{cf.}~\textsc{Figure}~\ref{gpidiag} for a conceptual depiction of the learning scheme).
As such, we only consider methods that alternate between policy evaluation and
policy improvement steps, in contrast with the distinct separation of these two problems that characterizes the
\textit{fitted} methods \cite{Gordon1995-er, Ernst2005-sf, Riedmiller2005-ok,
Antos2007-bw, Neumann2008-tm, Lange2010-mq, Hafner2011-rv}.
Note, LSPI \cite{Lagoudakis2003-gp}
does not involve an explicit policy, the policy is defined implicitly from the learned value.

When it comes down to the modern state-of-the-art offline RL approaches
which we critically evaluate in this work,
one can divide them up into three distinct categories.
These differ by how they encode the constraint
enticing the agent to stay close to the offline dataset.
They do so
\textit{1)} with \emph{explicit} constraints on the \emph{policy}
(\textit{e.g.}
BRAC \cite{Wu2019-nl},
BEAR \cite{Kumar2019-rw},
WOP \cite{Jaques2020-gb}), or
\textit{2)} with \emph{explicit} constraints on the \emph{value}
(\textit{e.g.}
BRAC \cite{Wu2019-nl},
CQL \cite{Kumar2020-zb}), or
\textit{3)} with \emph{implicit} constraints on the \emph{policy}
(\textit{e.g.}
BC \cite{Pomerleau1989-nh, Pomerleau1990-lm, Ratliff2007-fc, Bagnell2015-ni}
BCQ \cite{Fujimoto2018-mj},
MARWIL \cite{Wang2018-dn},
AWR \cite{Peng2019-hu},
ABM \cite{Siegel2020-lo},
CRR \cite{Wang2020-sr},
AWAC \cite{Nair2020-gd}).
Note, we here only list out the model-free approaches --- we will discuss the model-based ones
momentarily.
Optimizing an objective subjected to an implicit constraint corresponds to optimizing
a \emph{reduction} of an objective constrained explicitly.
After reduction (\emph{e.g.} via the Lagrangian method,
with KKT conditions assumed to be satisfied in case of inequality constraint)
the optimization problem is not constrained anymore, but it is still derived from an explicitly constrained problem,
hence the \textit{``implicit''} denomination.
The derivation from a constrained objective to the unconstrained one optimized in
MARWIL \cite{Wang2018-dn},
AWR \cite{Peng2019-hu},
ABM \cite{Siegel2020-lo},
CRR \cite{Wang2020-sr}, and
AWAC \cite{Nair2020-gd}
is laid out (albeit under a generalized form) in \textsc{Section}~\ref{pi}.
Seeing BCQ \cite{Fujimoto2018-mj}
as a perturbed, more sophisticated extension of BC,
itself consisting in maximizing the likelihood of the learned policy over the data,
we can equivalently interpret the objective of these algorithms as
the minimization of the forward KL divergence between the true data distribution and the learned policy.
As such, we \emph{can} see these (BC and BCQ) as \emph{implicitly} constrained.

Pretraining with offline datasets has been studied extensively in recent years.
The pursued goal is simply to leverage offline datasets so as to increase the learning speed and
final performance of downstream tasks in online RL, imitation learning \cite{Bagnell2015-ni},
and even offline RL (with a distinct task however).
Such desideratum is pursued notably in
\cite{Yang2021-ec},
\cite{Ajay2020-vj},
\cite{Singh2020-nm}, and
\cite{Nair2020-gd}.
Nonetheless, we are \emph{not} interested in this type of bootstrap and transfer capabilities in this work.

The endeavors we carry out in \textsc{Section}~\ref{pe}, including the formulation of the proposal
action distributions (from which the next actions are to be sampled) relate to a slew of works that
also attempted to skew the \textit{SARSA} variant of Bellman's equation (used in usual actor-critic's)
towards the \emph{optimal} variant of Bellman's equation (used in Q-learning
\cite{Watkins1989-ir, Watkins1992-gl}).
Although plenty of modifications to the Bellman operator have been proposed throughout the years
to entice the policy and value to adopt a certain desired behavior,
we are here only interested in approaching the \emph{optimal} action-value.
The discussion that follows therefore focuses on how previous works have successfully been able to
inject bias towards the optimal behavior in their respective agents.
We align the notion of \emph{optimality} with Bellman's principle of optimality.
As such, a policy is optimal if and only if its value is solution to the optimal version of Bellman's equation.
Whether an \textit{expert} dataset (possibly human-generated) should be considered \textit{optimal} is
a valid question --- discussed in \cite{Rust1992-wa}.
While one can just substitute the \textit{SARSA} operator with the Q-learning one
when dealing with discrete actions (\textit{cf.}~\cite{Crites1995-hn}
for the \textit{a priori} earliest adoption of such substitution in an actor-critic),
how to proceed in continuous action spaces is less obvious
(the $\max$ operation over actions creates an extra inner-loop optimization problem to solve \emph{every iteration}).
Different approaches have been adopted to circumvent this hindrance.
As such, we discern four main strategies to tackle the $\max$ operation in continuous action spaces:
\textit{1)} discretizing the continuous action space
into a discrete space of tractable dimension, then use Q-learning over the crafted discretized space
(\textit{e.g.} \cite{Millan2002-sl, Kimura2007-di, Metz2017-ol}),
\textit{2)} enforcing a strict structure over the Q-value in a way that facilitates the resolution of
said extra inner-loop maximization, as done \textit{e.g.} in NAF \cite{Gu2016-zd},
\textit{3)} getting rid of the hard $\max$ operation altogether by using a \emph{soft} update
instead \cite{Haarnoja2017-wj}, and finally
\textit{4)} tackling the resolution of inner-loop maximization problem every iteration,
despite the leap in computational cost caused by the continuous nature of the action space.
Approach \textit{4)} has, for instance, been implemented via the use of the derivative-free, black-box,
Cross-Entropy Method (CEM) \cite{Rubinstein1999-qt, Mannor2003-sj} optimizer
to estimate the maximizing action for the given state at each step,
\textit{e.g.} in QT-Opt \cite{Kalashnikov2018-ck, Quillen2018-wp},
and in the Actor-Expert framework \cite{Lim2018-ey}.
By contrast, the Amortized Q-learning (AQL) \cite{Van_de_Wiele2018-qs} method
follows approach \textit{4)} by taking inspiration from the amortized inference literature and
displays a conceptually simpler formulation.
In effect, these methods replace the optimal policy
that should in theory be used to generate the next action
in the SARSA variant of Bellman's equation (to obtain the Q-learning variant of Bellman's equation),
by a tractable approximation of it --- which we refer to as \emph{proposal} policy in
\textsc{Section}~\ref{pe}.
For example, QT-Opt \cite{Kalashnikov2018-ck, Quillen2018-wp}
craft said proposal policy as the function that, for the given state, at the given iteration,
returns the solution of the inner-loop,
per-iteration maximization sub-problem obtained via the CEM \cite{Rubinstein1999-qt, Mannor2003-sj}.
Tractable approximations of intractable policies in continuous action spaces
with proposal distributions over actions have also notably been carried out in \cite{Hunt2019-yu}
and \cite{Wiehe2018-id}.
Finally, the most noteworthy method implementing approach \textit{4)}
is perhaps the operator introduced in EMaQ \cite{Ghasemipour2020-ro},
in the offline regime specifically.
As we discuss in \textsc{Section}~\ref{pe}, this operator uses a proposal policy that interpolates between
a tractable approximation of the optimal policy estimated via AQL,
and an estimated clone of the policy underlying the offline dataset.
The framework we propose for policy evaluation in \textsc{Section}~\ref{pe} subsumes such operator.
Finally, the optimality \textit{tightening} \cite{He2017-si} technique
also aligns with the optimality desideratum pursued here.

Based on how brittle state-action values estimated via temporal-difference learning are in the offline regime
--- due to the risk of using out-of-distribution actions
for bootstrapping the value,
one might want to ensure or even guarantee \emph{safer} updates.
This safety desideratum naturally links offline RL
to the line of \textit{Safe Policy Improvement} (SPI)
methods (\textit{e.g.} \cite{Thomas2015-ex, Petrik2016-yc, Laroche2019-ar}), in the online regime.
As reminded in SPIBB \cite{Laroche2019-ar}
(SPI by \textit{Bootstrapping} the Q-value with a \textit{Baseline}, not with the learned policy),
the concept of \emph{safety} in RL is overloaded to say the least
(\textit{cf.}~\cite{Garcia2015-qk}).
Among all the possible hindrances one would desire to be shielded from
--- parametric (epistemic) uncertainty, internal (aleatoric) uncertainty,
interruptibility, exploration in a hazardous environment --- SPIBB sets out to design a method that
guarantees safety against the potential damages that might be caused by an excessive epistemic uncertainty
in the learned models.
SPIBB builds on the work of Petrik (\textit{cf.}~\cite{Petrik2016-yc}),
that designs a learning update rule for the actor that analytically guarantees a
safe policy improvement,
whatever the parameters of the model.
Aligning its take on safety with Petrik's,
SPIBB provides its agent access to a dataset as well as to a \emph{baseline}.
For most of their experimental endeavors, the dataset is assumed to be distributed as the baseline.
Crucially, that is a condition that need be satisfied for the theory backing SPIBB to hold.
Nevertheless, the authors also perform a set of experiments in which this
restrictive assumption is relaxed.
The relaxed scenario (the dataset was \emph{not} collected with the baseline)
is artificially implemented by replacing the dataset optained from the baseline
by a dataset collected by a random policy,
while the action bootstrap is still carried out using the baseline.
This second setting shares a property we investigate in \textsc{Section}~\ref{pe}:
how does detaching
the proposal policy used to bootstrap from
the policy that carried out the data collection,
impact performance.
We however do not have access to a baseline safeguard and our proposal policies
do not require any extra information; they are solely using the offline dataset (at most).
The safe policy improvement rule proposed by SPIBB articulates as follows:
when the agent is confident that it can improve upon the baseline,
it will use its own learned policy, otherwise it will use the oracle baseline as a fallback.
We take inspiration from these to formulate three of our proposal distributions,
dubbed \emph{safe}, and used in \textsc{Sections}~\ref{pe} and \ref{pi}.
As such, our work therefore also subsumes BRPO \cite{Sohn2020-ay},
which also proposes a learning update aligned with the one introduced in the SPI line of works.

By consistently opting for the safer option when there is a potential risk (based on an estimated measure of
uncertainty), SPI methods follow a \emph{pessimistic} heuristic, conceptually opposing the principle of
\textit{Optimism in the Face of Uncertainty Learning} (OFUL) which are
ubiquitous in the Multi-Armed Bandits (MAB) and Online Learning (OL) literature.
Note, any optimistic measure can trivially be made pessimistic by composing it with a monotonically decreasing
function over reals --- and optimistic measures of uncertainty or novelty are plentiful in the
MAB, OL, and online RL literature to infuse the agent with exploratory incentives.
Nonetheless, considering how the \emph{offline} agent are unable to learn from their own mistakes (since their
interactions are not recorded in the dataset used to train it), encourage exploration by advocating for
optimism in the face of uncertainty should be avoided.
Research endeavors in \emph{offline} RL have instead turned to
\emph{pessimism}, as analyzed through a theoretical lens in \cite{Buckman2020-rs}.
Pessimistic modifications to usual algorithms have recurrently been carried out via \emph{model-based} approaches,
consisting in replacing the rewards used by the learning agent with
pessimistic reward surrogates.
The first occurrence of such technique appears in the RaMDP technique propose by
Petrik in \cite{Petrik2016-yc},
where the author also introduced SPI,
stressing how both SPI and RaMDP promote the adoption of pessimism when confidence is low.
RaMDP transforms the rewards from the dataset via the application of a penalty.
A reward from a given transition in the dataset
will be penalized less if said transition appears often in the dataset.
Conversely, it will be penalized more if the transition appears rarely in the dataset.
The frequency of occurrence, used as a measure of confidence, is estimated via a
\emph{pseudo-count} \cite{Bellemare2016-ab, Tang2016-ys, Ostrovski2017-ww}.
Years later, MOPO \cite{Yu2020-vh}
and MoREL \cite{Kidambi2020-ph}
concurrently propose virtually identical model-based techniques penalizing the rewards in offline RL.
In contrast with RaMDP, they formulate their reward penalties
based on the uncertainty of a \emph{forward} model
(aiming to model the inner workings of the MDP).
On the topic of reward re-shaping, \cite{Lange2012-cc}
suggests that using a smoother reward signal might help in stabilizing the learned Q-value,
as first proposed and corroborated empirically in \cite{Hafner2011-rv}.
Yet, our preliminary investigation of reward smoothing did not yield improvement for the considered datasets.
In this work, we consider only \emph{model-free} approaches and, like mentioned earlier,
dabble in pessimism only via SPI-based techniques.

Finally, the involvement of constraints consisting of KL divergences in the policy improvement objectives
discussed and derived in \textsc{Section}~\ref{pi} echoes the entire \textit{``KL-control''} line of work,
originating in \cite{Kakade2001-hw, Kakade2002-kw, Kober2008-wb, Peters2008-mw, Vlassis2009-kc,
Theodorou2010-yy, Furmston2010-nz, Peters2010-vd, Kober2010-hy, Neumann2011-hn, Kappen2012-hp}.
Honorable mentions that distinguish themselves from these,
yet are tightly related,
could arguably be G-learning \cite{Fox2015-fr},
and $\psi$-learning \cite{Rawlik2013-cf}.
In the offline regime, WOP \cite{Jaques2020-gb}
notably implements a KL-control approach.

\section{Background}
\label{bg}

\subsection{Setting}

In this work, we tackle the problem of \emph{offline} RL
--- also sometimes referred to as \emph{batch} RL \cite{Lange2012-cc}:
the autonomous agent must learn how to interact optimally in an environment without being allowed to
interact with it during training.

On the flip side, our learning agent has access to a collection of interactions (including the received feedback
in the form of \emph{reward})
from another distinct agent.
This \emph{counterfactual} interactive information storage is called the offline dataset, noted $\mathcal{D}$.
The offline dataset is made available before the learning process starts, and is kept frozen throughout the
entirety of the training procedure. Note, we do not consider the \textit{``growing batch''} setting, in which
the dataset \emph{can} grow during training, either by the hands of the learning agent, or by an external source.
Leveraging solely the offline dataset $\mathcal{D}$,
the agent must learn an \emph{online}, interactive
policy that will enable the accumulation of the high rewards during \emph{evaluation} phases.
Note, we forbid the agent from accessing the offline dataset at evaluation time --- allowing it would place our
work in the \textit{``observational learning''} setting.
Albeit fairly realistic, we want our agents to be purely online when let loose for evaluation,
with no means of tapping into pre-existing repositories of interaction data.
As such, our agent uses only the offline dataset $\mathcal{D}$
without interacting online with its environment $\mathcal{E}$
at training time, but interacts online with $\mathcal{E}$
without ever using $\mathcal{D}$ at evaluation time
--- neither for \textit{reading} $\mathcal{D}$, nor \textit{writing} in $\mathcal{D}$.

\subsection{World and agent}

We model $\mathcal{E}$ as a memoryless, infinite-horizon, and stationary
Markov Decision Process (MDP) \cite{Puterman1994-pf}, noted $\mathbb{M}$.
Formally, $\mathbb{M} \coloneqq (\mathcal{S}, \mathcal{A}, p, \rho_0, u, \gamma)$, where
$\mathcal{S} \subseteq \mathbb{R}^n$ and $\mathcal{A} \subseteq \mathbb{R}^m$
are respectively the state space and action space.
The dynamics of the world are determined by the stationary, stochastic transition function $p$, and
the initial state probability density $\rho_0$.
In effect, $p(s' | s, a)$ is the conditional probability density
concentrated at the state $s'$ when action $a$ is executed in state $s$.
The reward feedback that $\mathcal{E}$ returns upon executing $a$ in $s$ is modeled as the outcome of
a stationary reward process $r$ that assigns real-valued rewards
distributed as $u(\cdot | s, a)$.
The remaining piece of $\mathbb{M}$ is $\gamma \in [0, 1)$, the discount factor.
The decision-making process of the learning agent is modeled by the parametric policy $\pi_\theta$,
under a neural representation with the parameter vector $\theta$.
The stochastic policy $\pi_\theta$ followed by the agent
maps states to probability distributions over actions, which we denote by
$\pi_\theta : \mathcal{S} \to \mathcal{P}(\mathcal{A})$, or by the compact notation
$\pi_\theta \in \mathcal{P}(\mathcal{A})^\mathcal{S}$, where
$\mathcal{P}(\mathcal{A})^\mathcal{S}
\coloneqq \{ \pi \, | \, \pi: \mathcal{S} \to \mathcal{P}(\mathcal{A}) \}$.
Concretely, the experiences of the agent are divided across discrete timesteps $t$, where $t\geq0$.
The analyses we carry out in this work do not require the involvement of a finite time horizon $T$,
hence our decision to adopt the infinite-horizon MDP setting
where $t$ is \textit{a priori} unbounded above in the
formalism.
Despite being infinite horizon, we make the MDP episodic
by assuming that every trace contains at least one absorbing state,
and that when the first is reached by the agent, $\gamma$ is artificially set to zero
to emulate termination, hence formally constructing an \textit{episode}.
At each timestep $t\geq0$, the agent is located at $s_t \in \mathcal{S}$,
and concentrates a probability density over actions from $\mathcal{A}$ that is equal to
$\pi_\theta(a | s_t)$ at action $a$.
We denote the action selected by the agent's policy $\pi_\theta$ at timestep $t$ by $a_t$,
and the received reward by $r_t$.

Lastly, we introduce the discounted state visitation frequency for an agent following
a policy $\pi$
in the MDP $\mathbb{M}$, denoted as $\rho^\pi_\mathbb{M}$,
and abbreviated $\rho^\pi$ in the absence of ambiguity.
Formally,
$\rho^\pi_\mathbb{M} (s)
\coloneqq
\sum_{t=0}^{+\infty} \gamma^t \mathbb{P}^\pi_\mathbb{M} [S_t=s]$,
where $\mathbb{P}^\pi_\mathbb{M} [S_t=s]$ is the probability of reaching state $s$
at timestep $t$ ($S_t$ is a random variable) when following $\pi$ in $\mathbb{M}$.
Since an immediate derivation gives $\sum_{s \in \mathcal{S}} \rho^\pi_\mathbb{M}(s) = 1 / (1-\gamma)$,
$\rho^\pi_\mathbb{M}$ can be seen as a probability distribution over states up to a constant factor.
Being in the episodic setting, we will use the \emph{undiscounted} counterpart of $\rho^\pi_\mathbb{M}$,
yet still artificially set $\gamma=0$ when the absorbing state (posited earlier to always exist)
is reached to emulate episode termination.

\subsection{Dataset}

We do not have access to the analytical form of the policy that generated the dataset,
nor do we have the ability to sample from it.
Besides, since none of the approaches mentioned in this work explicitly leverage the fact that the offline dataset might
have been produced by multiple distinct sources,
we posit \textit{w.l.o.g.} that the offline dataset $\mathcal{D}$ contains interaction traces
of a single conceptual policy $\beta$, dubbed the \emph{offline} distribution.
Despite being composed of traces of interaction, these are not necessarily available in connex trajectories, but
rather as a shuffled collection of individual transitions.
We do not know how the data was collected, in particular what the underlying strategy $\beta$ was optimizing for,
nor do we know the proficiency of $\beta$ in satisfying the chased objective other
than what we can infer and hopefully extrapolate from the offline dataset.
$\mathcal{D}$ might have been collected with a single fixed snapshot of a policy, in which case $\beta$ is this
very snapshot; or $\mathcal{D}$ might be the training history of a policy, in which case $\beta$
is a mixture of past iterates.
For a given dataset $\mathcal{D}$,
$\beta$ might be focusing on covering the most ground while not paying too much attention
to the collected rewards, or conversely covering the least amount of ground while accumulating
rewards as greedily as possible.
Note, in imitation learning and inverse RL \cite{Bagnell2015-ni}
where rewards are not known,
one assumes the \textit{expert} ($\beta$'s counterpart) was acting optimally when collecting the demonstrations.
This is \emph{not} the case in this work.
We use datasets whose quality range from \textit{expert}-grade
data (where $\beta$ is close to optimality)
to \textit{random} data (where $\beta$ wanders seemingly aimlessly)
(\textit{cf.}~\textsc{Appendix}~\ref{experimentalsetting}).
Our agents are never made aware of $\mathcal{D}$'s quality or $\beta$'s proficiency.
The offline dataset $\mathcal{D}$ is formally defined as a collection of \textit{SARS}-formatted transitions
$(s, a, r, s')$
collected by the underlying offline distribution $\beta$ through interactions with $\mathbb{M}$.
Being in the episodic setting, transitions also contain a termination indicator
in practice, taking value $1$ when the
transition is the last one in the episode, and $0$ otherwise.
When $\mathcal{D}$ has a richer structure --- from transitions being \textit{SARSA}-formatted to transitions being
sequenced in full connex trajectories --- we say so explicitly in the text.
As noted in \cite{Fujimoto2018-mj} and \cite{Laroche2019-ar},
we have in effect two MDPs in the tackled offline setting:
\textit{1)} the real, non-observable, online MDP $\mathbb{M}$ underlying the
inaccessible environment $\mathcal{E}$, and
\textit{2)} the fictitious, observable, offline MDP $\mathbb{M}^\textsc{off}$
effectively communicated through the dataset to the agent.
As such, while $\beta$ interacted with $\mathbb{M}$ to collect $\mathcal{D}$,
$\pi_\theta$ interacts with $\mathbb{M}^\textsc{off}$
--- in effect \emph{detached} from the real world $\mathcal{E}$
--- and collects nothing.
Every state $s$ in $\mathcal{D}$ is then distributed as $\rho^\beta_\mathbb{M}(\cdot)$.
For legibility purposes, we will use $\rho^\beta$
as a shorthand for $\rho^\beta_\mathbb{M}$.
In practice, with a slight abuse of notation,
we can note $(s,a,r,s') \sim \mathcal{D}$ to indicate that the transition $(s,a,r,s')$
is in effect obtained by sampling from the offline dataset $\mathcal{D}$.
Nevertheless, we will often opt for the explicit notation
$\mathbb{E}_{s \sim \rho^\beta(\cdot), a \sim \beta(\cdot | s),
s' \sim \rho^\beta(\cdot)}[\cdot]$ --- for \textit{SARS}-formatted transitions,
and $\mathbb{E}_{s \sim \rho^\beta(\cdot), a \sim \beta(\cdot | s),
s' \sim \rho^\beta(\cdot), a' \sim \beta(\cdot | s')}[\cdot]$ for \textit{SARSA}-formatted transitions ---
as a drop-in replacement for
$\mathbb{E}_{(s,a,r,s') \sim \mathcal{D}}[\cdot]$
and $\mathbb{E}_{(s,a,r,s',a') \sim \mathcal{D}}[\cdot]$,
with $r$ being replaced by $r(s,a,s')$ in the operand of the expectation
not to overload the $\mathbb{E}$ notation.
We express the reward function as a function of the next state $s'$ as well,
as it is the most general setting and can be reduced to $r(s,a)$ by positing
trivial assumptions.
We do not indicate the states (and actions when applicable)
at which the conditional densities are evaluated in the outer expectations throughout this work,
to lighten the notation as much as possible.

\subsection{Objective}

We here describe the concepts that we need to add to our tool set so as to properly deal with the
delayed nature of the reward feedback.
In the infinite-horizon regime we placed our agent into, the \emph{return}
$R_t^\gamma$ is
the discounted sum of rewards from timestep $t \geq 0$ onwards, and is at the core of how RL methods
attempt to solve the credit assignment problem created by said delayed feedback. We formalize the return
of the state-action pair associated with timestep $t$ as
$R_t^\gamma \coloneqq \sum_{k=0}^{+\infty} \gamma^k r_{t+k}$.
The expectation of the return along traces of the arbitrary policy $\pi$ starting from the execution of
action $a_t$ in $s_t$ defines the state-action value $Q^\pi$ (also called action-value, or simply Q-value)
such as
$Q^\pi(s_t, a_t) \coloneqq
\mathbb{E}_{
s_{t+1} \sim p(\cdot | s_t, a_t),
a_{t+1} \sim \pi(\cdot | s_{t+1}), \ldots}
[R_t^\gamma]$
(\textit{abbrv.} $\mathbb{E}_\pi^{>t}[R_t^\gamma]$).
We say that a policy $\pi$ acts \textit{optimally}
at timestep $t \geq 0$ if the action $\pi$ selects at
state $s_t$, noted $a_t$ verifies the following identity:
$a_t = \argmax_{a \in \mathcal{A}} Q^\pi (s_t, a)$.
Such behavior coincides with $\pi$ being purely greedy with respect to the \textit{exact}
action-value $Q^\pi$  coupled with $\pi$ (\textit{cf.}~Q-value definition right above).
By extension, $\pi$ acts optimally at \emph{every} timestep $t \geq 0$ if and only if,
for any given start state $s_0 \sim \rho_0$, the policy $\pi$ maximizes
$V^\pi (s_0) \coloneqq \mathbb{E}_{a_0 \sim \pi(\cdot | s_0)}[Q^\pi(s_0, a_0)]$.
Our objective is for the agent to learn a policy $\pi_\theta$ such that $\pi_\theta$
is \emph{a} solution of the optimization problem
$\pi_\theta \in \argmax_{\pi \in \Pi} U_0(\pi)$
for any given start state $s_0 \sim \rho_0$,
where $U_t(\pi) \coloneqq V^\pi(s_t)$ is the performance objective
(or \emph{u}tility)
and $\Pi$ is the neural search space for $\pi_\theta$.

\subsection{Convenience}
\label{convenience}

\paragraph{KL.}

We adopt an alternative notation for the KL divergences
by adding an arrow within the operator in order to disambiguate
from a quick glance whether it is the \textit{forward} (\textit{inclusive} \cite{MacKay2003-qn})
or the \textit{reverse} (\textit{exclusive} \cite{MacKay2003-qn})
KL divergence,
due to its asymmetric nature. The orientation of said arrow indicates the order in which one should write the
predicted and target distributions respectively in the original KL divergence notation.
As such, denoting the predicted and target distributions as $q_\theta$ and $p$ respectively
--- the parameter ``$\theta$'' being used here to highlight which distribution is
parameterized and learned, and which is fixed and set as target ---
we write, for a given state $s$,
$D^p_{\overrightarrow{\textsc{kl}}}[q_\theta](s)$
to replace $D_{\textsc{kl}}\big(p(\cdot | s) \, || \, q_\theta(\cdot | s)\big)$, and
$D^p_{\overleftarrow{\textsc{kl}}}[q_\theta](s)$
instead of $D_{\textsc{kl}}\big(q_\theta(\cdot | s) \, || \, p(\cdot | s)\big)$.
With this notation, the target distribution is always indicated in the exponent, of the divergence operator,
while the learned distribution to fit to the target is alone in the operand of the operator.
By following the arrow, we can then read, without ambiguity, for a given state $s$,
\textit{a)} $D^p_{\overrightarrow{\textsc{kl}}}[q_\theta](s)$ as
\textit{``the forward KL between $q_\theta$ and $p$''}, and
\textit{b)} $D^p_{\overleftarrow{\textsc{kl}}}[q_\theta](s)$ as
\textit{``the reverse KL between $q_\theta$ and $p$''}.

\paragraph{Hyper-parameter stiffness.}

Finally, we propose the notion of \textit{hyper-parameter stiffness}, defined by analogy with the notion
of stiff equation in physics:
we say that an algorithm $A$ is \textit{stiff} with respect to the hyper-parameter
$\lambda$ if slight changes in the value of $\lambda$ cause large and non-monotonic
variations in $A$'s asymptotic performance,
--- or, to a more extreme extent, make $A$ numerically unstable.
In effect, from a practitioner's perspective, that translates into $\lambda$ being tedious to tune.
We introduce this notion due to various offline RL methods having been reported for their
sensitivity and brittleness \textit{w.r.t.} hyper-parameter choices ---
for instance in \cite{Le_Paine2020-sb} and \cite{Monier2020-tq}.
The universal desideratum is to design algorithms $A$ that are stiff \textit{w.r.t.}
none of their hyper-parameters $\lambda$,
and this is naturally what we aspire to in this work.

\section{Critical evaluation of the offline RL landscape}
\label{landscape}

\subsection{Experimental setting}
\label{baselines}

As the first contribution of this work, we perform a thorough PyTorch \cite{Paszke2019-zf}
re-implementation of the state-of-the-art
competing baselines in offline RL under a common computational infrastructure,
and release it as an open-source\footnote{Code made available at the URL:
\texttt{https://github.com/lionelblonde/giwr-pytorch}.} project.
This allows not only for meaningful comparisons in which we can vary in a controlled manner
different design choices, but also for the easy development of new offline RL algorithms.
We lay out a comprehensive exposition of our re-implementations,
and the associated design choices we made to ensure their fairness,
in \textsc{Appendix}~\ref{baselinesdetails}.

We carry out every experiment
reported throughout this work
in the D4RL suite \cite{Fu2020-ic},
composed of environment-dataset pairs.
We give more details about how we use the suite to carry out our empirical investigations in
\textsc{Appendix}~\ref{evalsuite}.

In this work
(and by extension, in the released codebase),
we investigate, analyze, and evaluate the following methods:
SAC \cite{Haarnoja2018-bm},
D4PG \cite{Barth-Maron2018-ot},
BCQ \cite{Fujimoto2018-mj},
BEAR \cite{Kumar2019-rw},
CQL \cite{Kumar2020-zb},
BRAC \cite{Wu2019-nl},
BC \cite{Pomerleau1989-nh, Pomerleau1990-lm, Ratliff2007-fc, Bagnell2015-ni},
CRR \cite{Wang2020-sr}, and
AWR \cite{Peng2019-hu}
(\textit{cf.}~\textsc{Section}~\ref{relatedwork} for a categorization of these algorithms in
semantic families by how they enforce \emph{closeness} with respect to the dataset $\mathcal{D}$
to avoid the hindering distributional shift).
These are all off-policy actor-critic \cite{Sutton1984-ce, Konda2000-ef, Konda2003-jh, Degris2012-gv}
architectures --- with the exception of BC, which is a pure imitation learning \cite{Bagnell2015-ni}
method (more on that momentarily).
The actor or policy is modeled with $\pi_\theta$ (introduced in \textsc{Section}~\ref{bg});
the critic with the parametric model $Q_\omega$.
We study actor-critics and not pure value methods
since we tackle \emph{continuous} action spaces in this work
(\textit{cf.}~\textsc{Appendix}~\ref{evalsuite} where we describe the
continuous control tasks
considered for this present work).
In all of these implementations, we tried to strike the right balance between
\textit{a)} following the official implementation when made available by the authors,
\textit{b)} using the hyper-parameter values recommended in the associated paper (when not conflicting with
the official implementation), and
\textit{c)} making the algorithms computational comparable in terms of flops, number of parameters,
runtime, difficulty of implementation, accessibility to privileged information.
Concretely, every single baseline
\textit{a)} starts learning \emph{from scratch} (no warm-start),
\textit{b)} is provided with exactly the same information as input
(\emph{only} the offline dataset $\mathcal{D}$ at training time,
\emph{only} the online environment $\mathcal{E}$ modeled by the MDP $\mathbb{M}$ at evaluation time),
\textit{c)} is given access to the exact same computational resources (1 modern high-end GPU),
and
\textit{d)} is allowed the same maximum runtime (12 hours).
We delve deeper into our re-implementations
and evaluations in \textsc{Appendices}~\ref{baselinesdetails} and
\ref{evalprotocol} respectively.

\subsection{Empirical evaluation}
\label{experimentalassessment}

\begin{figure}
  \begin{subfigure}{\textwidth}
    \caption{Return upon training completion}
    \center\scalebox{0.18}[0.18]{\includegraphics{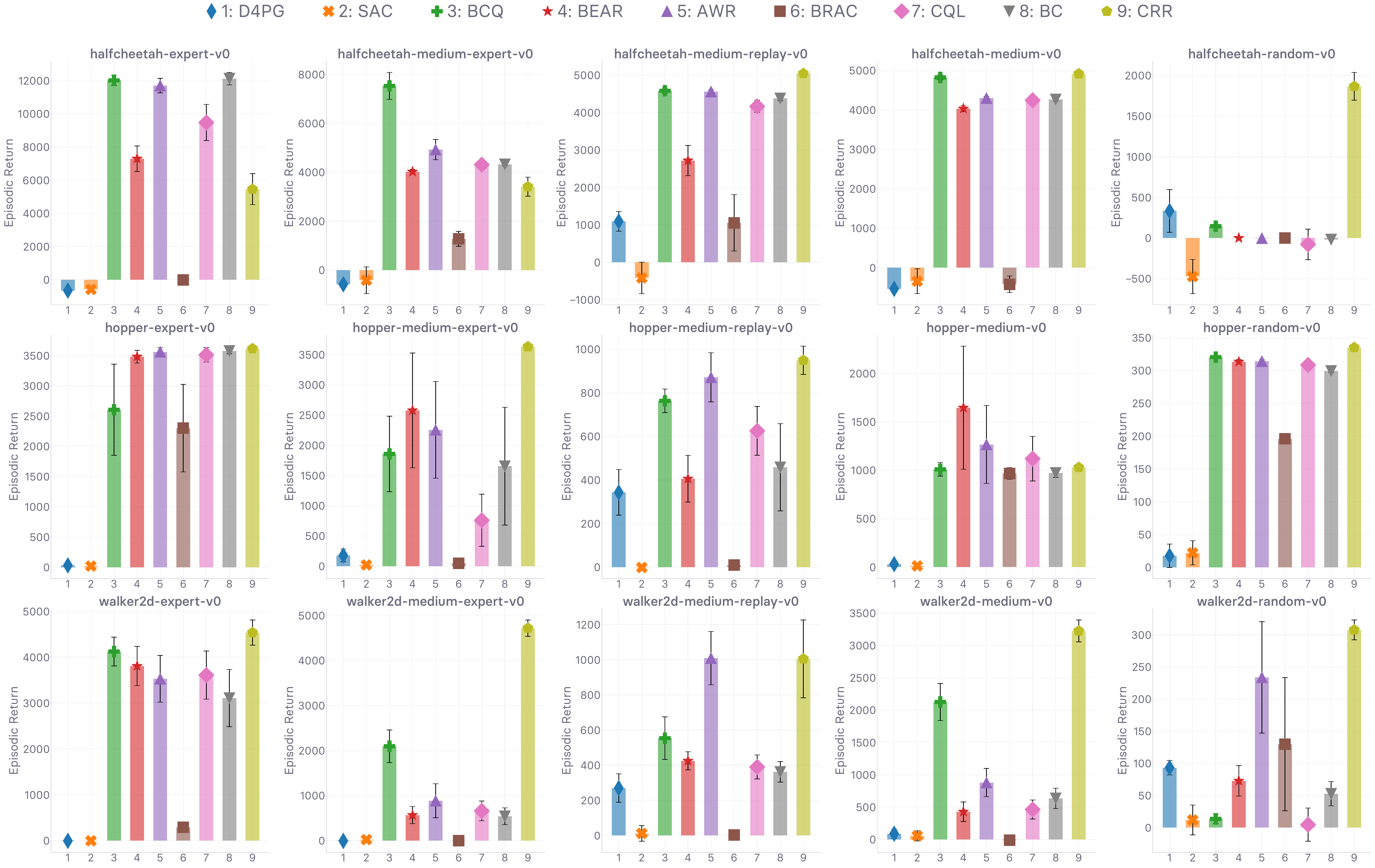}}
  \end{subfigure}
  \begin{subfigure}{\textwidth}
    \caption{Evolution of the return during training}
    \center\scalebox{0.18}[0.18]{\includegraphics{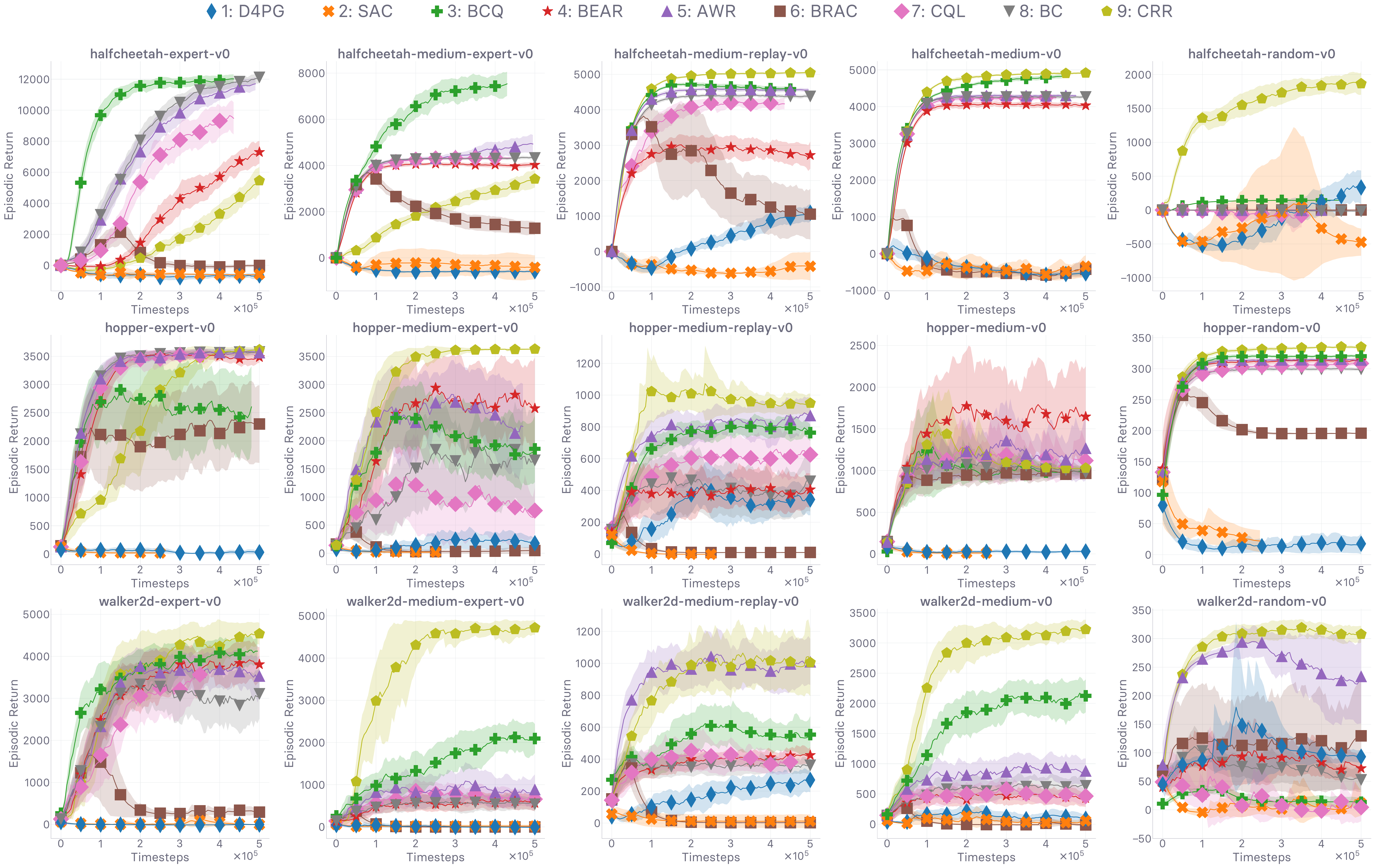}}
  \end{subfigure}
  \caption{Empirical evaluation of our unified re-implementations of the offline RL baselines:
  SAC \cite{Haarnoja2018-bm},
  D4PG \cite{Barth-Maron2018-ot},
  BCQ \cite{Fujimoto2018-mj},
  BEAR \cite{Kumar2019-rw},
  CQL \cite{Kumar2020-zb},
  BRAC \cite{Wu2019-nl},
  BC \cite{Pomerleau1989-nh, Pomerleau1990-lm, Ratliff2007-fc, Bagnell2015-ni},
  CRR \cite{Wang2020-sr}, and
  AWR \cite{Peng2019-hu}.
  (a) The first three rows give the return mean and standard deviation on training completion.
  (b) The last three rows give the evolution of the return.
  Runtime is 12 hours. Best seen in color.}
  \label{baselines:barplot}
\end{figure}

In \textsc{Figure}~\ref{baselines:barplot},
we depict the empirical comparison of the offline RL baselines laid out in \textsc{Section}~\ref{baselines},
following the evaluation protocol described in \textsc{Appendix}~\ref{evalprotocol}.
\textsc{Figure}~\ref{baselines:barplot} does not allow us to name a definitive \emph{best} approach,
as none of the baselines seem to perform well across the board.
Yet, overall, it looks like the three main contenders for the title would be
BCQ \cite{Fujimoto2018-mj},
CRR \cite{Wang2020-sr}, and
AWR \cite{Peng2019-hu}.
As expected (and documented in past offline RL literature),
the port of
SAC \cite{Haarnoja2018-bm}
to the offline regime (described in \textsc{Section}~\ref{baselines})
performs extremely poorly in every single dataset and environment.
Motivated by the promise of distributional RL \cite{Bellemare2017-yr, Dabney2017-to, Dabney2018-ya, Dabney2020-nm}
in the offline setting underlined by REM \cite{Agarwal2020-eu},
D4PG \cite{Barth-Maron2018-ot}
displays higher returns than SAC \cite{Haarnoja2018-bm},
but is still mediocre compared to the baselines designed specifically for the offline regime.
Nonetheless, the contrast in performance between SAC and D4PG
suggests we might prompt noteworthy return gains simply by replacing the traditional critic used in
natively offline RL methods with a distributional one.
At first glance, it might appear as surprising that behavioral cloning (BC) gathers such high returns
in a number of cases given that it is a
\emph{pure} imitation learning \cite{Bagnell2015-ni} approach.
Indeed, a BC agent need only use the state-action pairs $(s,a)$ extracted from the transitions provided through
the offline dataset $\mathcal{D}$, which is considerably less information than what the usual offline RL
approaches use.
BC discards the reward signal present in every transition,
which essentially acts as a score on the decision made by $\beta$ to execute action $a$ at state $s$
in the MDP $\mathbb{M}$.
Instead, the BC agent only knows what $\beta$ did, without knowing how proficient $\beta$ was
at accumulating rewards.
As such, it is to be expected that BC \emph{only} performs well when $\beta$ does well, and becomes
increasingly worse as the $\beta$ underlying the offline dataset $\mathcal{D}$ strays from optimality.
This is indeed what we observe in \textsc{Figure}~\ref{baselines:barplot}: BC's return is highest for
the \textit{expert} datasets, and decreases relatively to its counterparts as we look
from left (expert) to right (random).
Since BC does not use the rewards in $\mathcal{D}$,
it does not possess counterfactual learning abilities \cite{Bottou2013-so}.
Instead, it can only replicate the behavior demonstrated in the offline dataset with extremely limited extrapolation
capabilities,
and therefore can only achieve optimality if $\beta$ is optimal for the task.
Besides, BC is incredibly easier to implement that most of its baseline counterparts, which heavily plays in
the imitation learning method's favor, from a purely practical perspective.
CRR comes in a close second if we were to rank the methods by that criterion.

\paragraph{What is \textit{the best} offline RL method?}

From this analysis alone, one is unable to name a victor across the range of dataset qualities
for all environments.
Guidelines as to which offline RL algorithm is the best need be
conditioned on the dataset quality
(\textit{i.e.} on how optimal one believes $\beta$ to be),
which is what we tackle in the following investigation.

\section{Dataset-grounded optimality inductive biases}
\label{inductivebiases}

\subsection{Conceptualization}
\label{conceptualization}

In \textsc{Figure}~\ref{baselines:barplot},
we observe that CRR
outperforms its competitors except in a handful of datasets and environments.
In contrast to BC which imitates the policy $\beta$ underlying the offline dataset $\mathcal{D}$,
CRR only enforces the actor's policy $\pi_\theta$ to remain \emph{somewhat} close to $\beta$
via an \emph{implicit} constraint (\textit{cf.}~\textsc{eq}~\ref{piineq} for the generalized version of
said constraint, and (\textit{cf.}~\textsc{Section}~\ref{pi} for a derivation that could trivially be boiled down
to obtain CRR's actor objective \textit{exactly}).
\textbf{In an effort to ground the discussion with more conceptual formalism, we introduce
the notion of \emph{optimality inductive bias} $\mathcal{B}$, grounded on the offline dataset $\mathcal{D}$.
Concretely, the amount of bias $\mathcal{B}$, noted $B$, injected into an agent
represents to what degree the agent builds upon the belief that $\beta$ underlying $\mathcal{D}$
is \emph{optimal} to update its policy $\pi_\theta$.}
Crucially, note, we do not built $\mathcal{B}$ to be aligned with the
\emph{genuine} intrinsic quality of the data present in $\mathcal{D}$,
but on the agent's belief that $\mathcal{D}$ contains \textit{expert}-grade data
--- that $\beta$ is optimal.
For example, BC
injects the maximum possible amount of bias $\mathcal{B}$,
$B_\textsc{max}$, into the agent since it \emph{posits} by design that the
provided demonstrations are originating from an optimal expert policy.
In fact, this statement applies to every method following the
imitation learning paradigm.
Natively-offline baselines that enforce an explicit closeness constraint with $\beta$ to avoid any
distributional shift caused by out-of-distribution actions inject an inductive bias $B_\textsc{expl}$
--- and by symmetry,
$B_\textsc{impl}$ for the ones that involve an implicit constraint, like CRR.
Intuitively, we loosely have: $B_\textsc{max} \geq B_\textsc{expl} \geq B_\textsc{impl} \geq 0$.
One can move $B_\textsc{expl}$ and $B_\textsc{impl}$ within the interval
$[0, B_\textsc{max}]$ by increasing the scaling coefficient associated with the constraint enforcing
said \textit{closeness} of $\pi_\theta$ with respect to $\beta$ (equivalently, $\mathcal{D}$).
Notably, we found that, in the case of BRAC \cite{Wu2019-nl},
--- and to a lesser extent in the case of CQL \cite{Kumar2020-zb} ---
finding the right level of bias $B_\textsc{expl}$ to inject,
proved to be tedious, and remarkably difficult to tune.
The scaling coefficients controlling the injection of the optimality inductive bias in these methods
thus qualify as stiff (in line with the notion of \emph{stiffness} we have defined in \textsc{Section}~\ref{bg}).
A similar observation has been made by \cite{Monier2020-tq} for CQL (BRAC was not tackled there).
Interestingly, BCQ \cite{Fujimoto2018-mj}
involves neither an explicit nor an implicit constraint between $\pi_\theta$ and $\beta$.
Rather, we would categorize BCQ as a perturbed imitation learning method.
As such, it injects an inductive bias $B \approx B_\textsc{max}$ into the agent.
This is clearly illustrated in \textsc{Figure}~\ref{baselines:barplot}, where BCQ
performs well on expert dataset, yet poorly on random datasets.

As a rule of thumb, the closer the injected bias $B$ is to $B_\textsc{max}$ (pure imitation learning),
\textit{a)} the better the method performs in \emph{expert} datasets, and
\textit{b)} the worse it performs in \emph{random} ones, on the other side of the quality spectrum.
Intuitively, treating everything as equally valuable in $\mathcal{D}$
is a bad idea if it is not the case, but is optimal if it is indeed the case.
From a practitioner's perspective, it then all comes down to how much is known about the contents of
the offline dataset.
Consider a condition, dubbed $\mathcal{C}$, that is verified whenever we \emph{know} that $\beta$ is \emph{optimal}
for the task.
When $\mathcal{C}$ is satisfied (we know that $\beta$ is optimal),
one should inject an inductive bias $B \approx B_\textsc{max}$ into the agent
(\textit{e.g.} via BCQ
or via an imitation learning method like BC).
Conversely, when $\mathcal{C}$ is \emph{not} satisfied (either we do not know at all what is in $\mathcal{D}$
quality-wise, or we know that $\beta$ is sub-optimal),
one should inject an inductive bias $B < B_\textsc{max}$ into the agent
(\textit{e.g.} via CRR).
Rephrasing what precedes, based on our results in \textsc{Figure}~\ref{baselines:barplot} it seems that the
\emph{best} course of action is for the offline RL practitioner to:
\textit{a)} use BC or BCQ (or any other method with high dataset-grounded bias)
when $\mathcal{C}$ is satisfied, and
\textit{b)} use CRR (which is in effect with an \emph{advantage re-weighted} BC)
when $\mathcal{C}$ is not satisfied.

In the next section, we corroborate these statements
empirically by showing that increasing the optimality bias of CRR
in a minimalist and parsimonious fashion quickly makes the resulting method better in expert datasets and
worse in random ones.
Crucially, the same method can achieve state-of-the-art performance across the considered spectrum
of datasets qualities via the adjustment of a single hyper-parameter, provided one knows whether
$\beta$ is optimal or not.

\paragraph{Algorithmic base.}

From this point forward,
in the entire remainder of this work,
\textbf{we will use CRR \cite{Wang2020-sr} (or equivalently, AWAC \cite{Nair2020-gd})
as base algorithm,
since it is the method that seems to perform consistently well across the board}
(across environments and dataset quality levels).
Given the central role it plays in what follows in this work,
we lay out the algorithm
in \textsc{Algorithm}~\ref{algobase}
under the name \textsc{Base}, which denotes either CRR or AWAC indifferently.

\IncMargin{1em}
\begin{algorithm}
\SetAlgoLined
\SetNlSty{}{}{}
\SetKwInput{KwInit}{init}
\KwInit{initialize
the random seeds of each framework used for sampling,
the random seed of the environment $\mathbb{M}$,
the neural function approximators' parameters
($\theta$ for the actor's policy $\pi_\theta$, and $\omega$ for the critic's action-value $Q_\omega$),
the critic's target network $\omega'$ as an exact frozen copy,
the offline dataset $\mathcal{D}$.}
\While{no stopping criterion is met}{
    \tcc{Train the agent in $\mathbb{M}^\textsc{off}$}
    Get a mini-batch of samples from the offline dataset $\mathcal{D}$\;
    Perform a gradient \emph{descent} step along
        $\nabla_\omega \, \ell_\omega$
        (\textit{cf.} below)
        using the mini-batch\;
    $$
    \ell_\omega \coloneqq
    \mathbb{E}_{s \sim \rho^\beta(\cdot), a \sim \beta(\cdot | s), s' \sim \rho^\beta(\cdot)}
    \bigg[
    \Big(
    Q_\omega(s,a) -
    \big(
    r(s, a, s') + \gamma \, \mathbb{E}_{a' \sim \pi_\theta(\cdot | s')}
    \big[
    Q_{\omega'}(s',a')
    \big]
    \big)
    \Big)^2
    \bigg]
    $$
    where $r(s, a, s')$ was introduced as syntactic sugar in \textsc{Section}~\ref{bg}\;
    Perform a gradient \emph{ascent} step along
        $\nabla_\theta \, \mathcal{U}_\theta$
        (\textit{cf.} below)
        using the mini-batch\;
    $$
    \mathcal{U}_\theta \coloneqq
    \mathbb{E}_{s \sim \rho^\beta(\cdot), a \sim \beta(\cdot | s)}
    \bigg[
    \exp (\frac{1}{\tau} A^{\pi_\theta}_\omega(s,a)) \log \pi_\theta(a | s)
    \bigg]
    $$
    where
    $A^{\pi_\theta}_\omega(s,a) \coloneqq Q_\omega(s,a) -
    \mathbb{E}_{\bar{a} \sim \pi_\theta}[Q_\omega(s,\bar{a})]$,
    and $\tau$ is a temperature hyper-parameter\;
    Update the target network $\omega'$ using the new $\omega$\;
    \tcc{Evaluate the agent in $\mathbb{M}$}
    \If{evaluation criterion is met}{
        \ForEach{evaluation step per iteration}{
            Evaluate the empirical return of $\pi_\theta$
                in $\mathbb{M}$
                (\textit{cf.} evaluation protocol in \textsc{Appendix}~\ref{evalprotocol})\;
        }
    }
}
\caption{\textsc{Base} (denotes either CRR \cite{Wang2020-sr} or AWAC \cite{Nair2020-gd} indifferently)}
\label{algobase}
\end{algorithm}
\DecMargin{1em}

\subsection{Evidence}
\label{evidence}

We now investigate an extension of
the \textsc{Base} approach.
We carry out a thorough analysis of the behavior of the method resulting from the addition of
the CQL \cite{Kumar2020-zb} constraints in \textsc{Base}.
Adding these constraints in effect provides us with a finely controllable handle on the further
injection of inductive bias $\mathcal{B}$
(\textit{cf.}~\textsc{Section}~\ref{experimentalassessment}) in \textsc{Base}.
These constrains introduced by CQL will be explicitly reported momentarily.
We call the composite method \textit{``Reinforce The Gap''} (\textit{abbrv.} RTG).
The notion of \textit{gap} (noted $\Delta_\textsc{gap}$) we use here aligns with the one introduced in
CQL \cite{Kumar2020-zb}:
\begin{align}
  \Delta_\textsc{gap} \coloneqq
  \mathbb{E}_{s \sim \rho^\beta(\cdot), a \sim \beta(\cdot | s)}
  \Big[
  \max
  \big\{
  Q_\omega\big(s, a^i\big) \, | \, a^i \sim \text{unif}(\mathcal{A}[s])
  \big\}_{i \in [1,m] \cap \mathbb{N}}
  - Q_\omega(s,a)
  \Big]
  \label{gap}
\end{align}
where $\mathcal{A}[s]$ is the set of actions from $\mathcal{A}$ that are feasible in state $s$.
The observation made in CQL is that the introduced constraints have the expected effect of
increasing the maximum gap $\Delta_\textsc{gap}$ (\textit{cf.} definition in \textsc{eq}~\ref{gap})
in action-value between random, uniformly-sampled actions, and actions from
the offline dataset $\mathcal{D}$
at a given state from $\mathcal{D}$.
We will report these gaps for both methods (\textsc{Base} with and without CQL constraints) momentarily.
Despite only being ---
in the context of our work ---
a toy extension of \textsc{Base} that allows us to study the bias $\mathcal{B}$
more closely in a controlled environment,
RTG also appears (very recently) in \cite{Monier2020-tq}
as the combination of two state-of-the-art offline RL methods.
As the direct combination of CQL and CRR,
\cite{Monier2020-tq} names the method \textit{conservative} CRR (\textit{abbrv.}~CCRR).
CCRR was empirically evaluated in a handful of datasets of different qualities.
We propose a far more fine-grained dataset design technique that enables us to finely control the percentage
of random (or expert) data in the dataset.

\begin{figure}
  \begin{subfigure}{\textwidth}
    \caption{Return}
    \label{basertg:barplot}
    \center\scalebox{0.18}[0.18]{\includegraphics{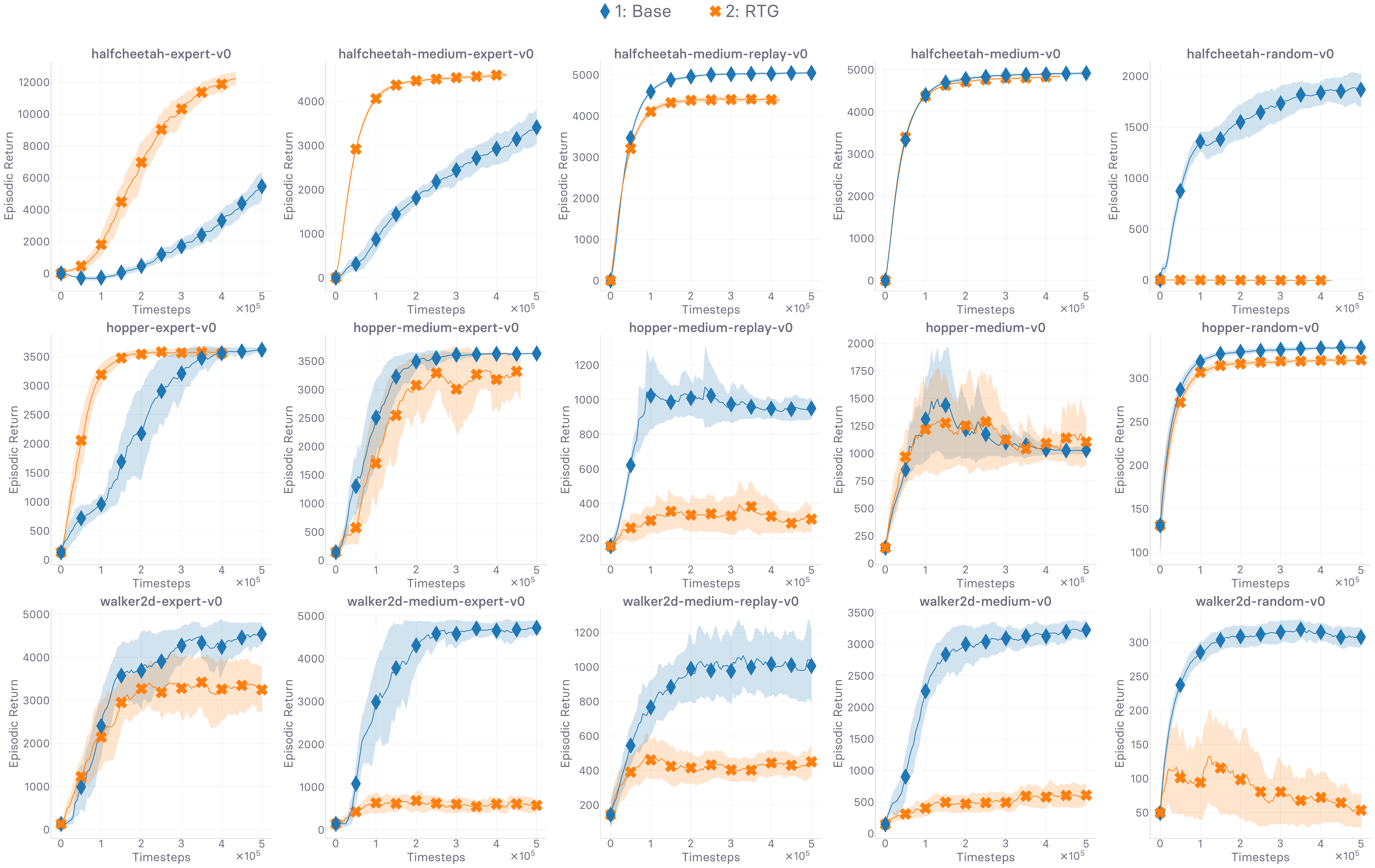}}
  \end{subfigure}
  \begin{subfigure}{\textwidth}
    \caption{Gap}
    \label{basertggap:barplot}
    \center\scalebox{0.18}[0.18]{\includegraphics{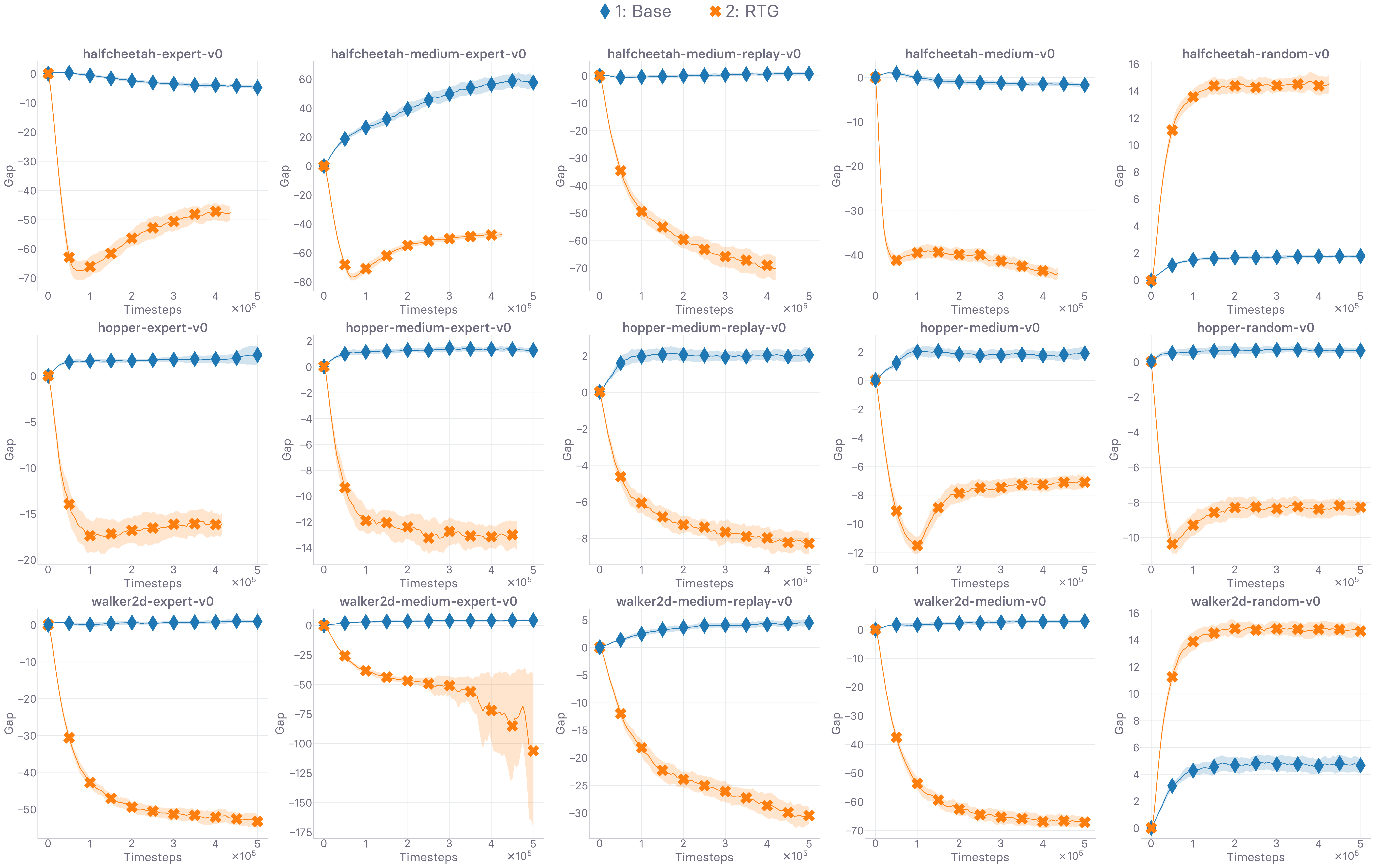}}
  \end{subfigure}
  \caption{Empirical evaluation of the (a) return
  and (b) gap (\textit{cf.}~\textsc{eq}~\ref{gap})
  of \textsc{Base} and RTG.
  Runtime is 12 hours.}
\end{figure}

\begin{figure}
  \begin{subfigure}{\textwidth}
    \caption{Return upon training completion}
    \center\scalebox{0.18}[0.18]{\includegraphics{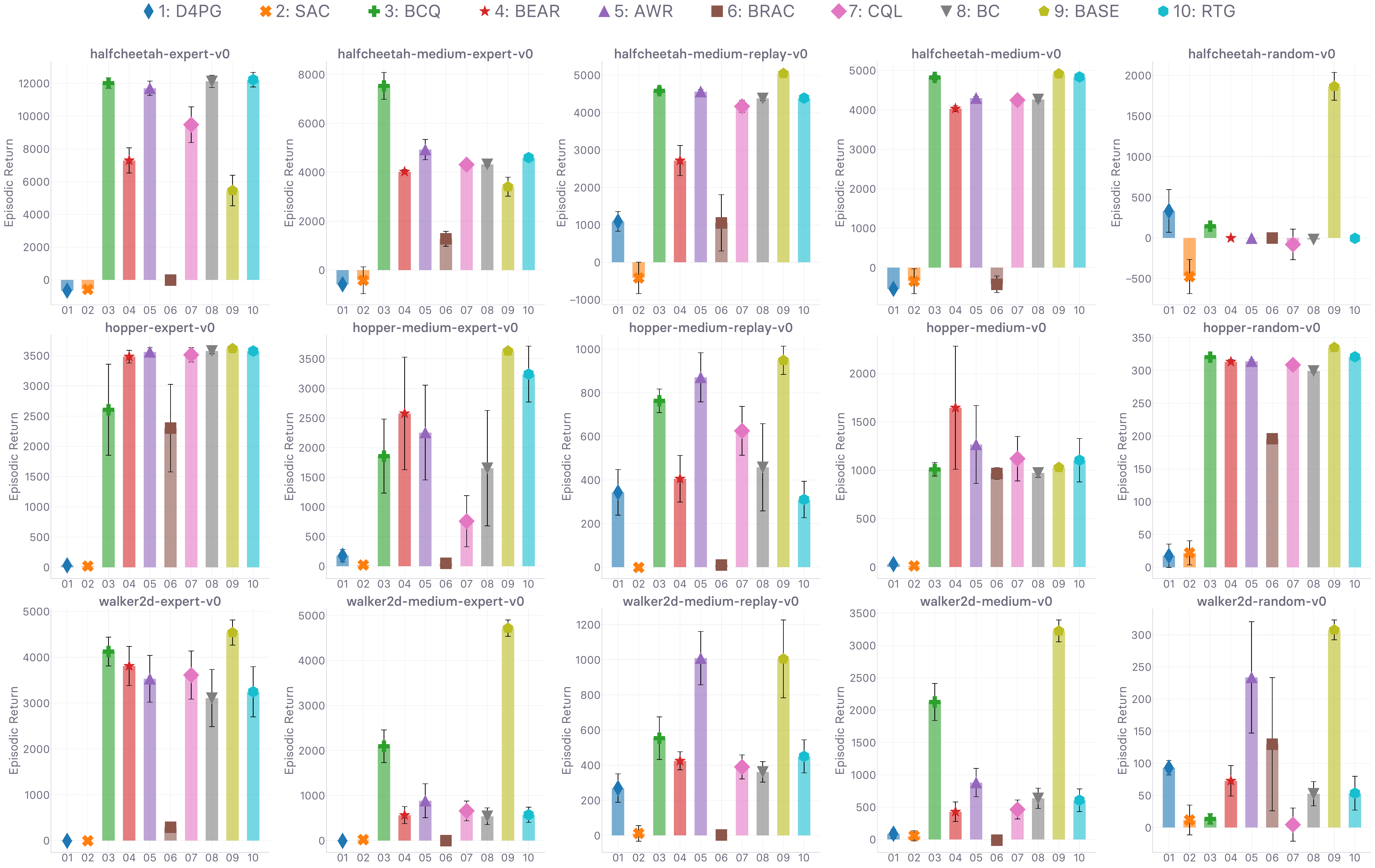}}
  \end{subfigure}
  \begin{subfigure}{\textwidth}
    \caption{Evolution of the return during training}
    \center\scalebox{0.18}[0.18]{\includegraphics{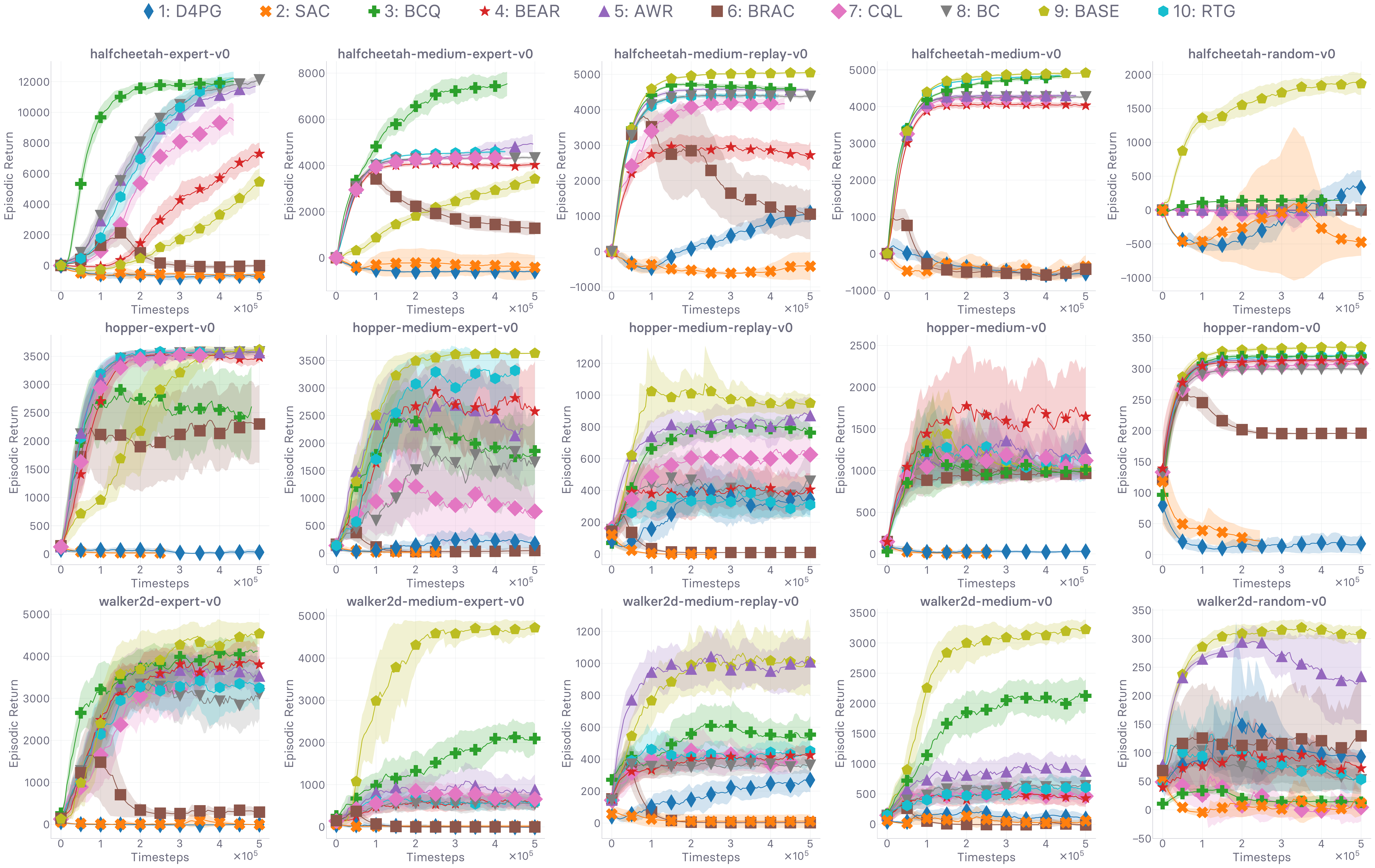}}
  \end{subfigure}
  \caption{Empirical evaluation of the return of RTG among the baselines
  treated in \textsc{Section}~\ref{baselines}.
  (a) The first three rows give the return mean and standard deviation on training completion.
  (b) The last three rows give the evolution of the return.
  Runtime is 12 hours. Best seen in color.}
  \label{rtgbaselines:barplot}
\end{figure}

We build RTG by adding both CQL's constraints in \textsc{Base}, as
add-on pieces to the loss optimized by $Q_\omega$ (\textit{cf.}~\cite{Kumar2020-zb}).
These are constraining $Q_\omega$ directly.
Formally, the loss optimized by CQL and RTG to learn $Q_\omega$ articulates as follows
(omitting numerical tricks; \textit{cf.}~\cite{Kumar2020-zb} for the various versions of CQL):
\begin{align}
\ell_\omega
\coloneqq
&\mathbb{E}_{s \sim \rho^\beta(\cdot), a \sim \beta(\cdot | s), s' \sim \rho^\beta(\cdot)}
\bigg[
\Big(
Q_\omega(s,a) -
\big(
r(s, a, s') + \gamma \, \mathbb{E}_{a' \sim \pi_\theta(\cdot | s')}
\big[
Q_{\omega'}(s',a')
\big]
\big)
\Big)^2
\bigg]
\label{cqltd}
\\
&+ \alpha
\Big(
\mathbb{E}_{s \sim \rho^\beta(\cdot), a \sim \text{unif}(\mathcal{A}[s])}
\big[
Q_\omega(s,a)
\big]
-
\mathbb{E}_{s \sim \rho^\beta(\cdot), a \sim \beta(\cdot | s)}
\big[
Q_\omega(s,a)
\big]
\Big)
\label{cqlcons}
\end{align}
The loss laid out in \textsc{eq}~\ref{cqltd} is the standard temporal-difference (or TD) error minimized
by CRR's critic $Q_\omega$ (as it appears in \textsc{Algorithm}~\ref{algobase}).
Brought over from CQL, the first piece of \textsc{eq}~\ref{cqlcons}
constitutes CQL's first constraint; it
tries to minimize the action value everywhere --- using uniformly sampled actions in $\mathcal{A}$
to apply the constraint onto.
The second piece of \textsc{eq}~\ref{cqlcons}
constitutes CQL's second constraint; it
tries to maximize the action-value over $\mathcal{D}$.
In effect, the loss laid out above
encourage the enlargement of the gap $\Delta_\textsc{gap}$ in \textsc{eq}~\ref{gap}.
Note, the gap  is signed.
RTG's name stems from this desideratum, by plugging CQL's constraints into CRR,
we urge the agent to deepen the gap in action-value between arbitrary actions and the ones from the offline dataset.
In effect, the aggregation of these two constraints \textit{increases} the learned $Q_\omega$ over $\mathcal{D}$,
and \textit{decreases} it everywhere else.
As such, the higher the scaling coefficient for these constraints, the more quantity optimality inductive bias
$\mathcal{B}$ we inject artificially into the CRR agent, while having a tight handle of how much we inject.
We see it as a minimal and parsimonious way to study the impact of such injection on the performance of CRR.
In \textsc{Figure}~\ref{basertg:barplot}, we show how RTG compares to CRR in terms of return.
Additionally, in \textsc{Figure}~\ref{basertggap:barplot}, we display the associated gaps $\Delta_\textsc{gap}$
(\textit{cf.} definition in \textsc{eq}~\ref{gap}).
\textsc{Figure}~\ref{rtgbaselines:barplot} puts things into perspective by depicting RTG's performance against the
baselines that we laid out in \textsc{Section}~\ref{baselines}.
We observe in \textsc{Figure}~\ref{basertg:barplot} that RTG improved upon CRR in the
3 top-left corner subplots, and displays significantly worse results in the 12 other subplots of the grid.
We arrive at the same expected conclusion: increasing a method's bias towards the optimality of $\beta$ is
a good idea if and only if $\beta$ is at least close to being optimal.
Naively forcing a method to imitate $\beta$ by injecting more of $\mathcal{B}$ is therefore not a decision to be
taken lightly and should be heavily grounded with respect to what we \emph{know} (and perhaps more importantly,
what we do \emph{not} know) about the contents of $\mathcal{D}$.
The greater bias is depicted clearly in \textsc{Figure}~\ref{basertggap:barplot} where we see that the gaps displayed by
RTG are consistently further away from zero than the ones in CRR.
Interestingly, there are two datasets (random, top-right and bottom-right)
in which the RTG gaps have high \emph{positive}
values, instead of low negative values like in the other subplots of
the grid in \textsc{Figure}~\ref{basertggap:barplot}.
Based on how the gap $\Delta_\textsc{gap}$ is defined (\textit{cf.}~\textsc{eq}~\ref{gap}),
this means that, according to the RTG agent,
actions inside the offline dataset have lower value than ones uniformly picked in $\mathcal{A}$.
This goes against the desideratum that motivated the introduction of the bias-inducing constraints,
and attests to the \emph{brittleness} of such biasing mechanism.
As expected, the RTG agent performs particularly badly in these two datasets (random, top-right and bottom-right
in \textsc{Figure}~\ref{basertg:barplot}),
in line with the nonsensical gaps displayed by RTG in these (\textit{cf.}~\textsc{Figure}~\ref{basertggap:barplot}).
Interestingly, \textsc{Figure}~\ref{rtgbaselines:barplot} shows that RTG beats CQL
in almost every tackled dataset.
Nevertheless, one should only use RTG when $\mathcal{C}$ is satisfied
--- otherwise, CRR.

\begin{figure}[H]
  \begin{subfigure}{\textwidth}
    \caption{\textsc{Base}}
    \center\scalebox{0.18}[0.18]{\includegraphics{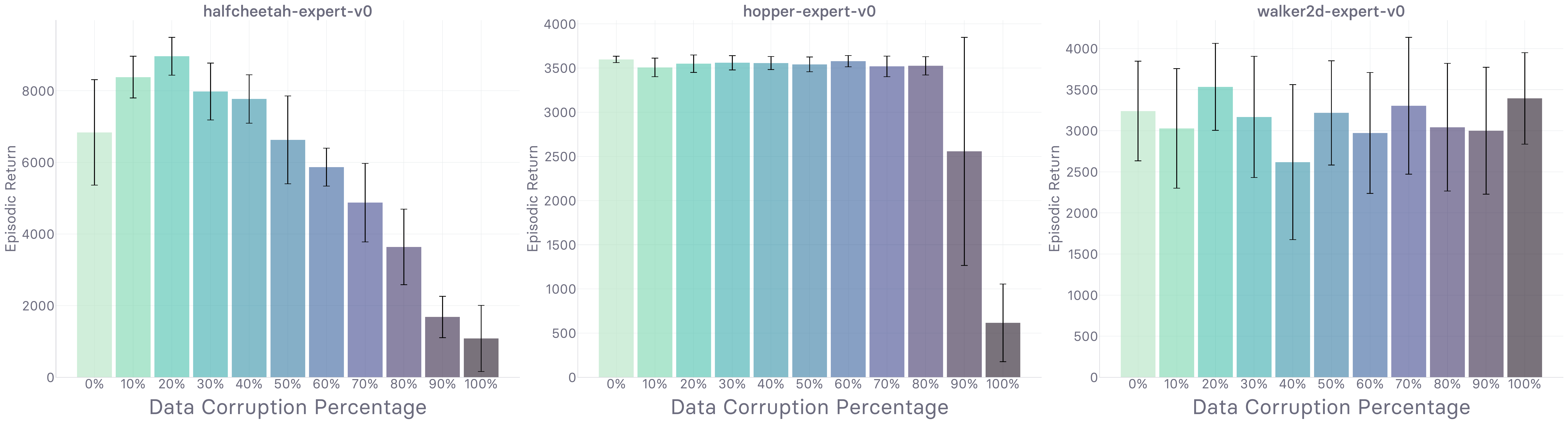}}
  \end{subfigure}
  \begin{subfigure}{\textwidth}
    \caption{RTG}
    \center\scalebox{0.18}[0.18]{\includegraphics{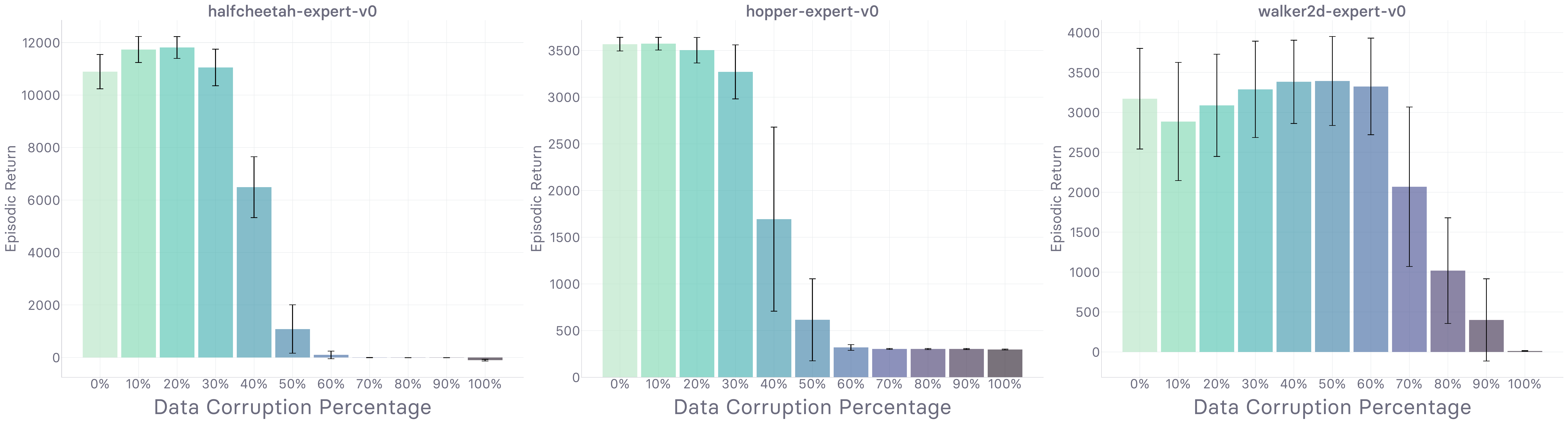}}
  \end{subfigure}
  \caption{Empirical evaluation of the return of  (a) \textsc{Base} and (b) RTG
  in \textbf{mixed datasets},
  in effect simulating a situation in which the expert-grade dataset suffers from
  \textbf{data corruption} with random data to various degrees
  (\textit{cf.} text for a complete description of the experimental design).
  We cover the range $p \in [0.0,1.0]$ with increments of $0.1$ (from 0\% to 100\% of corruption
  by increments of 10\%).
  Runtime is 12 hours. Best seen in color.}
  \label{basertgmixed:barplot}
\end{figure}

We have established through a series of experiments that RTG performs well on expert datasets and
poorly on random datasets, due to how much optimality bias $\mathcal{B}$ is injected into the agent.
\textsc{Base} does not inject as much bias, and thus
\textit{a)} behaves far better when $\beta$ is sub-optimal,
but \textit{b)} considerably lags behind RTG in expert datasets.
We would like to know how both methods perform in between these dataset qualities, \textit{i.e.}
what happens when the grade of data in the offline dataset $\mathcal{D}$ gradually decreases from
the maximum level (expert) to the minimum level (random).
To answer this, we investigate how both \textsc{Base} and RTG perform
in a series of \emph{mixed} datasets,
in effect simulating a situation in which the expert-grade dataset suffers from
\textbf{data corruption with random data} to various degrees.
These are crafted by aggregating a portion $p \in [0,1]$ of the \emph{expert} dataset
for a given environment with a portion $1-p$ of the \emph{random} dataset for the same environment.
In our experiments, $p$ covers the range $p \in [0.0,1.0]$ with increments of $0.1$.
Note, we \emph{shuffle} the datasets $A$ and $B$ using the agent's random seed,
before merging the portion $p$ extracted from dataset $A$ with the portion $1-p$ extracted from dataset $B$.
Since we average every reported run across a set of random seeds fixed beforehand
(\textit{cf.}~\textsc{Appendix}~\ref{evalprotocol}),
the results reported
in \textsc{Figure}~\ref{basertgmixed:barplot}
for this set of runs with mixed datasets are all the more robust
and reproducible.
As expected, RTG drops far quicker in performance than \textsc{Base} as we increase
the proportion of random data.
Yet, RTG still manages to accumulate a \textit{``fair''} return when the portion of random data in $\mathcal{D}$
is as high as $1-p = 0.4$ across the range of environments.
As such, the results of \textsc{Figure}~\ref{basertgmixed:barplot} show us once more that
\emph{knowing} about $\mathcal{D}$'s quality (\textit{i.e.} whether condition $\mathcal{C}$ is verified or not)
to then chose a method with the \emph{right} level of optimality inductive bias $\mathcal{B}$
is preferable over designing an algorithm than can \textit{``do it all''}.
Without this knowledge (\textit{i.e.} condition $\mathcal{C}$ is not verified),
then \textsc{Base} is the practitioner's best bet.
Besides, in practical scenarios where the data source can oftentimes be compromised and polluted with random data,
it is far easier for us to recommend the use of \textsc{Base} over RTG
(\textit{cf.}~\textsc{Figure}~\ref{basertgmixed:barplot}) --- or any other method with high bias.

\section{Towards revisiting generalized policy iteration in offline RL}
\label{gpirevisitationintro}

Since in practice the optimality condition $\mathcal{C}$ is never satisfied,
we set out to investigate how to improve the \textsc{Base} approach
by revisiting the
\emph{Generalized Policy Iteration} (GPI) learning procedure
(subsuming \textsc{Base}, as well as \emph{any} actor-critic touched on or investigated in this work)
in the \emph{offline} regime
(\textit{cf.}~\textsc{Figure}~\ref{gpidiag}).
In essence, methods that implement GPI alternate between a policy \emph{evaluation} step
(during which the value $Q_\omega$ is updated to be consistent \textit{w.r.t.}, or evaluate, the policy $\pi_\theta$)
and policy \emph{improvement} step
(during which the policy $\pi_\theta$ is updated to be greedy \textit{w.r.t.} its coupled value $Q_\omega$).
The loss optimized by \textsc{Base}'s critic $Q_\omega$ is laid out in \textsc{Algorithm}~\ref{algobase}
and in \textsc{eq}~\ref{cqltd}.
Learning the critic $Q_\omega$ offline,
characterized by the inability to acquire more data via interactions with the MDP,
exposes said critic to a distributional shift due to
out-of-distribution (or OOD) actions that can be involved in the Bellman target part of \textsc{eq}~\ref{cqltd}.
This phenomenon can manifest simply because any model likely evaluates arbitrarily poorly on
data located outside the distribution said model was trained on.
As such, since the Bellman target part of \textsc{eq}~\ref{cqltd}
involves an evaluation of the critic $Q_\omega$ on an action from the learned policy $\pi_\theta$ (in line with GPI)
then these evaluations might yield nonsensical values as soon as $\pi_\theta$'s are too far off
$\beta$'s predictions (\textit{i.e.} too far off the distribution underlying the dataset,
which \emph{is} the training distribution in the considered offline setting).
Most of the methods touched on when we laid out the related works in \textsc{Section}~\ref{relatedwork},
and studied in our first investigation in \textsc{Section}~\ref{baselines},
stave off OOD actions by forcing the learned policy $\pi_\theta$ be be close to $\beta$,
the distribution underlying the offline dataset $\mathcal{D}$.
Since, in GPI, $\pi_\theta$ is used to generate the action employed in the Bellman target,
updating $\pi_\theta$ in the vicinity of $\beta$ allows $Q_\omega$ not to suffer the instabilities
that would be caused by a distributional shift in target actions.
In line with the goal of GPI,
the alternation of policy evaluation and improvement must lead the estimated value and policy
to coincide with their optimal counterparts in the sense of Bellman's optimality,
\emph{while} being tied to $\beta$ for stability concerns
(as illustrated in \textsc{Figure}~\ref{gpidiag}).
Unless the offline distribution $\beta$ is optimal (\textit{i.e.} the optimal policy coincides with $\beta$),
the agent must face the following trade-off: \textbf{\emph{to what extent should one aim for optimality
at the expense of stability?}}

As such, we investigate unifying generalizations of the value objective
and policy objectives that consider how close to optimality the agent can get
without being exposed to the dreaded distributional shift that hinders the offline agent.
These policy evaluation and improvement studies
are carried out in \textsc{Sections}~\ref{pe} and \ref{pi} respectively.
These generalized evaluation and improvement objectives
can be aligned with the traditional actor-critic ones implementing GPI
in particular cases.
Our investigations involve the introduction of a wide spectrum of \textbf{\emph{proposal policies}}.
These proposal policies or distributions
act as placeholders or substitutes for a slew of different action distributions,
some \textbf{\emph{safer}} than others in terms of exposure to distributional shift due to OOD actions.
These investigations all take place
over \textsc{Base}.

We chose CRR
\textit{a)}
for the same reason we have done so in the investigation carried out in this section
(it is the method that seems to perform consistently well across the board,
as shown and concluded in \textsc{Section}~\ref{baselines}),
but also
\textit{b)}
because we have just shown in this section that simply injecting dataset-grounded optimality bias
in CRR (crystallized as RTG) enables the method to compete with top-performing baselines in the
three datasets CRR was falling behind.

\section{Generalizing policy evaluation: diagnosing the choice of bootstrap policy
to assess which proposal distribution should Q evaluate}
\label{pe}

In this section, we introduce a generalized objective for policy evaluation
whose optimization urges the learned action-value Q to evaluate a proposal policy.
We consider a range of distinct proposal distributions over actions, and study how
they impact the performance of the agent as they bootstrap the Q-value in our generalized
temporal-difference loss.
In short, they differ by how they tackle the trade-off between striving for optimality
and avoiding OOD actions.

\subsection{Unifying operators}
\label{operators}

Before listing out the proposal distributions $\zeta$'s considered in this work,
we first define homomorphic functional operators over the space of
functions mapping states from $\mathcal{S}$
to (state-conditioned) probability densities over actions from $\mathcal{A}$.
These operators --- denoted by $\mathcal{T}_\textsc{Eval}$ and
$\mathcal{T}_\textsc{Max}^{\omega, m}$ --- transform
stochastic policies from $\mathcal{P}(\mathcal{A})^\mathcal{S}$
into stochastic policies from $\mathcal{P}(\mathcal{A})^\mathcal{S}$,
and differ by how the sampling unit of the former is used to build the samples of the latter:
$(\forall \pi \in \mathcal{P}(\mathcal{A})^\mathcal{S})$,
sampling from the $s$-conditioned policy $\mathcal{T}_\textsc{Max}^{\omega, m}[\pi](\cdot | s)$
corresponds to sampling $m$ actions from the policy $\pi$ at $s$ and picking the action $a$ among the $m$ sampled ones
that has the highest estimated action-value at $s$, $Q_\omega(s,a)$.
Formally, $\mathcal{T}_\textsc{Max}^{\omega, m}[\pi] \in \mathcal{P}(\mathcal{A})^\mathcal{S}$
is defined to satisfy the following equivalence:
\begin{align}
  \big(\forall \pi \in \mathcal{P}(\mathcal{A})^\mathcal{S}\big)
  (\forall s \in \mathcal{S})
  \qquad
  a \sim \mathcal{T}_\textsc{Max}^{\omega, m}[\pi](\cdot | s)
  \iff
  a = \argmax_{a^i} \big\{ Q_\omega(s, a^i) \, | \, a^i \sim \pi(\cdot | s)\big\}_{i \in [1,m] \cap \mathbb{N}}
  \label{tmaxop}
\end{align}
For conceptual symmetry with $\mathcal{T}_\textsc{Max}^{\omega, m}$, we similarly introduce
$\mathcal{T}_\textsc{Eval}$, defined as the identity
homomorphic operator from and to the space of stochastic policies from $\mathcal{S}$ to $\mathcal{A}$.
Trivially, $(\forall \pi \in \mathcal{P}(\mathcal{A})^\mathcal{S})$,
sampling from the $s$-conditioned policy $\mathcal{T}_\textsc{Eval}[\pi](\cdot | s)$
corresponds to sampling a single action $a$ from the policy $\pi$ at $s$ and picking this action.
Maintaining the symmetry in notations, $\mathcal{T}_\textsc{Eval}[\pi] \in \mathcal{P}(\mathcal{A})^\mathcal{S}$
is formally defined to satisfy the following equivalence:
\begin{align}
  \big(\forall \pi \in \mathcal{P}(\mathcal{A})^\mathcal{S}\big)
  (\forall s \in \mathcal{S})
  \qquad
  a \sim \mathcal{T}_\textsc{Eval}[\pi](\cdot | s)
  \iff
  a \sim \pi(\cdot | s)
\end{align}
We will leverage these operators as building blocks to assemble
proposal policies with the maximum amount of notational overlap to keep our notation's verbosity to a bare minimum.
We notably use these operators to craft conditional operators
(we report them in \textsc{Appendix}~\ref{spioperators}),
able to adapt their output depending on the value of a given condition,
that we will later use to design learning objectives
reminiscent of Safe Policy Improvement (SPI) updates \cite{Petrik2016-yc}.

\subsection{Offline dataset distribution clones}
\label{offlineclones}
Finally, we introduce the policies $\beta_\textsc{c}$ and $\beta_\textsc{c}^\xi$, the last prerequisites
before laying out the proposal policies we considered for policy evaluation and improvement.
$\beta_\textsc{c}$ is a clone of $\beta$, the policy underlying
the offline dataset $\mathcal{D}$. Concretely, the $\beta_\textsc{c}$ policy
is modeled via a state-conditional variational auto-encoder (VAE) \cite{Kingma2014-hf, Rezende2014-ef}
trained to reconstruct the state-action pairing displayed in $\mathcal{D}$,
effectively cloning $\beta$ via behavioral cloning (BC),
making it a policy one can sample from --- given a state $s$ --- at training and evaluation time.
While $\beta_\textsc{c}$ stochastically generates actions from given states,
$\xi$ maps state-action pairs to actions,
and should therefore be interpreted as a \textit{state-conditional action perturbation} rather than as
a policy. Leveraging the perturbation model $\xi$,
we introduce $\beta_\textsc{c}^\xi$ to satisfy the following equivalence:
\begin{align}
  (\forall s \in \mathcal{S})
  \qquad
  a \sim \beta_\textsc{c}^\xi(\cdot | s)
  &\iff
  a = a_{\beta_\textsc{c}} \: + \: \Phi \; a_\xi \\
  &\text{with} \quad
  a_{\beta_\textsc{c}} \sim \beta_\textsc{c}(\cdot | s)
  \;\, \text{and} \;\,
  a_\xi = \xi(s, a_{\beta_\textsc{c}})
  \label{xieq}
\end{align}
Such a policy (perturbed clone $\beta_\textsc{c}^\xi$) was first introduced in BCQ \cite{Fujimoto2018-mj},
where the authors suggest the relative action scaling value of $\Phi = 0.05$, which we adopt in this work.
As in \cite{Fujimoto2018-mj},
we update the state-conditional action perturbation of the action predicted by the probabilistic clone
to maximize $Q_\omega(s, \bar{a})$, with $\bar{a} \sim \beta_\textsc{c}^\xi$
by leveraging the deterministic policy gradient theorem \cite{Silver2014-dk}.
Note, since the action sampled from the $\beta$-clone $\beta_\textsc{c}$
is an \textit{input} to the perturbation model $\xi$,
and that this is the only source of stochasticity in the $\beta_\textsc{c}^\xi$ policy,
the optimization of $\xi$ does not involve any reparametrization trick
--- in contrast with the optimization of $\beta_\textsc{c}$ which does
(\textit{cf.}~\cite{Kingma2014-hf, Rezende2014-ef}).

\subsection{Proposal policies and value simplex}
\label{proposalsimplex}
\begin{figure}[!h]
  \center\scalebox{0.3}[0.3]{\includegraphics{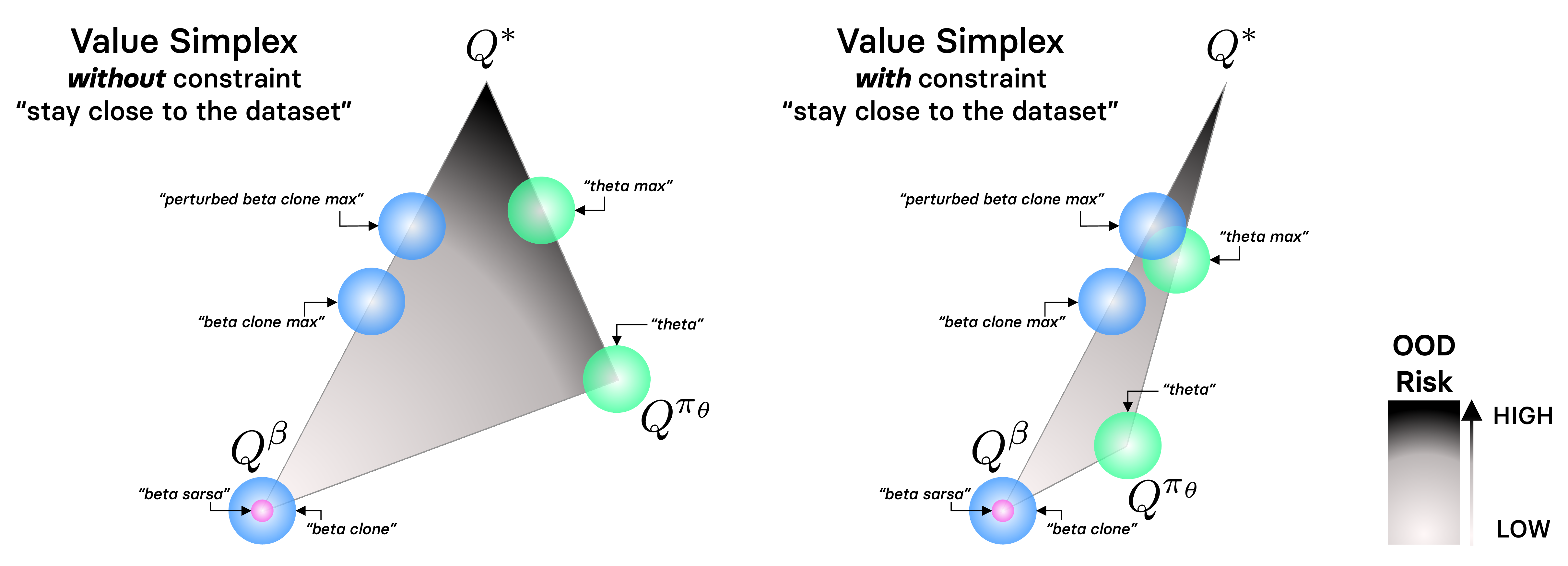}}
  \caption{Abstract representation of the relative positioning of the action-values learned
  using the various proposal distributions laid out in \textsc{Section}~\ref{pe}
  to generate the actions used to bootstrap Bellman's equation.
  These Q-values are depicted by disks over the simplex spanned by the optimal value $Q^*$, the value function
  perfectly evaluating the learned actor $\pi_\theta$, $Q^{\pi_\theta}$, and the exact value function $Q^\beta$
  associated with the policy underlying the offline dataset, $\beta$.
  The disk diameter roughly depicts how confident one can be about the placement of the various
  values associated with the tackled proposal policies (\textit{cf.}~\textsc{Section}~\ref{pe}) on the abstract simplex.
  Albeit only crudely estimating the actual geometry of the action-value simplex,
  this diagram can nevertheless help us categorize the
  different proposal distributions with respect to how they expose to agent and its value to
  \emph{out-of-distribution} (OOD) actions \emph{at training time}.
  Best seen in color:
  \textit{pink} signifies that the proposal distribution is $\beta$ --- requiring access to
  \textit{SARSA}-formatted transitions from the dataset, \textit{blue} that the proposal distribution relies on an
  estimate of the $\beta$ distribution, and \textit{green} that only the actor $\pi_\theta$ is used
  to bootstrap.}
  \label{valuesimplex}
\end{figure}

We now lay out the proposal policies $\zeta$ considered in this work.
In the context of policy evaluation (the focus of this section),
the proposal policy $\zeta$ is a placeholder for
the state-conditioned distribution from which the \textit{next} action $a'$ is sampled
to bootstrap $Q_\omega$ with at the \textit{next} state $s'$.
Formally, the proposal policy $\zeta$ satisfies the following schema:
\begin{align}
  a' \sim \zeta(\cdot | s')
\end{align}
and the form of Bellman's equation considered in this work is the one where the target policy is not
the optimal policy like in Q-learning \cite{Watkins1989-ir, Watkins1992-gl},
but the proposal policy $\zeta$. In other words, we involve the variant of Bellman's equation
that urges $Q_\omega$ to \textit{evaluate} the proposal policy $\zeta$,
\textit{i.e.} that makes $Q_\omega$ consistent with $\zeta$.
Consequently, employing such a recursive equation to design the temporal difference update rule
--- with which $Q_\omega$ is updated via stochastic gradient descent ---
will in effect make $Q_\omega$ approximate $Q^\zeta$, hence $Q_\omega \approx Q^\zeta$.
For completeness,
the loss used to update the action-value's parameter vector $\omega$ is the following:
\begin{align}
  \ell_\omega \coloneqq
  \mathbb{E}_{s \sim \rho^\beta(\cdot), a \sim \beta(\cdot | s), s' \sim \rho^\beta(\cdot)}
  \bigg[
  \Big(
  Q_\omega(s,a) -
  \big(
  r(s, a, s') + \gamma \, \mathbb{E}_{a' \sim \zeta(\cdot | s')}
  \big[
  Q_{\omega'}(s',a')
  \big]
  \big)
  \Big)^2
  \bigg]
  \label{criticloss}
\end{align}
Nevertheless, since the quadruple $(s, a, r, s')$ (\textit{abbrv.}~``\textit{SARS} transition'')
is always coming from the offline dataset $\mathcal{D}$
assumed to have been generated by the interactions of a behavior policy or baseline $\beta$,
and is therefore distributed as such, we can only have $Q_\omega \approx Q^\zeta$
if (and only if) the offline dataset $\mathcal{D}$ was generated by
an artificial agent following the policy $\zeta$ when interacting with the world $\mathcal{E}$.
In other words, we can write the \emph{intuitive} equivalence:
\begin{align}
  Q_\omega \approx Q^\zeta
  \iff
  \zeta \approx \beta
  \label{omegazetaequiv}
\end{align}
As such, the closer to $\beta$ we model and train $\zeta$ to be,
the more we can expect the learned action-value function
approximator $Q_\omega$ to accurately \textit{evaluate} the proposal distribution $\zeta$,
in which case $Q_\omega$ is also a good surrogate for $Q^\beta$.
This scenario corresponds to the virtual absence of \emph{distributional shift},
since there is little discrepancy between the distribution underlying the dataset, $\beta$, and the proposal
policy $\zeta$ used to generate the actions $Q_\omega$ must evaluate.
In the specific case where $\zeta$ coincides exactly with the offline behavior policy $\beta$ generating the
offline data, \textit{i.e.}~$\zeta = \beta$,
the critic loss $\ell_\omega$ laid out in \textsc{eq}~\ref{criticloss}
is in effect equivalent to a SARSA update \cite{Rummery1994-qp, Thrun1995-sz, Sutton1996-ky, Van_Seijen2009-yw},
which is an on-policy (and therefore effectively \textit{online}) update
for the learned action-value $Q_\omega$
--- which, as we have previously established in \textsc{eq}~\ref{omegazetaequiv},
then approximates $Q^\zeta$.
Indeed, in that scenario, we would have the states $s$ and $s'$ distributed as $\rho^\zeta$,
and the actions $a$ and $a'$ distributed as $\zeta$,
making the behavior and target policies coincide in an on-policy fashion, as follows:
\begin{align}
  \ell^\text{SARSA}_\omega \coloneqq
  \mathbb{E}_{(s,s') \sim \rho^\zeta, (a,a') \sim \zeta}
  \bigg[
  \Big(
  Q_\omega(s,a) -
  \big(
  r(s, a, s') + \gamma \,
  Q_{\omega'}(s',a')
  \big)
  \Big)^2
  \bigg]
  \label{sarsaloss}
\end{align}
In this scenario, since $\zeta = \beta$, we can equivalently write the exact same expression for
$\ell^\text{SARSA}_\omega$ with $\beta$ instead of $\zeta$.
A natural first candidate for our proposal policy $\zeta$ is therefore $\beta$ (exactly, not an approximation),
which can be achieved by leveraging the availability of the \textit{next} action for each
\textit{SARS} transition in the offline dataset $\mathcal{D}$.
In the context of this specific proposal policy strategy, which we name \textit{``beta sarsa''},
we therefore in effect use \textit{SARSA} transitions from $\mathcal{D}$.
Despite the setting laid out in \textsc{Section}~\ref{bg},
we here make an exception and benefit from the extra sequential information about $\beta$ provided by
these next actions attached to each transition.
Importantly, none of the other proposal policy strategies use any privileged information of this kind,
and stick to using \textit{SARS} transitions to learn $Q_\omega$.
Thus, in the \textit{``beta sarsa''} strategy,
the proposal policy is $\beta$, and
the next action $a'$ is coming directly from the transition sampled from the dataset $\mathcal{D}$.
Using the operators we have introduced at the beginning of \textsc{Section}~\ref{pe}, we can write:
\begin{align}
  \zeta \coloneqq \beta = \mathcal{T}_\textsc{Eval}\big[\beta\big]
  \quad \implies \quad
  a' &\sim \mathcal{T}_\textsc{Eval}\big[\beta\big](\cdot | s')
  \\ \tag*{\textsc{``beta sarsa''}}
  \label{betasarsa}
\end{align}
When the offline dataset $\mathcal{D}$ only contains \textit{SARS} transitions,
we can still, albeit to a lesser extent, leverage $\beta$'s by-design protection
against distributional shift caused by out-of-distribution \textit{next} actions in $Q_\omega$
by using a learned clone $\beta_\textsc{c}$ of $\beta$,
which we introduced in \textsc{Section}~\ref{pe}.
We name the strategy employing $\beta_\textsc{c}$ as proposal policy \textit{``beta clone''}.
Note, as a side-effect, we can expect this new approach to reduce the exposure of $Q_\omega$ to overfitting,
compared to adopting the parameter-free approach of simply using the available \textit{SARSA} transitions
(especially if the offline dataset coverage is poor).
We can therefore expect $Q_\omega$ to generalize better when using the proposal $\beta_\textsc{c}$ than $\beta$,
making it less likely to inject out-of-distribution actions in $Q_\omega$
--- unless the dataset covers $\mathcal{S} \times \mathcal{A}$ well,
in which case both strategies are equally capable.
Avoiding action-value overfitting is especially critical in actor-critic methods
since the actor $\pi_\theta$, trained to be greedy with respect to $Q_\omega$, tends to overfit itself
on spurious maxima of the action-value.
Overcoming this compounding effect from critic to actor is as crucial during training
--- provided $\pi_\theta$ is used in the proposal policy design ---
as it is crucial at evaluation time, in the case of \emph{on-policy} evaluation
(\textit{cf.}~\textsc{Section}~\ref{experimentalsetting} for a description of our experimental setting
and evaluation methods adopted in this work).
\begin{align}
  \zeta \coloneqq \mathcal{T}_\textsc{Eval}\big[\beta_\textsc{c}\big]
  \quad \implies \quad
  a' &\sim \mathcal{T}_\textsc{Eval}\big[\beta_\textsc{c}\big](\cdot | s')
  \\ \tag*{\textsc{``beta clone''}}
  \label{betaclone}
\end{align}
By using either $\zeta = \beta$ or $\zeta = \beta_\textsc{c} \approx \beta$
to produce $a'$ in \textsc{eq}~\ref{criticloss},
the equivalence of \textsc{eq}~\ref{omegazetaequiv} yields
$Q_\omega \approx Q^\beta$ in both cases.
We illustrate this, albeit through an abstract lens, in the diagrams of \textsc{Figure}~\ref{valuesimplex},
where the values learned by the strategies \ref{betasarsa} and \ref{betaclone} are depicted by
concentric disks centered at $Q^\beta$, signifying that both are approximating this value in functional space
--- the greater diameter for \ref{betaclone} echoes the wider trust region of the $Q_\omega$ approximation,
due to $\beta_\textsc{c}$ being itself an estimate of $\beta$.
As such, using proposal policies that cause $Q_\omega$ to be near $Q^\beta$ on the value simplex
depicted in \textsc{Figure}~\ref{valuesimplex} ($\zeta = \beta$ or $\zeta = \beta_\textsc{c}$)
ensures $Q_\omega$ will not be evaluated at out-of-distribution actions.
These proposal strategies are therefore \textit{safe} with regards to distributional shift in $Q_\omega$.

The offline RL algorithm we set out to use as baseline from \textsc{Section}~\ref{inductivebiases} onwards
uses the actor's policy $\pi_\theta$ as proposal policy.
We name this strategy \textit{``theta''},
add $Q^{\pi_\theta}$ as a corner of the action-value simplex in \textsc{Figure}~\ref{valuesimplex},
and can similarly write:
\begin{align}
  \zeta \coloneqq \mathcal{T}_\textsc{Eval}\big[\pi_\theta\big]
  \quad \implies \quad
  a' &\sim \mathcal{T}_\textsc{Eval}\big[\pi_\theta\big](\cdot | s')
  \\ \tag*{\textsc{``theta''}}
  \label{theta}
\end{align}
Learning $Q_\omega$ with the loss $\ell^\text{SARSA}_\omega$ and $\zeta = \pi_\theta$
would yield $Q_\omega \approx Q^{\pi_\theta}$,
as is commonplace in the online RL setting,
where the \textit{SARS} transitions
originate either from $\pi_\theta$ (online, and on-policy),
or an evolving mixture of previous iterates of $\pi_\theta$ (online, and off-policy with experience replay).
By contrast, in \textit{offline} RL,
the \textit{SARS} transitions are generated by the offline behavior policy $\beta$
that has no ties with $\pi_\theta$, and the dataset $\mathcal{D}$ produced by $\beta$ remains
frozen throughout the entirety of the learning process.
Maintaining $\pi_\theta$ in the vicinity of $\beta$ in some metric ---
as enforced in every single successful offline RL method reported in \textsc{Section}~\ref{baselines} ---
makes $\ell^\text{SARSA}_\omega$ using $\zeta = \beta$ coincide in analytical form with
$\ell_\omega$ using $\zeta = \pi_\theta$.
In other words, optimizing the offline loss $\ell_\omega$
with $\pi_\theta$ as proposal policy $\zeta$ while constraining $\pi_\theta$
(and therefore by construction $\zeta$) to be close to $\beta$ (\textit{i.e.}~$\zeta \approx \beta$),
we obtain, via the intuitive equivalence in \textsc{eq}~\ref{omegazetaequiv},
$Q_\omega \approx Q^{\pi_\theta}$.
Moreover, we have $Q_\omega \approx Q^\beta$ by transitivity, since $\pi_\theta$ is kept close to $\beta$
according to some metric which has an effect on the \textit{``closeness''} encoded here by the symbol
\textit{``$\approx$''} used as operator between action-value functions.
We illustrate the effect of encouraging $\pi_\theta$ to be close to $\beta$ in \textsc{Figure}~\ref{valuesimplex}
by representing the values $Q^{\pi_\theta}$ and $Q^\beta$
(corresponding to the values learned using $\ell^\text{SARSA}_\omega$ with
$\zeta = \pi_\theta$ and $\zeta = \beta$ or $\beta_\textsc{c}$ respectively) closer to each other.
As they get closer, the action-value simplex shrinks along the edge linking $Q^{\pi_\theta}$ to $Q^\beta$.
Before shifting our attention to the third corner of the simplex depicted in \textsc{Figure}~\ref{valuesimplex},
the optimal action-value $Q^*$, note how forcing $\pi_\theta$ to be somewhat close to $\beta$
to shield $Q_\omega$ from being evaluated at out-of-distribution actions (\textit{black} gradient
on the simplex of \textsc{Figure}~\ref{valuesimplex}) can have the \emph{averse} effect
of preventing $Q_\omega$ from ever reaching said optimal value $Q^*$.
This undesirable consequence is depicted both in
\textsc{Figure}~\ref{gpidiag} and \textsc{Figure}~\ref{valuesimplex}.

The diagrams of \textsc{Figure}~\ref{gpidiag} reminds us that the ultimate objective of
generalized policy iteration (GPI) \cite{Sutton1998-ow}
is for $Q_\omega$ to converge to the \emph{optimal} action-value $Q^*$,
represented as a corner of the simplex in \textsc{Figure}~\ref{valuesimplex}.
The canonical loss $\ell^*$ one uses to learn $Q^*$
is derived from the optimal version of Bellman's equation
--- the one used in Q-learning \cite{Watkins1989-ir, Watkins1992-gl},
and is defined as follows:
\begin{align}
  \ell^*_\omega \coloneqq
  \mathbb{E}_{s \sim \rho^\beta(\cdot), a \sim \beta(\cdot | s), s' \sim \rho^\beta(\cdot)}
  \bigg[
  \Big(
  Q_\omega(s,a) -
  \big(
  r(s, a, s') + \gamma \,
  \max_{a'}Q_{\omega'}(s',a')
  \big)
  \Big)^2
  \bigg]
  \label{qlearningloss}
\end{align}
By opting for an action-value $Q_\omega$ learned via Q-learning with $\ell^*_\omega$,
instead of via SARSA updates with $\ell^\text{SARSA}_\omega$
--- an adoption studied first in \cite{Crites1995-hn},
then in \cite{Lim2018-ey}
where the result of such an adoption was named an ``actor-\textit{expert}'' algorithm ---
we align the signal returned by $Q_\omega$ with the identification of whether a given action $a$ is
the \emph{best} action $a^*=\pi^*(s)$,
rather than evaluating the proposal policy $\zeta$ used in
\textsc{eq}~\ref{sarsaloss}.
As a result, due to the intertwined roles of the actor and critic
(even in proposal policy strategies where $\pi_\theta$ is not involved in $Q_\omega$'s update,
$Q_\omega$ is \emph{always} used in $\pi_\theta$'s update),
learning $Q_\omega$ via Q-learning will have a direct impact on $\pi_\theta$,
whose parameters will be updated to assign higher densities to actions that
$Q_\omega$, now estimating $Q^*$, believes are optimal.
Provided the estimation $Q_\omega \approx Q^*$ is viable,
this method has the clear advantage of
\emph{compartmentalizing} (containing and detaching from each other)
$Q_\omega$ and $\pi_\theta$,
therefore preventing the compounding of errors
(due to distributional shift and out-of-distribution actions,
inherent to offline RL)
in the alternating learning scheme
between policy and value that characterizes GPI \cite{Sutton1998-ow}.
We could also write the last operand of \textsc{eq}~\ref{qlearningloss} as
$\gamma \, Q_{\omega'}(s',\argmax_{a'}Q_{\omega'}(s',a'))$,
which has the added benefit of reminding us that the optimal policy $\pi^*$ greedy with respect to $Q^*$
is deterministic, since $\ell^*_\omega$ coincides with $\ell_\omega$ where $\zeta = \pi^*$.
While the $\argmax$ operation is tractable in a reasonable compute time when the actions are discrete
with a low number of dimensions, it is not a viable option \textit{as is}
when there is a plethora of discrete actions, or for continuous action spaces.
As such, the loss $\ell^*_\omega$ is not always the best candidate to learn
an estimate of the optimal value $Q^*$,
and we reported the slew of works that designed alternatives to the raw $\argmax$ operation
in these unviable scenarios in \textsc{Section}~\ref{relatedwork}.
We here opt for a simple stochastic sample-based \emph{relaxation}, leveraging the operator
$\mathcal{T}_\textsc{Max}^{\omega, m}$
introduced earlier in \textsc{eq}~\ref{tmaxop}:
\begin{align}
\argmax_{a'}Q_{\omega'}(s',a') \approx
\argmax_{a^i} \big\{ Q_{\omega'}(s', a^i) \, | \, a^i \sim \pi(\cdot | s')\big\}_{i \in [1,m] \cap \mathbb{N}}
= \mathcal{T}_\textsc{Max}^{\omega', m}[\pi](\cdot | s')
\label{relaxation}
\end{align}
where $\omega'$ is the parameter vector of the critic's target network introduced in \textsc{Section}~\ref{baselines},
and $\pi$ is a placeholder for a proposal distribution that the relaxation calls for, and for which we
consider the following candidates: $\beta_\textsc{c}$, $\beta_\textsc{c}^\xi$, and $\pi_\theta$.
The relaxation proposed in \textsc{eq}~\ref{relaxation} therefore
approximates the intractable loss
$\ell^*_\omega$ (which corresponds to $\ell_\omega$ where $\zeta = \pi^*$)
with the tractable loss $\ell_\omega$ where $\zeta = \mathcal{T}_\textsc{Max}^{\omega', m}[\pi]$,
with $\pi \in \{ \beta_\textsc{c},\beta_\textsc{c}^\xi,\pi_\theta \}$
(\textit{cf.}~\textsc{Section}~\ref{offlineclones} for the definitions of
$\beta_\textsc{c}$ and $\beta_\textsc{c}^\xi$, the proposal policies derived from the offline dataset policy $\beta$).
As such, in addition to the proposal policies $\zeta$ already introduced above,
we now also have the following ones, derived from
$\beta_\textsc{c}$, $\beta_\textsc{c}^\xi$, and $\pi_\theta$ respectively:
\begin{align}
  \zeta \coloneqq \mathcal{T}_\textsc{Max}^{\omega', m}\big[\beta_\textsc{c}\big]
  \quad \implies \quad
  a' &\sim \mathcal{T}_\textsc{Max}^{\omega', m}\big[\beta_\textsc{c}\big](\cdot | s')
  \\ \tag*{\textsc{``beta clone max''}}
  \label{betaclonemax} \\
  \zeta \coloneqq \mathcal{T}_\textsc{Max}^{\omega', m}\big[\beta_\textsc{c}^\xi\big]
  \quad \implies \quad
  a' &\sim \mathcal{T}_\textsc{Max}^{\omega', m}\big[\beta_\textsc{c}^\xi\big](\cdot | s')
  \\ \tag*{\textsc{``perturbed beta clone max''}}
  \label{perturbedbetaclonemax} \\
  \zeta \coloneqq \mathcal{T}_\textsc{Max}^{\omega', m}[\pi_\theta]
  \quad \implies \quad
  a' &\sim \mathcal{T}_\textsc{Max}^{\omega', m}[\pi_\theta](\cdot | s')
  \\ \tag*{\textsc{``theta max''}}
  \label{thetamax}
\end{align}
As for \ref{betasarsa}, \ref{betaclone}, and \ref{theta},
there is one colored disk depicted in \textsc{Figure}~\ref{valuesimplex} for
\ref{betaclonemax}, \ref{perturbedbetaclonemax}, and \ref{thetamax}.
The three latter are drawn closer to $Q^*$ than the three former.
In particular, \ref{perturbedbetaclonemax} is depicted closer to $Q^*$ than \ref{betaclonemax}
since $\beta_\textsc{c}^\xi$ is a clone $\beta_\textsc{c}$ perturbed slightly to maximize $Q_\omega$,
and is therefore pushing $Q_\omega$ further towards the optimal action-value $Q^*$.
Moreover, the value chosen for the hyper-parameter $m$ can be used to modulate where these disks are
located \textit{a)} on the segment joining $Q^\beta$ to $Q^*$ for
$\mathcal{T}_\textsc{Max}^{\omega', m}\big[\beta_\textsc{c}\big]$
and $\mathcal{T}_\textsc{Max}^{\omega', m}\big[\beta_\textsc{c}^\xi\big]$,
and \textit{b)} on the segment joining $Q^{\pi_\theta}$ to $Q^*$ for
$\mathcal{T}_\textsc{Max}^{\omega', m}[\pi_\theta]$.
Indeed, the proposal policies
$\mathcal{T}_\textsc{Max}^{\omega', m}\big[\beta_\textsc{c}\big]$,
$\mathcal{T}_\textsc{Max}^{\omega', m}\big[\beta_\textsc{c}^\xi\big]$, and
$\mathcal{T}_\textsc{Max}^{\omega', m}[\pi_\theta]$
are \textit{a priori} expected to become better approximations of
$\pi^*$ when $m$ is set to larger values.
The quality of such approximation nevertheless strongly depends on
the proposal distribution  $\pi$ introduced in \textsc{eq}~\ref{relaxation}
for the relaxation of $\ell^*_\omega$,
where $\pi \in \{ \beta_\textsc{c},\beta_\textsc{c}^\xi,\pi_\theta \}$.
Hence, as $m$ increases, we could draw the colored disks in \textsc{Figure}~\ref{valuesimplex}
--- associated with the proposal distributions
$\mathcal{T}_\textsc{Max}^{\omega', m}\big[\beta_\textsc{c}\big]$,
$\mathcal{T}_\textsc{Max}^{\omega', m}\big[\beta_\textsc{c}^\xi\big]$, and
$\mathcal{T}_\textsc{Max}^{\omega', m}[\pi_\theta]$ ---
increasingly closer to $Q^*$
(\textit{cf.}~\textsc{Appendix}~\ref{tmaxops} for a sweep over several values of the
hyper-parameter $m$, showing a trade-off between performance and
computational cost).

All in all, tackling an offline RL task is a \textbf{balancing act}: we want to move
$Q^{\pi_\theta}$ (the action-value consistent with the actor's policy $\pi_\theta$)
closer to $Q^*$, while not creating too much distance between $Q^{\pi_\theta}$ and $Q^\beta$
to avoid unforgiving distributional shifts during training.
What \textsc{Figure}~\ref{valuesimplex} does not depict,
in contrast with \textsc{Figure}~\ref{gpidiag},
is what the diagram would look like
for various qualities of dataset $\mathcal{D}$.
In particular, if the dataset contains near-optimal data (\textit{i.e.}~$\beta \approx \pi^*$)
the edge linking $Q^\beta$ to $Q^*$ would be considerably shorter,
such that $Q^\beta$ would almost overlap with $Q^*$.
In other words, the objective of offline RL is to \emph{shrink} this simplex
until a \emph{``sweet spot''} is reached.
When the dataset contains optimal data, we want
the simplex to shrink and collapse onto a single point ($Q^{\pi_\theta} = Q^\beta = Q^*$).
When the dataset contains sub-optimal data, we want
the simplex to reach a sweet spot that ought to manifest
\emph{before} all three corners collapse onto a single point,
since $\beta \neq \pi^*$.

To conclude, in this section, we have introduced several proposal policies $\zeta$ to sample
the next action $a'$ from, in $\ell_\omega$:
\ref{betasarsa}, \ref{betaclone}, \ref{theta},
\ref{betaclonemax}, \ref{perturbedbetaclonemax}, and \ref{thetamax}.
In an attempt to strike such balance, we also create proposal distributions which,
by leveraging the conditional operators introduced in \textsc{Appendix}~\ref{spioperators},
enable us to orchestrate value learning objectives that directly
echo the SPI update rule first proposed in \cite{Petrik2016-yc}.
We introduce these new SPI-inspired proposal distributions
(dubbed \ref{spibetaclone}, \ref{spibetaclonemax}, and \ref{spiperturbedbetaclonemax} by symmetry)
in \textsc{Appendix}~\ref{spiproposalsimplex}.

The pseudo-code of the algorithm we use to conduct the experiments reported in \textsc{Section}~\ref{peresults}
(that immediatly follows)
coincides with the one laid out in \textsc{Algorithm}~\ref{algobase},
\emph{except} for $\ell_\omega$ that we replace with the following critic loss:
\begin{align}
\ell_\omega \coloneqq
\mathbb{E}_{s \sim \rho^\beta(\cdot), a \sim \beta(\cdot | s), s' \sim \rho^\beta(\cdot)}
\bigg[
\Big(
Q_\omega(s,a) -
\big(
r(s, a, s') + \gamma \, \mathbb{E}_{a' \sim \textcolor{red}{\zeta}(\cdot | s')}
\big[
Q_{\omega'}(s',a')
\big]
\big)
\Big)^2
\bigg]
\end{align}
where the sole change from \textsc{Algorithm}~\ref{algobase} is colored in red.

We now report and discuss our experimental findings.

\subsection{Experimental results}
\label{peresults}
As before,
we rely on the experimental setting thoroughly described in \textsc{Appendix}~\ref{experimentalsetting}
to carry out the empirical investigation laid out here.

\begin{figure}[!h]
  \center\scalebox{0.18}[0.18]{\includegraphics{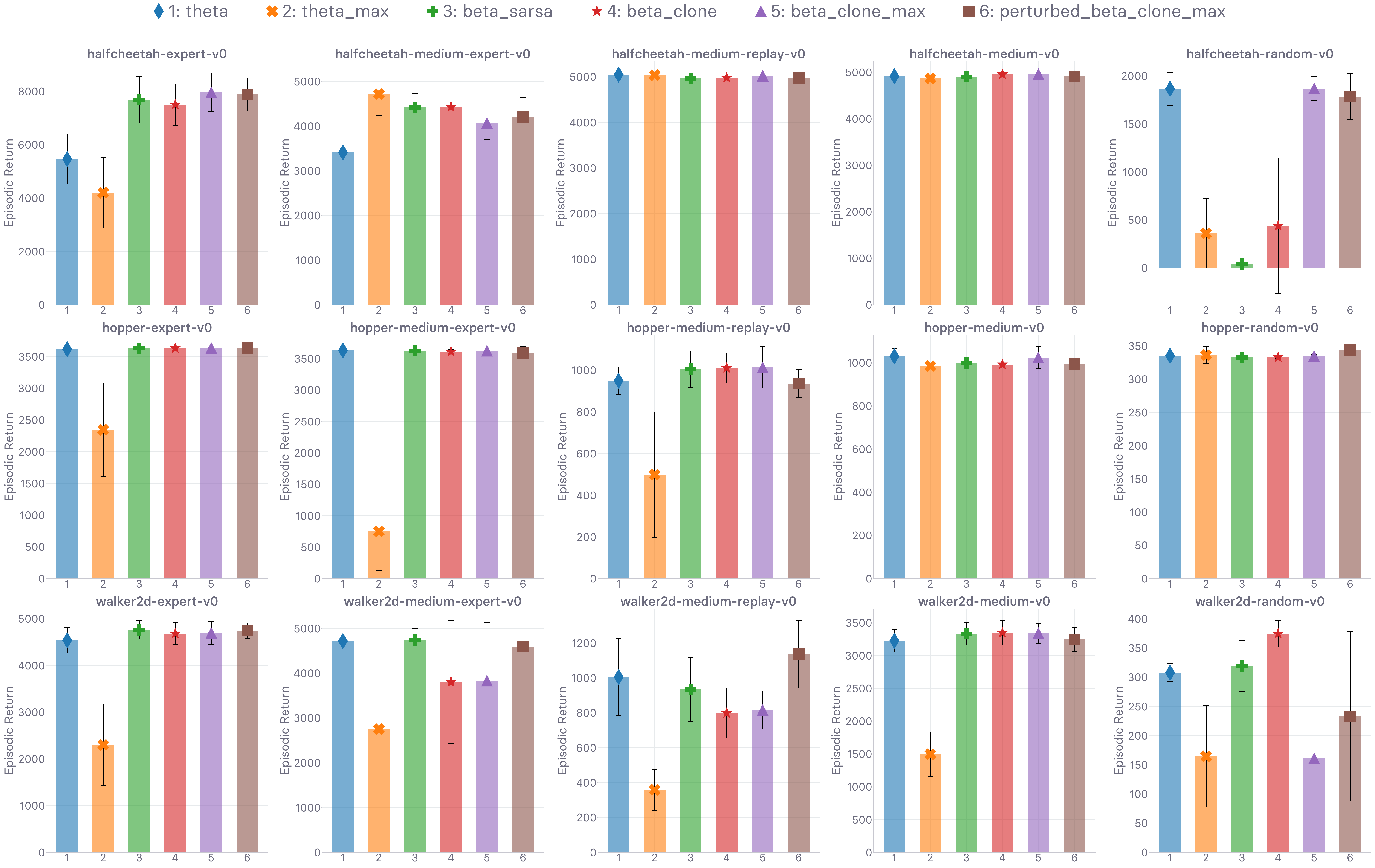}}
  \caption{Final performance of \textsc{Base} (\textit{cf.}~\textsc{Algorithm}~\ref{algobase})
  with the policy evaluation carried out
  under the different proposal distributions that we introduced in \textsc{Section}~\ref{proposalsimplex}.
  Everything except the proposal policy $\zeta$ used to sample the \textit{next} action from is identical.
  Runtime is 12 hours. Best seen in color.}
  \label{next:barplot}
\end{figure}

Notably, albeit perhaps naive in appearance, the results reported in
\textsc{Figure}~\ref{next:barplot}
show that proposal policies as simple as \ref{betasarsa} perform strikingly well
compared to considerably more sophisticated
in most environments and datasets
--- although one need access to \textit{SARSA}-formatted transition in the offline dataset $\mathcal{D}$
to be able to leverage this proposal distribution strategy.
Indeed, \ref{betasarsa} only performs worse than the baseline strategy \ref{theta} in a \emph{single}
environment (equivalently, in a single dataset), in the top-right sub-plot,
and does so considerably.
Nevertheless, in the 14 other environment-dataset couples, the empirical performance of the
\ref{betasarsa} proposal distribution strategy is on par with the baseline \ref{theta}, and even often outperforms
it, despite not involving the actor's policy $\pi_\theta$
in its policy evaluation update of $Q_\omega$ via $\ell_\omega$.

On the opposite side of the spectrum in \textsc{Figure}~\ref{next:barplot},
the proposal distribution \ref{thetamax} shows the poorest results in almost every datasets,
which is \textit{in fine} not too surprising considering the higher chances of
out-of-distribution actions in the Bellman backup it incurs.
As we laid out earlier in \textsc{Section}~\ref{projectionoptions},
involving a maximization operator over estimated action-values $Q_\omega(s,a)$ (for a given pair
$(s,a) \in \mathcal{S} \times \mathcal{A}$)
will have the harmful effect of compounding onto the overestimation bias $Q_\omega$ is prone to
\cite{Thrun1993-or},
locking the actor's policy onto arbitrarily overestimated action values when bootstrapping $Q_\omega$
via Bellman's equation.
Such effect is exacerbated in the offline RL setting \cite{Levine2020-hz},
although this distributional shift is still present when further interactions are allowed,
in the online scenario \cite{Fu2019-kb}.
The visual aid depicted in \textsc{Figure}~\ref{valuesimplex} illustrates the faced challenge:
involving a maximization operation to move $Q_\omega$ closer to $Q^*$, while making sure the proposal policy
that generates the \textit{next} action to bootstrap with is not too far off being distributed as the offline policy
$\beta$, underlying the offline dataset $\mathcal{D}$.
Still, \ref{thetamax} can display high return in some isolated cases,
as $Q_\omega$ \emph{can} suffer from said overestimation bias, yet not predictably or controllably so,
and therefore not consistently across the benchmark.
As such, attempting to move $Q_\omega$ closer to $Q^*$ directly from $Q^{\pi_\theta}$
is objectively not the safest route for the offline RL practitioner to take.

The proposal distributions of the \textit{clone} group
(composed of \ref{betaclone}, \ref{betaclonemax}, and \ref{perturbedbetaclonemax})
all yield similar returns.
\textbf{None of the methods within the group outperforms the two other consistently across the benchmark,
although \ref{perturbedbetaclonemax} shows the best performance within the group overall.}
Interestingly, these methods only rarely beat the \ref{betasarsa} heuristic, which despite needing
\textit{SARSA}-formatted transitions, does not rely on an additional state-conditional generative model of
the offline dataset $\mathcal{D}$
($\beta_\textsc{c}$ or its perturbed version $\beta_\textsc{c}^\xi$)
effectively \textit{cloning} $\beta$.
Such a trade-off can, in practice, be addressed differently depending on how complex the data distribution $\beta$ is
(therefore more difficult to estimate accurately via behavioral cloning),
and how feasible it is for the practitioner to gather the dataset to be used offline such that the
collected transitions are sequentially ordered in \emph{connex} trajectories
(transitions stored in $\mathcal{D}$ can then be \textit{SARSA}-formatted).

Finally, in \textsc{Figure}~\ref{nextappendix:barplot},
we report the empirical performance of using the SPI proposal policies
introduced in \textsc{Section}~\ref{spiproposalsimplex} in $Q_\omega$,
and discuss these in \textsc{Appendix}~\ref{spiperesults}.
In short, the involvement of the highly-unstable \ref{thetamax} (as we have just reported above)
in every SPI proposal distribution makes them overall not worth their tuning cost.

\paragraph{Which proposal distribution should Q evaluate?}

All in all, what \textsc{Figure}~\ref{next:barplot} shows is that
there is not \textit{one} proposal distribution that dominates the others across the entire spectrum of
dataset qualities, across every environment.
Involving $\beta$ in the proposal distribution
that is used to generate the bootstrap action in the TD update
(be it by using it directly, provided \textit{next} actions are available through $\mathcal{D}$,
or by cloning it and using the clone instead)
should be the preferred route to train the action-value $Q_\omega$ in offline RL
\textbf{\emph{only} when the dataset is known to contain transition from an expert},
\textit{i.e.} a near-optimal policy for the task at hand.
Indeed, as soon as the quality of $\mathcal{D}$ lowers,
the performance across the \textit{clone} group is less consitent as a whole:
one or several proposal distributions of the group often show high variance across random seeds,
and it is not rare to witness a surprisingly low performance for one proposal compared to the other
distributions of the group.
In these non-expert scenarios, it seems far preferable to stick to \ref{theta}
and avoid the \textit{clone} group.
This phenomenon is likely due to the fact that $\beta$ (and therefore also a clone of $\beta$)
has low entropy for expert datasets and high entropy for rando datasets.
Evaluating $Q_\omega$ on out-of-distribution actions is less likely when the entropy is low than when it is high.
Therefore, a distributional shift is increasingly likely to occur as the grade of the offline dataset
$\mathcal{D}$ decreases
from expert to random when $Q_\omega$ is evaluated with a proposal policy from the clone group.
\textbf{We can therefore articulate the following guidelines for the choice of proposal distribution
in the policy evaluation objective $\ell_\omega$.
\textit{1)} If the dataset grade is \emph{known}, use \ref{perturbedbetaclonemax} for near-expert-grade datasets,
and \ref{theta} otherwise.
\textit{2)} If the dataset grade is \emph{unknown}, the agnostic best choice is to
use \ref{theta}.}

\paragraph{Unifying previous offline RL works under \emph{one} policy evaluation framework.}

Among the proposal policies that we have formalized in a unified framework and empirically evaluated
in the section,
some have a \emph{counterpart} in prior offline RL
algorithms introduced in recent years.
Note, we consider the policy evaluation step in isolation from the GPI cycle
it is embedded in (\textit{cf.}~\textsc{Figure}~\ref{gpidiag}).
The proposal distribution $\zeta \coloneqq \pi_\theta$ set in \ref{theta} coincides with the usual
\textit{SARSA} update \cite{Rummery1994-qp, Thrun1995-sz, Sutton1996-ky, Van_Seijen2009-yw}
(as opposed to Q-learning update \cite{Watkins1989-ir, Watkins1992-gl})
adopted by most actor-critic architectures whose action-value's loss is based on the \textit{evaluation}
version of Bellman's equation
\cite{Silver2014-dk, Lillicrap2016-xa, Barth-Maron2018-ot, Haarnoja2018-bm,
Kumar2019-rw, Wu2019-nl, Kumar2020-zb, Siegel2020-lo, Wang2020-sr, Nair2020-gd}.
By bootstrapping $Q_\omega$ with $a' \sim \mathcal{T}_\textsc{Max}^{\omega', m}[\pi_\theta](\cdot | s')$,
$\forall s' \in \mathcal{S}$,
the proposal strategy \ref{thetamax} sets out to \textit{emulate} a Q-learning update to urge the actor
$\pi_\theta$, by design greedy with respect to the critic $Q_\omega$, to increase the probability density
of actions that $Q^*$ views as optimal (as opposed to the usual $Q^{\pi_\theta}$ critic in
\textit{SARSA}-like off-policy online actor-critic architectures \cite{Lim2018-ey}).
Such desideratum has been sought after in a slew of works released concurrently,
among which \textit{Amortized} Q-learning (AQL) \cite{Van_de_Wiele2018-qs}
draws the closest resemblance, albeit being an \textit{online} off-policy method
(\textit{cf.}~\textsc{Section}~\ref{relatedwork} for a rundown of said concurrent works by differ by
how they \emph{relax} the intractable maximization operation over $\mathcal{A}$
in the Q-learning version of Bellman's equation).
Later, \ref{thetamax} has been used in the BEAR-QL method \cite{Kumar2019-rw}
in an \emph{offline} RL context.
By replacing $\pi_\theta$ in the latter by a clone $\beta_\textsc{c}$
of the distribution underlying the offline dataset $\mathcal{D}$
such that $a' \sim \mathcal{T}_\textsc{Max}^{\omega', m}[\beta_\textsc{c}](\cdot | s')$,
$\forall s' \in \mathcal{S}$, we obtain the \ref{betaclonemax} proposal strategy,
which one can find as a standalone contribution in the EMaQ method \cite{Ghasemipour2020-ro}.
Further, by replacing $\pi_\theta$ by a \textit{perturbed} clone $\beta_\textsc{c}^\xi$,
such that $a' \sim \mathcal{T}_\textsc{Max}^{\omega', m}[\beta_\textsc{c}^\xi](\cdot | s')$,
$\forall s' \in \mathcal{S}$, we obtain the \ref{perturbedbetaclonemax} proposal strategy,
which is part of the BCQ method \cite{Fujimoto2018-mj} (the perturbed clone $\beta_\textsc{c}^\xi$ is
effectively the actor in BCQ, such that $\pi_\theta \coloneqq \beta_\textsc{c}^\xi$
at evaluation time).
Finally, BRPO \cite{Sohn2020-ay} can be cast as an instance of \ref{spibetaclone},
a safe policy improvement proposal distribution that we introduced,
among others, in \textsc{Appendix}~\ref{spiproposalsimplex}.

\paragraph{Complementary investigations \#1: the impact of Baird's advantage-learning.}

So as to complement our analysis on how to better carry out policy evaluation in offline RL,
we conduct two additional sets of experiments.
First, we investigate the effect of Baird's advantage-learning \cite{Baird1993-qa, Baird1999-dq}
as re-adapted to modern objective designs in \cite{Bellemare2015-ii}.
The purpose of Baird's advantage-learning is to \emph{increase} the gap in action-value
between optimal and sub-optimal actions, so that the greedy actor is less likely to select sub-optimal action
because of misestimation or simply numerical precision.
Notably, in offline RL, \cite{Wu2019-nl} and \cite{Kumar2020-zb}
undertook to increase the gap between actions that are close to
being distributed as the offline distribution $\beta$ and actions that seem not to be.
Despite being motivated by different desiderata --- avoiding sub-optimal action for Baird's advantage-learning,
\textit{i.e.}~$a \nsim \pi^*(\cdot | s)$;
avoiding out-of-distribution actions for works like \cite{Wu2019-nl} and \cite{Kumar2020-zb} ---
both Baird's advantage-learning
and \textit{Q-constrained} offline RL $Q_\omega$ objectives
(\textit{e.g.}~\cite{Wu2019-nl}, \cite{Kumar2020-zb})
are similar in spirit.
Concretely, we add the $\alpha$-scaled \emph{advantage}
$\alpha \, A^{\pi_\theta}_\omega(s,a)$ to $Q_\omega(s,a)$'s target
in the temporal-difference objective $\ell_\omega$
that updates $Q_\omega$ over the offline dataset $\mathcal{D}$
\big(\textit{i.e.}~$s \sim \rho^\beta(\cdot)$, $a \sim \beta(\cdot | s)$, $s' \sim \rho^\beta(\cdot)$\big).
We define the advantage $A^{\pi_\theta}_\omega$ over $\mathcal{D}$ as
$A^{\pi_\theta}_\omega(s,a) \coloneqq Q_\omega(s,a) -
\mathbb{E}_{\bar{a} \sim \pi_\theta}[Q_\omega(s,\bar{a})]$,
where the expectation is estimated with the usual unbiased empirical mean.
Note, $A^{\pi_\theta}_\omega$ takes values in $\mathbb{R}$, while $\alpha > 0$.
Since tuning the scale of bonuses or penalties added to $Q_\omega$'s target (\textit{e.g.}~\cite{Wu2019-nl})
or $Q_\omega$'s objective (\textit{e.g.}~\cite{Kumar2020-zb})
has proved tremendously tedious due to the stiffness (\textit{cf.} definition in \textsc{Section}~\ref{bg})
of such hyper-parameter,
we conducted a grid search over values separated by equal spaces and ranging from $0.1$ to $0.9$.
The latter upper bound is set to such value
since that is the highest, still theoretically-principled, value that
can be assumed by Baird's advantage-learning scaling coefficient according to \cite{Bellemare2015-ii}
--- although it has later been argued otherwise in \cite{Lu2019-tp}.
Note, our advantage-learning bonus is \emph{only} applied on points from the offline dataset $\mathcal{D}$,
as our add-on \emph{concretely} changes $\ell_\omega$ as described in \textsc{eq}~\ref{criticloss} into
the following objective:
\begin{align}
  \ell^\textsc{al}_\omega \coloneqq
  \mathbb{E}_{s \sim \rho^\beta(\cdot), a \sim \beta(\cdot | s), s' \sim \rho^\beta(\cdot)}
  \bigg[
  \Big(
  Q_\omega(s,a) -
  \big(
  r(s, a, s') + \gamma \, \mathbb{E}_{a' \sim \zeta(\cdot | s')}
  \big[
  Q_{\omega'}(s',a') + \alpha \, A^{\pi_\theta}_\omega(s,a)
  \big]
  \big)
  \Big)^2
  \bigg]
  \label{criticlossal}
\end{align}
We study how using the loss defined in \textsc{eq}~\ref{criticlossal}
affects final performance over the grid of $\alpha$ values reported above,
and report our empirical findings in \textsc{Appendix}~\ref{bairdal}, \textsc{Figure}~\ref{bairdal:barplot}.
We observe that there is little to gain from such an add-on, and perhaps more importantly
that there is a lot to lose judging by how stiff the new problem is
\textit{w.r.t.} the newly introduced hyper-parameter $\alpha$. Consequently, we do not use
this revisited version of Baird's advantage-learning in any of our experiments
except the ablation we have carried out in \textsc{Appendix}~\ref{bairdal}, \textsc{Figure}~\ref{bairdal:barplot}.

\paragraph{Complementary investigations \#2: the effect of the hyper-parameter $m$ in the
$\mathcal{T}_\textsc{Max}^{\omega, m}$ operator.}

Finally, we investigate the impact of the hyper-parameter $m$ on the agent's performance,
where $m$ is involved in the operator $\mathcal{T}_\textsc{Max}^{\omega, m}$ introduced in
\textsc{Section}~\ref{operators} to be used in the design of the proposal distributions
$\mathcal{T}_\textsc{Max}^{\omega', m}\big[\beta_\textsc{c}\big]$,
$\mathcal{T}_\textsc{Max}^{\omega', m}\big[\beta_\textsc{c}^\xi\big]$, and
$\mathcal{T}_\textsc{Max}^{\omega', m}[\pi_\theta]$
--- \ref{betaclonemax}, \ref{perturbedbetaclonemax}, and \ref{thetamax}, respectively.
Concretely, $m$ is the number of times we sample actions before selecting the one with the highest value
in said operators.
In essence, $m$ controls the degree of interpolation with $Q^*$, as discussed earlier in
\textsc{Section}~\ref{proposalsimplex}.
We report our empirical findings in \textsc{Appendix}~\ref{tmaxops},
\textsc{Figures}~\ref{betaclonemaxm:barplot}, \ref{perturbedbetaclonemaxm:barplot}, and
\ref{thetamaxm:barplot} respectively.
In short, these figures show that while the \ref{betaclonemax} and \ref{perturbedbetaclonemax} proposal
distributions are fairly resilient (and opposed to stiff)
to changes in the value of $m$, \ref{thetamax} often displays significant gaps in performance between distinct
values of $m$, in line with our previous discussions about the \ref{thetamax} proposal being far more exposed to
out-of-distribution actions than \ref{betaclonemax} and \ref{perturbedbetaclonemax}.
Increasing $m$ increases the chance of involving an arbitrarily overestimated $Q_\omega$ value to the set
of $m$ values the operator $\mathcal{T}_\textsc{Max}^{\omega, m}$ takes the $\argmax$ over,
which explains the greater spread in performance for the proposal that does not involve a mechanism to
ensure the actor's policy $\pi_\theta$ remains close to $\beta$.

\section{Generalizing policy improvement: the Generalized Importance-Weighted Regression (GIWR) framework}
\label{pi}

In \textsc{Section}~\ref{pe}, we undertook the design of several proposal policies $\zeta$ to
sample the \textit{next} action from in the TD learning update
\cite{Sutton1984-ce, Sutton1988-to, Sutton1999-ii}
of the critic $Q_\omega$, giving rise to as many variants of Bellman's equation.
We analyzed and reported the impact of each of these on the agent's learning dynamics and final asymptotic
performance in \textsc{Section}~\ref{peresults}, by changing the proposal policy used in the policy evaluation
strategy while keeping the policy improvement subroutine of each policy iteration step identical and fixed.
\textbf{Here, instead, we vary the proposal distribution used in the new actor update
method we introduce, while keeping the policy evaluation method set to the standard
\textit{SARSA} actor-critic update, ubiquitous in both online and offline RL, \ref{theta}.}
There might exist synergies between considered form of the policy improvement update and the use of
proposal distributions other than \ref{theta} in the policy evaluation objective $\ell_\omega$ from
\textsc{Section}~\ref{pe}.
We believe looking into such synergies in depth to be an avenue of interest for future reaserch,
although the compute required for such inquiry is considerable.
\textbf{Still, we do evaluate the configurations that show promise and that are expected
to reveal such synergies between design choices.}

\paragraph{N.B.}

The framework still allows for the use of any proposal strategy introduced
in \textsc{Section}~\ref{pe} in policy evaluation.

\subsection{Generalized constrained policy improvement}
\label{actorupdate}
Our derivation is inspired from the derivations of the almost-identical constrained optimization problem
carried out in several KL-control works, which we divide in three waves based on when
the respective works appeared:
it was first reported in REPS \cite{Peters2010-vd},
RWR \cite{Peters2007-qb, Kober2010-hy},
and LAWER \cite{Neumann2008-tm},
then later in AWR \cite{Peng2019-hu},
whose reminiscent elements appear in TRPO's derivation as well \cite{Schulman2015-jt},
to finally reemerge later in the concurrent works CRR \cite{Wang2020-sr}
and AWAC \cite{Nair2020-gd}.
Despite sharing most of the mechanisms overlapping in each of these waves,
our derivation involves a slightly altered constraint in the initial constrained optimization problem
formulation. As such, we solve the said problem analytically from the start, to arrive at a tractable solution
taking into account our change in the original formulation. The proposed problem alteration and its provably-adapted
and computationally-tractable solution provide a \textbf{generalized framework that allows the practitioner
to involve additional constraints to the original \textit{KL-control}-based constrained optimization problem
using any proposal distribution introduced and discussed earlier} in \textsc{Section}~\ref{pe}.

While REPS \cite{Peters2010-vd}
and RWR \cite{Peters2007-qb, Kober2010-hy},
maximize the expected return $J(\pi)$,
LAWER \cite{Neumann2008-tm},
CPI \cite{Kakade2002-kw},
TRPO \cite{Schulman2015-jt},
MARWIL \cite{Wang2018-dn},
MPO \cite{Abdolmaleki2018-sp},
AWR \cite{Peng2019-hu},
CRR \cite{Wang2020-sr}
and AWAC \cite{Nair2020-gd}
maximize the expected improvement $\eta(\pi)$.
In other words, while the former group intends to learn policies that maximize the \emph{action-value}
from the start state, the latter group cares about the maximization of the \emph{advantage} from the start state:
$\eta(\pi) \coloneqq \mathbb{E}_{s \sim \rho^\pi(\cdot), a \sim \pi(\cdot | s)}[A^\pi(s,a)]$.
Nevertheless, since we work under the offline RL setting, we only have access to states $s$ coming from
the offline dataset $\mathcal{D}$, \textit{i.e.}~distributed as $s \sim \rho^\beta(\cdot)$.
We therefore define a \emph{surrogate} objective $\tilde{\eta}^\beta(\pi)$ that, by contrast with $\eta(\pi)$,
we \emph{can} evaluate in the offline setting:
$\tilde{\eta}^\beta (\pi) \coloneqq \mathbb{E}_{s \sim \rho^\beta(\cdot), a \sim \pi(\cdot | s)}[A^\pi(s,a)]$.
The severity of this relaxation depends on how well
the state visitation distribution $\rho^\pi$ matches the one
observed in the offline dataset, \textit{i.e.}~$\rho^\beta$.
As such, if $\pi$ and $\beta$ lead the agent to the same states
such that $\rho^\pi \approx \rho^\beta$,
then $\tilde{\eta}^\beta(\pi)$ approximates $\eta(\pi)$ well.
The relaxation is then \emph{mild}.
Crucially, since we often encourage the learned policy $\pi$ to remain close to $\beta$ in offline RL to
avoid out-of-distribution actions in $Q_\omega$, then the approximation
$\tilde{\eta}^\beta(\pi) \approx \eta(\pi)$
is even more likely to be satisfied in the offline setting.

We tie the new iterate of the actor's policy $\pi_{\theta^\text{new}}$ to the previous one, $\pi_{\theta^\text{old}}$,
via the constrained optimization problem that follows (\textit{cf.}~\textsc{Section}~\ref{bg} for a reminder of
how we denote either direction of the KL divergence in this work):
\begin{align}
  \pi_{\theta^\text{new}}
  \in
  &\argmax_{\pi \in \mathcal{P}(\mathcal{A})^\mathcal{S}} \;\,
  \tilde{\eta}^\beta (\pi)
  \label{piobjectiveorig1} \\
  &\text{s.t.} \quad
  (\forall s \in \mathcal{S}) \quad
  D^\zeta_{\overleftarrow{\textsc{kl}}}[\pi](s) \leq \delta
  \label{piobjectiveorig2} \\
  &\phantom{\text{s.t.}} \quad
  (\forall s \in \mathcal{S}) \quad
  \int_{a \in \mathcal{A}} \pi(a | s) \, da = 1
  \label{piobjectiveorig3}
\end{align}
Similarly to how we could not evaluate $\eta(\pi)$ and had to use its
relaxation $\tilde{\eta}^\beta (\pi)$ using states from $\mathcal{D}$
as a surrogate objective,
we also apply an identical relaxation to the equality constraint.
As such,
``$(\forall s \in \mathcal{S})$, $D^\zeta_{\overleftarrow{\textsc{kl}}}[\pi](s) \leq \delta$'' becomes
``$\mathbb{E}_{s \sim \rho^\beta(\cdot)}
\big[D^\zeta_{\overleftarrow{\textsc{kl}}}[\pi](s)\big] \leq \delta $''.
Here, instead of attempting to enforce the
equality constraint over the entirety of  $\mathcal{S}$,
we restrict the constraint's field of view to $\mathcal{D}$,
the only subspace over which we \emph{can} enforce it.
Lastly, we relax the theoretical placeholder of the advantage $A^\pi$ with
$A^{\pi_{\theta^\text{old}}}_\omega$, defined as
$A^{\pi_{\theta^\text{old}}}_\omega(s,a) \coloneqq Q_\omega(s,a) -
\mathbb{E}_{\bar{a} \sim \pi_{\theta^\text{old}}}[Q_\omega(s,\bar{a})]$,
where the expectation is estimated with the usual unbiased empirical mean.
The equality constraint (urging $\pi$ to describe a probability distribution over $\mathcal{A}$,
$\forall s \in \mathcal{S}$) can not be relaxed
as we need this property to be distilled into the learned $\pi$ entirely.
After applying these relaxations, the constrained optimization problem we set out to solve is:
\begin{align}
  \pi_{\theta^\text{new}}
  \in
  &\argmax_{\pi \in \mathcal{P}(\mathcal{A})^\mathcal{S}} \;\,
  \mathbb{E}_{s \sim \rho^\beta(\cdot), a \sim \pi(\cdot | s)}
  \big[A^{\pi_{\theta^\text{old}}}_\omega(s,a)\big]
  \label{piobjective} \\
  &\text{s.t.} \quad
  \mathbb{E}_{s \sim \rho^\beta(\cdot)}\big[D^\zeta_{\overleftarrow{\textsc{kl}}}[\pi](s)\big] \leq \delta
  \label{piineq} \\
  &\phantom{\text{s.t.}} \quad
  (\forall s \in \mathcal{S}) \quad
  \int_{a \in \mathcal{A}} \pi(a | s) \, da = 1
  \label{pieq}
\end{align}
Via \emph{Lagrange-Duality}, we define the following Lagrangian from the relaxed objective
laid out in \textsc{eq}~\ref{piobjective}, subjected to both
\textit{a)} the inequality constraint in \textsc{eq}~\ref{piineq}
encouraging the learned policy $\pi$ to be close in reverse KL to
the proposal policy $\zeta$, and
\textit{b)} the equality constraint in \textsc{eq}~\ref{pieq}
ensuring $\pi$ defines a proper conditional probability distribution.
In particular, to deal with the inequality constraint (\textit{cf.}~\textsc{eq}~\ref{piineq}),
we carry out the derivations under the KKT conditions:
\begin{align}
  \mathcal{L}(\pi, \lambda_\textsc{kl}, \lambda) \coloneqq
  & \;\, \mathbb{E}_{s \sim \rho^\beta(\cdot), a \sim \pi(\cdot | s)}
  \big[A^{\pi_{\theta^\text{old}}}_\omega(s,a)\big] \nonumber \\
  &- \lambda_\textsc{kl}
  \Big[
  \mathbb{E}_{s \sim \rho^\beta(\cdot)}\big[D^\zeta_{\overleftarrow{\textsc{kl}}}[\pi](s)\big] - \delta
  \Big]
  - \int_{s \in \mathcal{S}} \lambda(s)
  \bigg[
  \int_{a \in \mathcal{A}} \pi(a | s) \, da - 1
  \bigg]
  \, ds
  \label{firstlagrangian}
\end{align}
where $\lambda_\textsc{kl} \in (0, \infty)$ is the Lagrange multiplier associated with
the \textit{KL}-based inequality constraint (\textit{cf.}~\textsc{eq}~\ref{piineq}),
and $\lambda: \mathcal{S} \to (0, \infty)$ is a function that, for all $s \in \mathcal{S}$,
returns a Lagrange multiplier for \emph{each} equality constraint (\textit{cf.}~\textsc{eq}~\ref{pieq}).
By expanding the expectations (and KL divergence)
into integrals in \textsc{eq}~\ref{firstlagrangian}, we trivially obtain
the following:
\begin{align}
  \mathcal{L}(\pi, \lambda_\textsc{kl}, \lambda) \coloneqq
  & \;\, \int_{s \in \mathcal{S}} \rho^\beta(s) \int_{a \in \mathcal{A}} \pi(a | s) \,
  A^{\pi_{\theta^\text{old}}}_\omega(s,a) \nonumber \\
  &- \lambda_\textsc{kl}
  \bigg[
  \int_{s \in \mathcal{S}} \rho^\beta(s) \int_{a \in \mathcal{A}} \pi(a | s)
  \big(
  \log \pi(a | s) - \log \zeta(a | s)
  \big) \, da
  - \delta
  \bigg] \nonumber \\
  &- \int_{s \in \mathcal{S}} \lambda(s)
  \bigg[
  \int_{a \in \mathcal{A}} \pi(a | s) \, da - 1
  \bigg]
  \, ds
  \label{integralslagrangian}
\end{align}
Then, by taking the first and second derivatives of $\mathcal{L}(\pi, \lambda_\textsc{kl}, \lambda)$ as
expressed in \textsc{eq}~\ref{integralslagrangian} with respect to $\pi(a | s)$, we get:
\begin{align}
  \pdv{\mathcal{L}(\pi, \lambda_\textsc{kl}, \lambda)}{\pi(a | s)}
  &= \rho^\beta(s) \, A^{\pi_{\theta^\text{old}}}_\omega(s,a)
  - \lambda_\textsc{kl} \, \rho^\beta(s)
  \big(\log \pi(a | s) + 1 - \log \zeta(a | s)\big)
  - \lambda(s)
  \\
  &= B - \lambda_\textsc{kl} \, \rho^\beta(s) \log \pi(a | s)
  \quad \text{with }
  B \coloneqq \rho^\beta(s) \, \Big(A^{\pi_{\theta^\text{old}}}_\omega(s,a)
  - \lambda_\textsc{kl} \, \big(1 - \log \zeta(a | s)\big)\Big) - \lambda(s)
  \label{firstderivlagragian}
  \\
  \pdv[2]{\mathcal{L}(\pi, \lambda_\textsc{kl}, \lambda)}{\pi(a | s)}
  &= - \frac{\lambda_\textsc{kl} \, \rho^\beta(s)}{\pi(a | s)} \leq 0
  \qquad \implies \text{\emph{critical points} of $\mathcal{L}(\pi, \lambda_\textsc{kl}, \lambda)$
  w.r.t. $\pi(a | s)$ are \emph{maxima}.}
  \label{secondderivlagragian}
\end{align}
Since $\mathcal{L}(\pi, \lambda_\textsc{kl}, \lambda)$ is \emph{concave} with respect to $\pi(a | s)$
over the studied spaces, we look for the $\argmax$ of the objective (\textit{cf.}~\textsc{eq}~\ref{piobjective})
seeking a critical point of the Lagrangian $\mathcal{L}(\pi, \lambda_\textsc{kl}, \lambda)$ along the $\pi$
dimension, which we name $\pi^*$:
\begin{align}
  \pdv{\mathcal{L}(\pi, \lambda_\textsc{kl}, \lambda)}{\pi(a | s)}\bigg|_{\pi^*} = 0
  \iff B - \lambda_\textsc{kl} \, \rho^\beta(s) \log \pi^*(a | s) = 0
  \iff \log \pi^*(a | s) = \frac{B}{\lambda_\textsc{kl} \, \rho^\beta(s)}
\end{align}
Hence,
\begin{align}
  \pi^*(a | s) = C(s) \zeta(a | s) \exp (\frac{1}{\lambda_\textsc{kl}} A^{\pi_{\theta^\text{old}}}_\omega(s,a))
  = C(s) \varphi(a | s)
  \quad \text{with }
  C(s) &\coloneqq \exp \bigg( - \bigg[ 1 + \frac{\lambda(s)}{\lambda_\textsc{kl} \,\rho^\beta(s)} \bigg] \bigg) \\
  \quad \text{and }
  \varphi(a | s) &\coloneqq
  \zeta(a | s) \exp (\frac{1}{\lambda_\textsc{kl}} A^{\pi_{\theta^\text{old}}}_\omega(s,a))
\end{align}
where $\varphi$ is the \emph{unnormalized} advantage-weighted counterpart of $\zeta$.
$\varphi$ is \emph{not} a PDF, and will need to be normalized to be one.
We do so using the equality constraint (\textit{cf.}~\textsc{eq}~\ref{pieq})
encoding our desideratum for $\pi$ to be a probability distribution, which \textit{a fortiori}
naturally also applies to the point $\pi^*$ maximizing $\mathcal{L}(\pi, \lambda_\textsc{kl}, \lambda)$:
\begin{align}
  (\forall s \in \mathcal{S}) \quad
  \int_{a \in \mathcal{A}} \pi^*(a | s) \, da = 1
  \iff
  C(s) \int_{a \in \mathcal{A}} \varphi(a | s) \, da = 1
  \iff
  C(s) = \frac{1}{\varphi(s)}
\end{align}
where $\varphi(s) \coloneqq \int_{a \in \mathcal{A}} \varphi(a | s) \, da$ is the Bayesian evidence,
or partition function.
As such, for $\pi^*(a | s) = C(s) \varphi(a | s)$ to define a PDF,
we need $C(s)$ to satisfy $C(s) = 1 \, \big/ \, \int_{a \in \mathcal{A}}
\big[\zeta(a | s) \exp (A^{\pi_{\theta^\text{old}}}_\omega(s,a) / \lambda_\textsc{kl})\big] \, da$.
Hence, $\pi^*$ verifying:
\begin{align}
  \pi^*(a | s)
  = \frac{1}{\varphi(s)} \zeta(a | s) \exp (\frac{1}{\lambda_\textsc{kl}} A^{\pi_{\theta^\text{old}}}_\omega(s,a))
  \label{pistar}
\end{align}
defines a PDF since $\int_{a \in \mathcal{A}} \pi^*(a | s) \, da = 1$.
\textit{In fine}, such $\pi^*$ is the \emph{normalized} advantage-weighted counterpart of
the proposal policy $\zeta$
--- the policy $\zeta$ being the \textit{trajectory distribution} in the \textit{KL}-control literature,
\textit{e.g.}~in \cite{Peters2007-qb, Kober2010-hy, Peters2010-vd, Neumann2011-hn}).
As such, we can refer to $\pi^*$ as defined in \textsc{eq}~\ref{pistar} as the \emph{advantage-weighted
proposal policy}. To disambiguate the notations, the advantage-weighted counterpart of the
proposal policy $\zeta$ will be denoted as $\zeta_\textsc{iw}$, \textit{i.e.}~$\zeta_\textsc{iw} \coloneqq \pi^*$
(the acronym ``\textsc{iw}'' standing for \textit{i}mportance-\textit{w}eighted,
where \textit{``importance''} here plays the role of universal, unifying placeholder for either \textit{reward} or
\textit{advantage} depending on the considered method).
As in all the previous work cited in this section for either reporting or building on the derivation of the present
derivation (or a variant thereof) we stick to the traditional E-M scheme.
Constructing $\zeta_\textsc{iw}$, whose assembly procedure is described in \textsc{eq}~\ref{pistar},
is nevertheless tedious since computing the evidence $\varphi(s)$ in \textsc{eq}~\ref{pistar} requires
an inordinate amount of compute to estimate exactly (\textit{cf.}~Bayesian ML,
energy-based models in particular, assembling Boltzmann distributions in a similar fashion).

\subsection{Projection options for distributional shift mitigation}
\label{projectionoptions}

Yet, instead of trying to relax said evidence or find a more computationally affordable surrogate,
we treat the intractable analytical solution $\zeta_\textsc{iw}$ (\textit{cf.}~\textsc{eq}~\ref{pistar})
as an \textit{input} in a subsequent, distinct, unconstrained optimization problem (in line with the E-M
procedural paradigm). Said optimization problem is defined as follows:
\begin{align}
  \theta
  \in
  &\argmin_{\theta \in \Theta} \;\,
  \mathbb{E}_{s \sim \rho^\beta(\cdot)}
  \Big[
  \Delta\big(\pi_\theta(\cdot | s), \zeta_\textsc{iw}(\cdot | s)\big)
  \Big]
  \quad \text{with }
  \Delta
  \text{ being a measure between probability distributions.}
  \label{piobjectivenew}
\end{align}
In this subsequent problem, we set out to find a tractable decision-making rule by directly \emph{projecting} the
intractable advantage-weighted proposal policy $\zeta_\textsc{iw}$
onto the manifold of parametric policies $\{ \pi_\theta \, | \, \theta \in \Theta \}$
we are able to estimate empirically.
By construction, we can therefore hope to afford to compute the solution to this problem.
We opt for the KL divergence as our choice of measure $\Delta$ to perform such projection.
Since this measure is asymmetric, we have two options:
we can either perform an \textit{I}-projection (reverse, exclusive KL),
or a \textit{M}-projection (forward, inclusive KL), whose respective advantages and drawbacks are discussed
in detail in the books of MacKay \cite{MacKay2003-qn},
Bishop \cite{Bishop2006-xz},
and Murphy \cite{Murphy2012-ih}.
Consider the projection in KL divergence
of the target distribution $p$ onto the set of parametric distributions
in which we look for $q_\theta$, where $\theta \in \Theta$.
In short, an \textit{M}-projection (\textit{``M''} for \textit{M}oment)
will have the effect of making $q_\theta$ \textit{cover} the modes of $p$
(\textit{``mode-covering''}),
caring much about \emph{not} assigning zero density wherever $p$ has non-zero probability (\textit{``zero-avoiding''}),
yet not caring much about wrongly assigning non-zero density outside the support of $p$.
Conversely, an \textit{I}-projection (\textit{``I''} for \textit{I}nformation)
will have the effect of making $q_\theta$ \textit{seek} the modes of $p$
(\textit{``mode-seeking''}),
caring much about \emph{not} assigning \emph{non-}zero density wherever $p$ has zero probability
(\textit{``non-zero-avoiding''}, by symmetry),
yet not caring much about assigning zero density inside the support of $p$, missing areas where $p$
has non-zero probability.
Said differently, the ultimate priority of an M-projection is to \emph{not miss anything inside} the support of $p$,
while the ultimate priority of an I-projection is to \emph{miss everything outside} the support of $p$.
As such, it is easy to see how using an I-projection (reverse KL)
on a target trajectory distribution would have the
indirect effect of learning \emph{``cost-averse''} policies,
while an M-projection (forward KL)
would make policies \emph{``reward-chasing''}.
In \cite{Gu2015-ax}, \textsc{Section~3},
the authors give four reasons as to why using an M-projection as optimization objective might be beneficial.
Their third reason posits that the projection resulting from a forward KL objective tends to
display a higher entropy than the target distribution, which makes the projected proposal policy a good
candidate for importance sampling, in various sub-areas (\textit{e.g.}~Monte-Carlo estimation in
\cite{Gu2015-ax}).
This claim is supported by \cite{MacKay2003-qn},
defending that the reverse KL, or I-projection, conversely does \emph{not} yield proposal distributions
suited for importance sampling.
Besides, higher-entropy policies are naturally equipped with dithering capabilities to trade off with their
greedy incentive to maximize $Q_\omega$, enabling them to be reasonably proficient at exploring their environment
without needing extra dithering mechanisms.

In offline RL however, high-entropy policies are more likely to evaluate the critic's value $Q_\omega$
at out-of-distribution actions
\textit{a)} in policy evaluation at training time,
provided the actor's policy $\pi_\theta$ is used by the proposal distribution
generating the next action in the temporal-difference objective, and
\textit{b)} in policy improvement at training time \emph{and} evaluation time.
Indeed, in the specific case of offline RL, it is far safer to perform I-projections, ensuring
the learned policies are \emph{not} assigning non-zero weight to actions \emph{outside} the support of the
target policy involved in the projection.
Otherwise (using M-projections), by trying to cover all the modes of the target policy, the projected policy
would be urged to \emph{fill in} gaps in between peaks of the target distribution by
assigning density where the target assigns none.
The propensity to put overshoot the assigned density is particularly detrimental when gaps are numerous,
\textit{i.e.}~when the target distribution is \emph{not} concave
(\textit{e.g.}~when the latter is multimodal).
As such, there would be a distributional shift between the projected and the target distributions,
causing severe instabilities both at training time and evaluation time, given that we are working in
the offline setting in which collecting more data via interactions with the world is not allowed.
Remaining \textit{in}-distribution is paramount, and keeping the entropy down by leveraging
I-projection rather than M-projection appears as the safest option to achieve this desideratum.
Besides, since the agent need only exploit --- and not explore --- in offline RL,
possessing the natural exploration capabilities enabled by following a higher-entropy policy
is void of benefit in offline RL, in contrast with online RL.
As such, the problem formulation in \textsc{Eqs}~\ref{piobjectiveorig1},
\ref{piobjectiveorig2}, and \ref{piobjectiveorig3} --- which has been adopted in a slew of works such as
REPS \cite{Peters2010-vd},
RWR \cite{Peters2007-qb, Kober2010-hy},
LAWER \cite{Neumann2008-tm},
CPI \cite{Kakade2002-kw},
VIP \cite{Neumann2011-hn},
TRPO/PPO \cite{Schulman2015-jt, Schulman2017-ou} (forward KL constraint instead of reverse KL),
MPO \cite{Abdolmaleki2018-sp},
AWR \cite{Peng2019-hu},
ABM \cite{Siegel2020-lo},
CRR \cite{Wang2020-sr}
and AWAC \cite{Nair2020-gd}
--- is particularly well-suited to the \emph{offline} RL setting
(under which
MARWIL \cite{Wang2018-dn},
AWR \cite{Peng2019-hu},
ABM \cite{Siegel2020-lo},
CRR \cite{Wang2020-sr}
and AWAC \cite{Nair2020-gd}
are framed)
where the cost-aversion encoded via and distilled by I-projections
enables the learned agents to prevent straying from the
behavior policy into a distributional shift where errors compound.
The reverse KL constraint in \textsc{eq}~\ref{piobjectiveorig2} leads to \textsc{eq}~\ref{piobjectivenew}
via the exhibited derivation (\textit{cf.}~beginning of \textsc{Section}~\ref{actorupdate}).
Since all of these past methods have gone through or have reused said derivations of similar flavor,
they (and we) all face the same \emph{new} objective as reported in \textsc{eq}~\ref{piobjectivenew}, and
are then subject to the same subsequent task consisting of choosing a measure $\Delta$.
All of these works have opted for a KL divergence.
Based on the arguments posited above, picking the \emph{reverse} KL, an I-projection, seem like the natural
choice given how detrimental and unforgiving naively chasing after rewards
(like an M-projection would dictate)
seems to be --- in offline RL above all else.
Starting from the objective in \textsc{eq}~\ref{piobjectivenew}, we now lay out the derivations
of said objective \textit{1)} using the forward KL for $\Delta$, and \textit{2)} using the reverse KL for $\Delta$.
Our aim is to highlight that, while there is a striking claim for an I-projection,based the discussion that precedes,
the M-projection is inordinately easier to compute than the I-projection, leaving us with a trade-off to balance.
We begin with the \emph{forward} KL, by unpacking the measure and the expectations into explicit integral form,
and injecting \textsc{eq}~\ref{pistar}:
\begin{align}
  \mathbb{E}_{s \sim \rho^\beta(\cdot)}
  \Big[
  \Delta\big(\pi_\theta(\cdot | s), \zeta_\textsc{iw}(\cdot | s)\big)
  \Big]
  &\coloneqq
  \mathbb{E}_{s \sim \rho^\beta(\cdot)}
  \Big[
  D^{\zeta_\textsc{iw}}_{\overrightarrow{\textsc{kl}}}[\pi_\theta](s)
  \Big]
  \label{mprojection}
\end{align}
\begin{align}
  \implies \quad
  \theta
  &\in
  \argmin_{\theta \in \Theta} \;\,
  \mathbb{E}_{s \sim \rho^\beta(\cdot)}
  \Big[
  \Delta\big(\pi_\theta(\cdot | s), \zeta_\textsc{iw}(\cdot | s)\big)
  \Big]
  \\
  &= \argmin_{\theta \in \Theta} \;\,
  - \int_{s \in \mathcal{S}} \rho^\beta(s) \int_{a \in \mathcal{A}}
  \zeta_\textsc{iw}(a | s)
  \log \pi_\theta(a | s) \, da \, ds \\
  &= \argmin_{\theta \in \Theta} \;\,
  - \int_{s \in \mathcal{S}} \rho^\beta(s) \int_{a \in \mathcal{A}}
  \zeta(a | s) \exp (\frac{1}{\lambda_\textsc{kl}} A^{\pi_\theta}_\omega(s,a))
  \log \pi_\theta(a | s) \, da \, ds \\
  &= \argmax_{\theta \in \Theta} \;\,
  \mathbb{E}_{s \sim \rho^\beta(\cdot), a \sim \zeta(\cdot | s)}
  \bigg[
  \exp (\frac{1}{\lambda_\textsc{kl}} A^{\pi_\theta}_\omega(s,a)) \log \pi_\theta(a | s)
  \bigg]
  \label{mprojectionargmin}
\end{align}
Conversely, by opting for the \emph{reverse} KL instead, the problem in \textsc{eq}~\ref{pistar}
reduces to the following problem:
\begin{align}
  \mathbb{E}_{s \sim \rho^\beta(\cdot)}
  \Big[
  \Delta\big(\pi_\theta(\cdot | s), \zeta_\textsc{iw}(\cdot | s)\big)
  \Big]
  &\coloneqq
  \mathbb{E}_{s \sim \rho^\beta(\cdot)}
  \Big[
  D^{\zeta_\textsc{iw}}_{\overleftarrow{\textsc{kl}}}[\pi_\theta](s)
  \Big]
  \label{iprojection}
\end{align}
\begin{align}
  \implies \quad
  \theta
  &\in
  \argmin_{\theta \in \Theta} \;\,
  \mathbb{E}_{s \sim \rho^\beta(\cdot)}
  \Big[
  \Delta\big(\pi_\theta(\cdot | s), \zeta_\textsc{iw}(\cdot | s)\big)
  \Big]
  \\
  &= \argmin_{\theta \in \Theta} \;\,
  \int_{s \in \mathcal{S}} \rho^\beta(s) \int_{a \in \mathcal{A}}
  \pi_\theta(a | s)
  \log \bigg(\zeta(a | s) \exp (\frac{1}{\lambda_\textsc{kl}} A^{\pi_\theta}_\omega(s,a))\bigg) \, da \, ds
  \nonumber \\
  & \qquad -
  \int_{s \in \mathcal{S}} \rho^\beta(s) \int_{a \in \mathcal{A}} \pi_\theta(a | s)
  \log \pi_\theta(a | s) \, da \, ds
  \\
  &= \argmin_{\theta \in \Theta} \;\,
  \mathbb{E}_{s \sim \rho^\beta(\cdot), a \sim \pi_\theta(\cdot | s)}
  \bigg[
  \log \zeta(a | s) + \frac{1}{\lambda_\textsc{kl}} A^{\pi_\theta}_\omega(s,a)
  \bigg]
  + \mathbb{E}_{s \sim \rho^\beta(\cdot)} \big[H\big(\pi_\theta(\cdot | s)\big)\big]
  \label{iprojectionargmin}
\end{align}
where $H\big(\pi_\theta(\cdot | s)\big)$ denotes the entropy of $\pi_\theta$ for a given state $s$
(for more detailed derivations, see
\textsc{Appendix}~\ref{agentobjectivedetail}).
Directly echoing our previous discussion about the hurdles of high-entropy policies in offline RL,
we observe that \textsc{eq}~\ref{iprojectionargmin} directly involves the entropy of the actor's policy
$H\big(\pi_\theta(\cdot | s)\big)$ estimated over states from the offline dataset $\mathcal{D}$.
Crucially, we see that when designing $\Delta$ as an I-projection, the problem of finding the parametric
policy that minimizes $\Delta\big(\pi_\theta(\cdot | s), \zeta_\textsc{iw}(\cdot | s)\big)$
over states from the dataset $\mathcal{D}$ reduces to a formulation in \textsc{eq}~\ref{iprojectionargmin}
where we need to \emph{minimize} the entropy of $\pi_\theta$ on said distribution of states $s \sim \rho^\beta(\cdot)$.
In other words, the I-projection urges the policy $\pi_\theta$ to have the \emph{lowest} bandwidth possible
so as to place the least amount of density outside the support of the target distribution,
fitting the modes tightly (albeit likely ignoring some of them in non-concave target scenarios, as developed earlier).
The I-projection however tends to make the learned policy collapse \cite{Abdolmaleki2018-sp}.
Despite being somewhat aligned with our conservative desiderata
--- and setting aside the policy's propensity to collapse since it can be alleviated via regularization
with relative ease,
the reduction in \textsc{eq}~\ref{iprojectionargmin} faces two hurdles
that make the reduced problem tedious to solve efficiently in practice.

First, we need a model of $\zeta$ that enables the evaluation of the likelihood of an action
at a given state $\zeta(a | s)$ --- which might be already readily available depending on how the proposal
policy $\zeta$ is defined (\textit{cf.}~we lay out the $\zeta$ options considered in this work
in \textsc{Section}~\ref{pe}).
In the particular case where $\zeta \coloneqq \beta$, \textit{i.e.}~sampling actions from $\zeta$ simply
means picking actions from the offline dataset $\mathcal{D}$, which means we do not have any way to evaluate the
likelihood of an action according to $\beta$ \emph{other than} modelling the offline distribution $\beta$
underlying the dataset (equivalently, $\zeta$) with a model that enables such evaluation.
Notably, one could sidestep the need for a likelihood model estimating probability densities
by leveraging a conditional \textit{score} density estimator,
whose returned score \emph{can} be used as proxies for said likelihoods.
Relaxing the problem even further, via Bayes' rule, one could craft a surrogate for said conditional score
from a joint score over $\mathcal{S} \times \mathcal{A}$ and a score over $\mathcal{S}$, which, on top
of being easier to estimate in most cases, would naturally regulate the scale of the assembled
conditional score. The density, novelty, or uncertainty estimator formalized as $\rho$ in
\textsc{eq}~\ref{rhodef} is a suitable candidate to build a proxy for $\zeta(a | s)$
when $\zeta \coloneqq \beta$.
We refer the reader to the discussion surrounding \textsc{eq}~\ref{rhodef} in \textsc{Section}~\ref{pe}
about potential practical candidates for $\rho$,
along with references to works leveraging such estimators.

The second reason why \textsc{eq}~\ref{iprojectionargmin} can be tedious to estimate is due to the presence of
an expectation over samples from the \textit{very} model we set out to update,
in both pieces of the operand.
In other words, directly implementing the reduced objective of
\textsc{eq}~\ref{iprojectionargmin} means sending gradients backwards through the stochastic sampling unit
``$\sim \pi_\theta(\cdot |s)$'' to update $\pi_\theta$ --- a non-differentiable operator as is.
There are nevertheless numerous tricks to bypass this hurdle.
Using a reparametrization trick is the most popular option, first popularized as such in the context of
variational auto-encoders in \cite{Kingma2014-hf, Rezende2014-ef} for Gaussian distributions,
then extended to a wider class of variational distributions (\textit{e.g.}~beta and gamma distributions)
in \cite{Ruiz2016-qe},
then concurrently adapted to the categorical distribution in \cite{Jang2017-tu, Maddison2017-gk}
by leveraging the Gumbel distribution, \cite{Gumbel1954-xu}.
These have seen wide adoption in RL since \cite{Heess2015-va}.
Considering that these encompass virtually every distribution usually used in RL to model the learned policy,
one rarely need look elsewhere. Still, in more exotic scenarios, one can turn to
REBAR \cite{Tucker2017-zj},
LAX or RELAX \cite{Grathwohl2018-dw},
the \textit{straight-through} estimator \cite{Bengio2013-hg},
or the archetype \textit{REINFORCE trick} \cite{Williams1992-xn}
--- as last resort due to high-variance gradients
(\textit{cf.}~\cite{Schulman2015-br, Schulman2016-wc}
for an in-depth dive into stochastic computational graphs).

By contrast, the objective resulting from the M-projection in \textsc{eq}~\ref{mprojectionargmin}
is burdened by none of the two previous hurdles. We only need to be able to sample from $\zeta$,
as opposed to having access to the likelihood $\zeta(a | s)$.
Like before, considering the particular case where $\zeta \coloneqq \beta$,
we do not even need access to a sampling unit, since we can directly use state-actions pairs picked from the
offline dataset $\mathcal{D}$. By defining $\Delta$ as an M-projection, one therefore reduces the
problem described in \textsc{eq}~\ref{piobjectivenew} (itself reduced from original one
\textit{cf.}~\textsc{eq}~\ref{piobjective}) into a strikingly simpler
problem (\textit{cf.}~\textsc{eq}~\ref{mprojectionargmin})
consisting in maximizing the \emph{re-weighted} likelihood of the actor's policy $\pi_\theta$
over the dataset $\mathcal{D}$.
Still, despite being comparatively tedious to estimate \emph{in practice},
the objective resulting from the I-projection in \textsc{eq}~\ref{iprojectionargmin} might be worth optimizing,
due to the greater resilience against out-of-distribution actions it invests the policy with, \textit{in theory}.

Among the past works that had to tackle the projection task in \textsc{eq}~\ref{piobjectivenew},
REPS \cite{Peters2010-vd},
RWR \cite{Peters2007-qb, Kober2010-hy},
LAWER \cite{Neumann2008-tm},
MPO \cite{Abdolmaleki2018-sp},
MARWIL \cite{Wang2018-dn},
AWR \cite{Peng2019-hu},
ABM \cite{Siegel2020-lo},
CRR \cite{Wang2020-sr},
and AWAC \cite{Nair2020-gd}
opted for a M-projection (forward KL), while VIP \cite{Neumann2011-hn}
chose to observe the problem through a variational inference lens and
went for an I-projection (reverse KL),
claiming that the cost-aversion induced via I-projections
alleviates plenty of issues that are attributed to M-projections
(\textit{cf.}~our discussion on these projections, earlier in \textsc{Section}~\ref{actorupdate}).
Nevertheless, the authors of \cite{Neumann2011-hn}
conclude that the reverse KL operation is considerably more difficult to compute,
which our previous discussion of the problem we arrived at in \textsc{eq}~\ref{iprojectionargmin}
corroborates.

In this work, we opt for the use of a \emph{forward} KL divergence, an M-projection, to define $\Delta$
in \textsc{eq}~\ref{piobjectivenew}. Indeed, we deem the trade-off to lean towards
computational feasibility and ease of implementation in modern settings,
despite the \emph{``reward-chasing''} behavior it can distill in the learned policies,
particularly destructive in offline RL.
Note, however, we still use a \emph{reverse} KL divergence
in the inequality constraint (\textit{cf.}~\textsc{eq}~\ref{piineq})
of the original optimization problem laid out in \textsc{eq}~\ref{piobjective},
in spite of using a \emph{forward} KL divergence in
the derived problem in \textsc{eq}~\ref{piobjectivenew}.
To sum up,
\textit{1)} we formulate a first problem (\textit{cf.}~\textsc{Eqs}~\ref{piobjectiveorig1},
\ref{piobjectiveorig2}, and \ref{piobjectiveorig3})
where the policy is urged to remain close to a proposal policy $\zeta$
in \emph{reverse} KL,
\textit{2)} we observe that the analytical closed-form solution of this constrained optimization problem
is the \emph{importance-weighted} counterpart $\zeta_\textsc{iw}$ of the proposal policy $\zeta$,
\textit{3)} we formulate a second problem (\textit{cf.}~\textsc{eq}~\ref{piobjectivenew})
where the policy is now urged to remain close to the
\emph{importance-weighted} proposal policy $\zeta_\textsc{iw}$
in \emph{forward} KL, and finally
\textit{4)} we observe that this second problem reduces to a final formulation that is simple, interpretable, and
light on compute.
\textit{In fine},
in practice, we update the actor's policy $\pi_\theta$ by minimizing
(via gradient descent) the loss $\ell_\theta$,
directly derived from \textsc{eq}~\ref{mprojectionargmin}:
\begin{align}
  \ell_\theta \coloneqq -
  \mathbb{E}_{s \sim \rho^\beta(\cdot), a \sim \zeta(\cdot | s)}
  \bigg[
  \exp (\frac{1}{\lambda_\textsc{kl}} A^{\pi_\theta}_\omega(s,a)) \log \pi_\theta(a | s)
  \bigg]
  \label{actorloss}
\end{align}

\subsection{Expansion to multiple distributions over actions}
\label{multisteams}
The actor's loss $\ell_\theta$ (\textit{cf.}~\textsc{eq}~\ref{actorloss})
involves the proposal distribution $\zeta$,
which we can define from any of the proposal policies we have laid out in \textsc{Section}~\ref{pe}, under
the same handle $\zeta$.
In particular, the past works
MARWIL \cite{Wang2018-dn},
AWR \cite{Peng2019-hu},
CRR \cite{Wang2020-sr},
and AWAC \cite{Nair2020-gd}
all update the actor's policy $\pi_\theta$ using $\ell_\theta$,
and using the offline policy $\beta$ as proposal policy $\zeta$, \textit{i.e.}~$\zeta \coloneqq \beta$
(\textit{cf.}~\ref{betasarsa} strategy in \textsc{Section}~\ref{proposalsimplex}).
As discussed in \textsc{Section}~\ref{projectionoptions}, when the proposal is $\beta$ the expectations
over state and action in $\ell_\theta$ are implemented in practice simply by taking samples from the offline
dataset $\mathcal{D}$, \textit{i.e.}~we need not have an explicit handle on $\zeta$, be it for sampling from $\zeta$
or for computing a likelihood estimate $\zeta(a | s)$ for a given state-action pair
$(s,a) \in \mathcal{S} \times \mathcal{A}$.
In addition to \emph{subsuming} the learning rules of
MARWIL \cite{Wang2018-dn},
AWR \cite{Peng2019-hu},
CRR \cite{Wang2020-sr},
and AWAC \cite{Nair2020-gd},
$\ell_\theta$ also subsumes \emph{both} policy improvement rules proposed in
ABM \cite{Siegel2020-lo},
where the proposal policy $\zeta$ plays the role of \emph{prior} (\textit{cf.}~ABM \cite{Siegel2020-lo}).

Importantly, as a design choice, we \emph{never} allow gradients to flow backwards through the $\zeta$ sampling unit
when the parameter vector $\theta$, parametrizing the actor's policy $\pi_\theta$, are used in the assembly of
the proposal policy $\zeta$ (if $\zeta$ does not use $\theta$, the $\zeta$ sampling unit
\textit{``$a \sim \zeta(\cdot | s)$''} is out of the computational graph for the actor update anyway).
We consequently need not involve stochastic computational graphs techniques, of which we gave an overview
earlier in \textsc{Section}~\ref{projectionoptions} when analyzing \textsc{eq}~\ref{iprojectionargmin}.
In practice, this means treating the actions sampled via $a \sim \zeta(\cdot | s)$ as \textit{inputs}
to the computational graph of the policy improvement update, or to \textit{detach} these samples from the graph.
We now consider this as a given and
will not involve \textit{stop-gradient} operations in the derivations that follow, whatever $\zeta$ contains.

Coming back to how $\ell_\theta$ subsumes the actor update
of ABM \cite{Siegel2020-lo},
we can replicate the one using the \textit{``BM''} prior (\textit{cf.}~\cite{Siegel2020-lo})
by setting $\zeta \coloneqq \beta_\textsc{c}$
(\textit{cf.}~\ref{betaclone} strategy in \textsc{Section}~\ref{proposalsimplex}),
where $\beta_\textsc{c}$ is a policy resulting from cloning the behavior policy $\beta$
underlying the dataset $\mathcal{D}$.
Furthermore, we can replicate the actor update rule using the \textit{``ABM''} prior
(\textit{cf.}~\cite{Siegel2020-lo})
by modelling $\zeta$ with an \emph{auxiliary} actor learned with
an $n$-step TD return \cite{Peng1996-xn}
hybrid between MARWIL/AWR (\cite{Wang2018-dn, Peng2019-hu}, pure MC return)
and CRR/AWAC (\cite{Wang2020-sr, Nair2020-gd}, $1$-step TD return).
In such a setting, the auxiliary actor would also use an auxiliary critic, learned via Monte-Carlo estimation,
in order to build its own advantage estimate, exclusively used by the \textit{``ABM''} prior.

Moreover, we observe that by aligning $\zeta$ with the actor itself
(\textit{cf.}~\ref{theta} strategy in \textsc{Section}~\ref{proposalsimplex})
--- more accurately, with a fixed, detached from the graph, copy of the previous actor update,
which can be denoted by $\pi_{\theta^\text{old}}$, akin to the notations adopted in TRPO \cite{Schulman2015-jt}
--- the inequality constraint depicted in \textsc{eq}~\ref{piineq} becomes
$\mathbb{E}_{s \sim \rho^\beta(\cdot)}\big[
D^{\pi_{\theta^\text{old}}}_{\overleftarrow{\textsc{kl}}}[\pi_\theta](s)\big]
= \mathbb{E}_{s \sim \rho^\beta(\cdot)}\big[
D_{\textsc{kl}}\big(\pi_\theta (\cdot | s) \, || \, \pi_{\theta^\text{old}} (\cdot | s)\big)\big]$
when applied to $\pi_\theta$.
This constraint coincide with the one adopted in MPO \cite{Abdolmaleki2018-sp},
and had it been a \textit{forward} KL instead of the reverse one,
this constraint would have matched the one used by
TRPO \cite{Schulman2015-jt},
and PPO \cite{Schulman2017-ou}:
$\mathbb{E}_{s \sim \rho^\beta(\cdot)}\big[
D^{\pi_{\theta^\text{old}}}_{\overrightarrow{\textsc{kl}}}[\pi_\theta](s)\big]
= \mathbb{E}_{s \sim \rho^\beta(\cdot)}\big[
D_{\textsc{kl}}\big(\pi_{\theta^\text{old}} (\cdot | s) \, || \, \pi_\theta (\cdot | s)\big)\big]$.
In essence, once one can instantiate the problem laid out in TRPO from one's framework, one can also do so
--- omitting minor irrelevant specificities --- for all the methods adopting a natural gradient
\cite{Amari2016-hl, Kakade2001-hw, Peters2008-mw}
approach, from which TRPO is inspired, such as
NPG \cite{Kakade2001-hw}, and
CPI \cite{Kakade2002-kw}.
All in all, the loss $\ell_\theta$ depicted in \textsc{eq}~\ref{actorloss} \textit{already} enables
us to instantiate a number of methods from the online and offline RL literature.
As such, the loss $\ell_\theta$ is not novel \emph{per se}, but the crafted framework
provides a unified view of the current state-of-the-art methods in offline RL
(CRR \cite{Wang2020-sr},
and AWAC \cite{Nair2020-gd}),
that are readily expressible under the framework for policy improvement we propose in this section.
Note, we established a similar unification earlier in \textsc{Section}~\ref{pe},
but over a wide range of distinct ways one could perform policy evaluation back then.

What the diagrams of \textsc{Figure}~\ref{valuesimplex} illustrated clearly in the context
of policy evaluation is that designing an
update rule (equivalently, loss function) for the action-value $Q_\omega$
is not a one-dimensional problem in offline RL like it is in online RL.
Re-using the nomenclature introduced in \textsc{Section}~\ref{pe}, the practitioner in charge of designing the
policy evaluation learning update for $Q_\omega$ has by construction tight control over where the learned
action-value $Q_\omega$ is located --- and how it will evolve and and travel --- over the value simplex
depicted in \textsc{Figure}~\ref{valuesimplex}.
While in online RL (at least in the traditional setting), said practitioner could design a critic's loss that
places $Q_\omega$ \emph{anywhere} on the closed line segment joining the
\textit{SARSA} critic $Q^{\pi_\theta}$ (perfectly consistent with $\pi_\theta$)
and the optimal critic $Q^*$ (perfectly consistent with $\pi^*$, and called
\textit{``expert''} instead of critic by \cite{Lim2018-ey}
to further emphasize the gap in their objectives).
As such, the DDPG \cite{Lillicrap2016-xa}
critic (among many others like SAC \cite{Haarnoja2018-bm}, etc.)
is effectively updated as a \textit{SARSA} critic where the next action injected in $\ell^\text{SARSA}_\omega$
(\textit{cf.}~\textsc{eq}~\ref{sarsaloss}) is from the (greedy) actor $\pi_\theta$,
while the actor-critic methods that stemmed from \cite{Crites1995-hn}
attempt for $Q_\omega$ to approximate $Q^*$ more directly.
Hence, one can easily place the action-value $Q_\omega$, for either summoned algorithm,
on the closed line segment joining $Q^{\pi_\theta}$ and $Q^*$ on
the value simplex of \textsc{Figure}~\ref{valuesimplex}.
The offline RL setting introduces $Q^\beta$
due to the added constraint discouraging the actor from straying from $\beta$.
The involvement of $\beta$, underlying the offline dataset $\mathcal{D}$, has the effect of
inflating the previous 1-dimensional closed line segment into said 2-dimensional simplex
(\textit{cf.}~\textsc{Figure}~\ref{valuesimplex}).

\begin{figure}[!h]
  \center\scalebox{0.3}[0.3]{\includegraphics{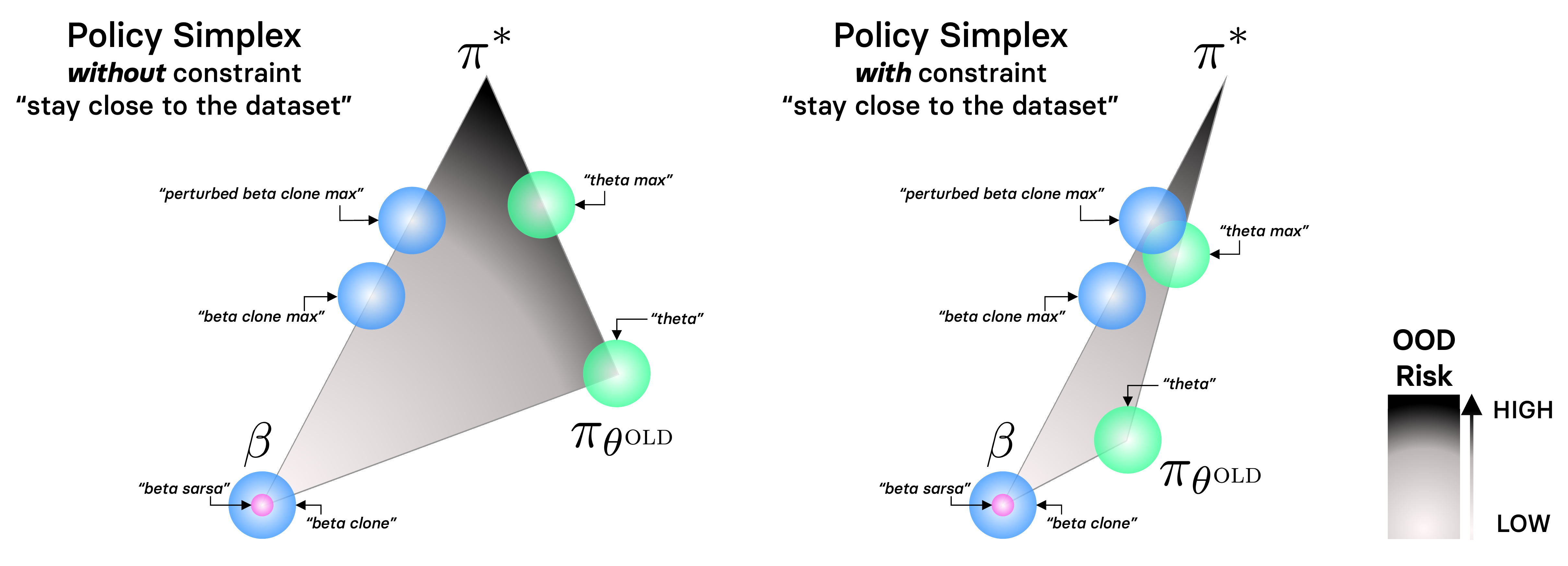}}
  \caption{Abstract representation of the relative positioning of the policies learned
  using the various proposal distributions $\zeta$ laid out in \textsc{Section}~\ref{pe}
  (whose names are depicted on the diagram)
  to sample actions from in $\ell_\theta$ (\textit{cf.}~\textsc{eq}~\ref{actorloss}).
  These policies are depicted by disks over the simplex spanned by the optimal policy $\pi^*$, the policy
  followed by the learned actor $\pi_\theta$, and the policy underlying the offline dataset, $\beta$.
  The diameter of said disks crudely depicts how confident one can be about the placement of the various
  tackled proposal policies (\textit{cf.}~\textsc{Section}~\ref{pe}) on the abstract simplex.
  Albeit only roughly estimating the actual geometry of the policy simplex,
  this diagram can nevertheless help us categorize the
  different proposal distributions with respect to how they expose to agent (and its value) to
  \emph{out-of-distribution} (OOD) actions \emph{at training time}, and crucially \emph{at evaluation time}.
  Note, we only consider the use of a \emph{single} proposal policy
  in $\ell^\textsc{giwr}_\theta$ (\textit{cf.}~\textsc{eq}~\ref{giwrloss})
  for the abstract, illustrative purposes of these diagrams.
  Best seen in color:
  \textit{pink} signifies that the proposal distribution is $\beta$ \emph{exactly},
  \textit{blue} that the proposal distribution relies on an
  estimate of the $\beta$ distribution, and \textit{green} that the proposal policy $\zeta$
  solely involves the actor $\pi_\theta$ --- in other words, the proposal distribution
  $\zeta$ is not derived from the offline dataset $\mathcal{D}$
  in any shape or form.
  We keep the residual denomination \textit{``SARSA''} from \textsc{Section}~\ref{pe},
  \textsc{eq}~\ref{betasarsa} to signify that $\zeta \coloneqq \beta$, for the sake of
  conceptual symmetry between \emph{evaluation} (\textit{cf.}~\textsc{Section}~\ref{pe})
  and \emph{improvement} (\textit{cf.}~\textsc{Section}~\ref{pi})
  --- despite the fact that the \textit{next} action $a'$ (last \textit{``A''} in \textit{``SARSA''})
  plays no functional role in the  policy improvement step.}
  \label{policysimplex}
\end{figure}

Constraining $\pi_\theta$ to remain close to $\beta$ in reverse KL divergence
as encoded by \textsc{eq}~\ref{piineq}, albeit instrumental in alleviating out-of-distribution actions
at training \emph{and} evaluation time, can (as a side effect) thwart the \emph{true} objective
the actor should aim at: converging towards $\pi^*$ for the task at hand.
Such reasoning is vividly echoing the discussion we carried out in \textsc{Section}~\ref{pe},
which we provided a retake of and pointers to in the previous paragraph.
Similarly to how the design of the policy evaluation step in offline RL makes $Q_\omega$ follow a certain path
(throughout the iterations) on a \textit{value} simplex whose vertices are
$\{
Q^*, Q^\beta, Q^{\pi_\theta}
\}$ (\textit{cf.}~\textsc{Figure}~\ref{valuesimplex}),
the design of the policy \emph{improvement} step in offline RL makes $\pi_\theta$ follow a certain path
(throughout the iterations) on a \textit{policy} simplex whose vertices are
$\{
\pi^*, \beta, \pi_{\theta^\text{old}}
\}$ (\textit{cf.}~\textsc{Figure}~\ref{policysimplex}).
Crucially, note, these simplices are asymmetrical: their verticies are not tied by a bijection,
\textit{i.e.}~there is not a one-to-one mapping linking each vertex of one simplex to its counterpart in the other.
Indeed, while the two verticies $\pi^*$ and $\beta$ are both coupled with their counterparts $Q^*$ and
$Q^\beta$ respectively by a \textit{``greedifies $\leftrightarrow$ evaluates''} relationship,
this is not at all the case for $\pi_{\theta^\text{old}}$ and $Q^{\pi_\theta}$.
This is due to the fact that there is an extra degree of estimation for the action-value compared to the policy.
While the estimated policy is $\pi_\theta$, the estimated action-value is $Q_\omega$, which is not necessarily
designed to be consistent with the estimated actor's policy $\pi_\theta$.

The result $Q_\omega \approx Q^{\pi_\theta}$ can be achieved \emph{only}
if the \ref{theta} strategy (proposed in \textsc{Section}~\ref{proposalsimplex})
is picked for policy evaluation.
The action-value ($Q^{\pi_\theta}$) perfectly consistent with the actor's policy ($\pi_\theta$)
that navigates the policy simplex (\textit{cf.}~\textsc{Figure}~\ref{policysimplex})
is a \emph{vertex} in the value simplex (\textit{cf.}~\textsc{Figure}~\ref{valuesimplex}),
and it is $Q_\omega$ that navigates the value simplex.
The policy simplex in \textsc{Figure}~\ref{policysimplex}
is simpler to interpret than the value simplex in the sense that the potential
proximity constraint imposed between $\pi_\theta$ and $\beta$ is directly observable since they both live in the
same space as the simplex.
This is not the case for the value simplex in \textsc{Figure}~\ref{valuesimplex}, for which the entities tied by
said proximity constraints (policies) do not live in the same space as the points of the simplex (action-values).

By aligning $\zeta$ with $\beta$ (\textit{cf.}~\ref{betasarsa} proposal strategy)
in \textsc{eq}~\ref{piineq} for the policy improvement step,
the actor update will \emph{attract} $\pi_\theta$ towards
the ``$\beta$'' corner of the simplex in \textsc{Figure}~\ref{policysimplex},
as it departs from $\pi_{\theta^\text{old}}$ and
makes a gradient step into the simplex.
On the next policy improvement step, the $\pi_{\theta^\text{old}}$ vertex will see its location overridden with
the freshly obtained $\pi_\theta$.
Note, like in the value simplex, the vertex involving
the parameter vector $\theta$ in the policy simplex
changes \emph{continually} as the agent iterates through the GPI steps, depicted in
\textsc{Figure}~\ref{gpidiag}.
Similarly, by setting $\zeta$ to be $\pi_{\theta^\text{old}}$
(\textit{cf.}~\ref{theta} proposal strategy)
in \textsc{eq}~\ref{piineq},
the actor update will attract $\pi_\theta$ towards
the ``$\pi_{\theta^\text{old}}$'' corner of the simplex
as it departs from $\pi_{\theta^\text{old}}$ and
makes a gradient step into the simplex,
effectively restricting the amplitude of updates $\pi_\theta$ goes through
--- as mentioned earlier when describing how our framework can implement the conservative KL constraints
akin to natural gradient \cite{Amari2016-hl, Kakade2001-hw, Peters2008-mw}
methods
from TRPO \cite{Schulman2015-jt},
PPO \cite{Schulman2017-ou},
and MPO \cite{Abdolmaleki2018-sp},
but here set in the context of \textsc{Figure}~\ref{policysimplex}.

Crucially, in this work, we want the learned actor policy $\pi_\theta$ to \emph{cover more ground} on
the policy simplex, similarly to our coverage of the value simplex in \textsc{Section}~\ref{pe} where we did so
by considering a slew of proposal policies $\zeta$ and using these to generate the \textit{next} action
to inject in Bellman's equation (\textit{cf.}~$\ell_\omega$ in \textsc{eq}~\ref{criticloss}).
We want to expand the navigation capabilities of $\pi_\theta$ over the policy simplex, and can do so
(as we have just laid out earlier in this paragraph) by using various designs of the proposal distribution
$\zeta$ in the inequality constraint (\textit{cf.}~\textsc{eq}~\ref{piineq}) at the source of the
policy improvement update rule we have just derived in \textsc{Section}~\ref{actorupdate}.
In \textsc{Section}~\ref{pe}, it is tedious to controllably navigate $Q_\omega$ over the value simplex
since the only entity we have control over with the proposal distribution $\zeta$ is the next action
$a' \sim \zeta(\cdot | s')$, to use in $\ell_\omega$.
We were able to interpolate between proposal policies by introducing \textit{SPI}-inspired designs
(\textit{cf.}~\textsc{Section}~\ref{proposalsimplex}),
but these can also only be used to output the next action for Bellman's equation, thwarting the
interpolation of proposal strategies by limiting their expressiveness.
Besides, making an action-value consistent with several proposal policies by optimizing a
combination of temporal-difference losses is a tall order, as it promises to be remarkably unstable.
Learning separate values in an ensemble could be an option, but is costly, and out of the scope of this work
(\textit{cf.}~\textsc{Section}~\ref{bg}).
By contrast, involving several proposal strategies
in our policy improvement objective $\ell_\theta$ in \textsc{eq}~\ref{actorloss}
seems significantly easier to optimize,
as it essentially \emph{augments} the dataset over which $\pi_\theta$ must maximize its
importance-weighted likelihood.
Such data augmentation, rather that trying to find a better representation to favor the resolution of
downstream tasks, aims to assist the agent towards a speedier resolution of the task currently at hand
by squeezing more juice out of the available information --- $\mathcal{D}$ in our case
(notable instances of such data augmentation include the works of
\cite{Kaelbling1993-dv} and \cite{Andrychowicz2017-mn}).

As such, we introduce a new framework, called GIWR (pronounced \textit{``giver''}) for
\textit{Generalized Importance-Weighted Regression}, whose defining objective
$\ell^\textsc{giwr}_\theta$
involves \emph{families} of proposal policies $Z_n \coloneqq (\zeta_i)_{i \in [1,n] \cap \mathbb{N}}$,
and their accompanying scaling coefficients $K_n \coloneqq (\kappa_i)_{i \in [1,n] \cap \mathbb{N}}$
with $\kappa_i > 0 \; (\forall i \in [1,n] \cap \mathbb{N})$.
We define $\ell^\textsc{giwr}_\theta$ as follows:
\begin{align}
  \ell^\textsc{giwr}_\theta \coloneqq -
  \mathbb{E}_{s \sim \rho^\beta(\cdot)}
  \Bigg[
  \sum^{n}_{i=1}
  \; \kappa_i \,
  \mathbb{E}_{a \sim \zeta_i(\cdot | s)}
  \bigg[
  \exp (\frac{1}{\lambda_\textsc{kl}} A^{\pi_\theta}_\omega(s,a)) \log \pi_\theta(a | s)
  \bigg]
  \Bigg]
  \label{giwrloss}
\end{align}
where the temperature hyper-parameter $\lambda_\textsc{kl}$ --- shared by all the $n$ contributions
to the GIWR actor loss $\ell^\textsc{giwr}_\theta$ defined in \textsc{eq}~\ref{giwrloss} ---
could be made specific per proposal policy $\zeta \in Z_n$.
We would then also have a family of $n$ temperatures, one for each $\zeta$ in $Z_n$,
making the hyper-parameter sweep considerably more tedious to complete.
We therefore opted for simplicity and sticked with the use a single, shared temperature $\lambda_\textsc{kl}$.
Since we conceived the loss in \textsc{eq}~\ref{giwrloss} as a \emph{multi-objective}
inflation of the loss in \textsc{eq}~\ref{actorloss},
we can interpret GIWR as introducing extra KL inequality constraints
following the schema of \textsc{eq}~\ref{piineq}, restricting
$\mathbb{E}_{s \sim \rho^\beta(\cdot)}\big[
D^{\zeta_k}_{\overleftarrow{\textsc{kl}}}[\pi_\theta](s)
\big]
=
\mathbb{E}_{s \sim \rho^\beta(\cdot)}\big[
D_{\textsc{kl}}\big(\pi_\theta (\cdot | s) \, || \, \zeta_k (\cdot | s)\big)
\big]$
to remain below a certain threshold,
for the $n$ proposal distributions $\zeta_k$
of the proposal family $Z_n$.
We refer the reader to \cite{Roijers2014-rt}
and \cite{Mossalam2016-aj}
for a survey and overview of the \textit{multi-objective RL} sub-field.

In particular, the work of \cite{Abdolmaleki2020-gu},
dresses the proposed MO-MPO framework as a multi-objective RL framework first and foremost,
but is in essence a multi-task RL framework that tackles the tasks at end via a multi-objective formulation,
and learns a single policy that must trade off across different \textit{tasks}, or interchangeably,
\textit{objectives} (\textit{cf.}~\textsc{Section 2.3} of their work \cite{Abdolmaleki2020-gu}).
They build on the premise that the agent is provided with a family of reward signals, a distinct one for each task
or objective, and learn an action-value for each.
They learn an action distribution (using our terminology, a proposal distribution) for each of these action-values,
and combine these distributions along with their associated task-specific values to obtain the next actor iterate.
Our proposed objective draws similarities with theirs, as they build on MPO \cite{Abdolmaleki2018-sp}
which we showed earlier can be instantiated under our current framework and shares the derivation re-purposed in
\textsc{Section}~\ref{actorupdate} like most of the approaches adopting the \textit{``RL as inference''}
paradigm. In this work, by contrast, the agent does not have access to a family of rewards
(world framed as a \textit{multi-objective MDP} in \cite{Abdolmaleki2020-gu}), but only
to the rewards collected by $\beta$ and
provided through the offline dataset $\mathcal{D}$
(\textit{cf.} the work of \cite{Fujimoto2018-mj} and \cite{Laroche2019-ar} for a formulation of \textit{offline}
or \textit{dataset-bound} MDP
underlying $\mathcal{D}$ and derived from the dynamics traces observed in $\mathcal{D}$).
As such, we only learn a single action-value.
Plus, despite also involving a proposal family (denoted by $Z_n$ in our work), the proposal distributions
are defined in a completely different way: while \cite{Abdolmaleki2020-gu}
has one per reward signal, we conceive the $\zeta$'s in $Z_n$ from the bare information available in
the offline setting (just $\mathcal{D}$), as strategies that empower the agent to cover the
policy and value simplicies
(\textit{cf.}~\textsc{Figures}~\ref{policysimplex} and \ref{valuesimplex})
to a greater extent, so as to achieve optimality faster and more reliably
while avoiding the pitfalls of offline RL.
Since the GIWR framework allows for the involvement of as many constraints one desires,
we can essentially instantiate both a trust-region constraint tying the next iterate to the previous one
$\pi_{\theta^\text{old}}$ and another constraint forcing it to be in the vicinity of
$\beta$ in the policy simplex (\textit{cf.}~\textsc{Figure}~\ref{policysimplex}).
As such, with a proposal family of two defined as $Z = (\zeta_\textsc{awr}, \pi_{\theta^\text{old}})$,
where $\zeta_\textsc{awr}$ is an \textit{auxiliary actor} learned with
$n$-step TD extension \cite{Peng1996-xn}
of the AWR \cite{Peng2019-hu} algorithm,
we can effectively replicate the variant of ABM \cite{Siegel2020-lo}, called ABM-MPO,
reported as achieving the highest performance in said work.
Alternatively, by replacing $\zeta_\textsc{awr}$ with $\mathcal{T}_\textsc{Eval}\big[\beta_\textsc{c}\big]$
(\textit{cf.}~\ref{betaclone} in \textsc{Section}~\ref{pe}) in the proposal family $Z$,
we get the empirically-weaker BM-MPO method, as reported in \cite{Siegel2020-lo}.
Nonetheless, these two last methods are reported in \cite{Nair2020-gd}
to be outperformed by AWAC \cite{Nair2020-gd}
--- and consequently also by the concurrent, virtually-identical CRR method \cite{Wang2020-sr}.

These two last methods (along with MARWIL \cite{Wang2018-dn} and AWR \cite{Peng2019-hu}
if we set aside how they estimate $Q_\omega$)
can be cast as instances of GIWR (\textit{cf.}~$\ell^\textsc{giwr}_\theta$ in \textsc{eq}~\ref{giwrloss})
where the proposal family $Z$ is a \emph{singleton} that contains only $\beta$
(and the scaling coefficient family $K$ a singleton that trivially contains only $1.0$).
These achieve state-of-the-art performance in most situations --- dynamics of the environment and
quality of the dataset loosely being the main differentiating factors,
as we have showcased in \textsc{Section}~\ref{experimentalassessment}.
As such, in the experiments we report in this work (\textit{cf.}~\textsc{Section}~\ref{piresults}),
every proposal family $Z$ that we consider contains $\beta$,
the proposal distribution \ref{betasarsa}. Not only has this proposal distribution proved to yield excellent
performance in AWAC \cite{Nair2020-gd} and CRR \cite{Wang2020-sr},
but it is also trivial to evaluate the expectation with respect to such proposal, as discussed earlier
in \textsc{Section}~\ref{actorupdate}, since we need only take data from the offline dataset
$\mathcal{D}$.
To keep the number of experiments to a reasonable amount without sacrificing the depth
of understanding we can get out out of them, we cap the cardinality of $Z$ at $2$.
Since we set $|Z| \leq 2$ and $\beta \in Z$ in the experiments of \textsc{Section}~\ref{piresults},
we can then use ``$\zeta$'' to unambiguously denote the \textit{other} proposal distribution
(distinct from $\beta$) in the family $Z$ when $|Z| = 2$.
Likewise,
we can then use ``$\kappa$'' to unambiguously denote the coefficient that scales the
contribution associated with $\zeta$ in $\ell^\textsc{giwr}_\theta$ (\textit{cf.}~\textsc{eq}~\ref{giwrloss}),
since we scale the contribution associated with $\beta$ in $\ell^\textsc{giwr}_\theta$ by $1.0$
consistently across experiments.
Concretely, as for policy evaluation in \textsc{Section}~\ref{pe}, we report and discuss our empirical findings
for 9 scenarios. The first corresponds to the case where only $\beta$ is used ($|Z| = 1$, coincides exactly
with AWAC \cite{Nair2020-gd} and CRR \cite{Wang2020-sr})
and plays the role of \textbf{baseline} in \textsc{Section}~\ref{piresults}.
The 8 other competing scenarios correspond to the cases where $Z = (\beta, \zeta)$
and $K = (1.0, \kappa)$, with
$\zeta$ covering the spectrum of proposal distributions that we introduced in \textsc{Section}~\ref{pe}
(and \textsc{Appendix}~\ref{spiproposalsimplex})
--- all except \ref{betasarsa}, which would be redundant with the baseline, involving only $\beta$.
This brings us to a total of 8 candidates for $\zeta$:
\ref{theta}, \ref{thetamax}
\ref{betaclone}, \ref{betaclonemax}, \ref{perturbedbetaclonemax}
(and the three SPI proposal distributions from \textsc{Appendix}~\ref{spiproposalsimplex}).
These scenarios were designed to maintain a high degree of symmetry with the previously reported experiments;
we could use \emph{any} policy for $\zeta$.
\textbf{Note, every method competing with the baseline is novel.}

We lay out the pseudo-code for \textsc{Section}~\ref{piresults}
in \textsc{Algorithm}~\ref{algopi}.

We now report and discuss our experimental findings.

\IncMargin{1em}
\begin{algorithm}
\SetAlgoLined
\SetNlSty{}{}{}
\SetKwInput{KwInit}{init}
\KwInit{initialize
the random seeds of each framework used for sampling,
the random seed of the environment $\mathbb{M}$,
the neural function approximators' parameters
($\theta$ for the actor's policy $\pi_\theta$, and $\omega$ for the critic's action-value $Q_\omega$),
the critic's target network $\omega'$ as an exact frozen copy,
the offline dataset $\mathcal{D}$.}
\While{no stopping criterion is met}{
    \tcc{Train the agent in $\mathbb{M}^\textsc{off}$}
    Get a mini-batch of samples from the offline dataset $\mathcal{D}$\;
    Perform a gradient \emph{descent} step along
        $\nabla_\omega \, \ell_\omega$
        (\textit{cf.} below)
        using the mini-batch\;
    \SetNlSty{}{\textcolor{red}{$\diamond$}}{}
    $$
    \ell_\omega \coloneqq
    \mathbb{E}_{s \sim \rho^\beta(\cdot), a \sim \beta(\cdot | s), s' \sim \rho^\beta(\cdot)}
    \bigg[
    \Big(
    Q_\omega(s,a) -
    \big(
    r(s, a, s') + \gamma \, \mathbb{E}_{a' \sim \textcolor{red}{\zeta^\textsc{pe}}(\cdot | s')}
    \big[
    Q_{\omega'}(s',a')
    \big]
    \big)
    \Big)^2
    \bigg]
    $$
    \SetNlSty{}{}{}
    where $r(s, a, s')$ was introduced as syntactic sugar in \textsc{Section}~\ref{bg}\;
    \textcolor{red}{(Note, as mentioned early in \textsc{Section}~\ref{actorupdate},
    we use $\zeta^\textsc{pe} \coloneqq \pi_\theta$ in the experiments reported in
    \textsc{Section}~\ref{piresults})}\;
    Perform a gradient \emph{ascent} step along
        $\nabla_\theta \, \mathcal{U}_\theta$
        (\textit{cf.} below)
        using the mini-batch\;
    \SetNlSty{}{\textcolor{red}{$\diamond$}}{}
    $$
    \mathcal{U}_\theta \coloneqq
    \mathbb{E}_{s \sim \rho^\beta(\cdot)}
    \Bigg[
    \textcolor{red}{
    \sum^{n}_{i=1}
    \; \kappa_i \,
    \mathbb{E}_{a \sim \zeta^\textsc{pi}_i(\cdot | s)}
    }
    \bigg[
    \exp (\frac{1}{\lambda_\textsc{kl}} A^{\pi_\theta}_\omega(s,a)) \log \pi_\theta(a | s)
    \bigg]
    \Bigg]
    $$
    \SetNlSty{}{}{}
    where
    $A^{\pi_\theta}_\omega(s,a) \coloneqq Q_\omega(s,a) -
    \mathbb{E}_{\bar{a} \sim \pi_\theta}[Q_\omega(s,\bar{a})]$,
    $\kappa_i$ are scaling coefficients,
    and $\lambda_\textsc{KL}$ is a temperature\;
    Update the target network $\omega'$ using the new $\omega$\;
    \tcc{Evaluate the agent in $\mathbb{M}$}
    \If{evaluation criterion is met}{
        \ForEach{evaluation step per iteration}{
            Evaluate the empirical return of $\pi_\theta$
                in $\mathbb{M}$
                (\textit{cf.} evaluation protocol in \textsc{Appendix}~\ref{evalprotocol})\;
        }
    }
}
\caption{GIWR \\
with proposal distributions:
$\zeta^\textsc{pe}$ for policy evaluation,
$(\zeta^\textsc{pi}_i)_{i \in [1,n] \cap \mathbb{N}}$ for policy improvement \\
\textcolor{red}{$\diamond$}: differing from \textsc{Base} (\textit{cf.}~\textsc{Algorithm}~\ref{algobase})}
\label{algopi}
\end{algorithm}
\DecMargin{1em}

\subsection{Experimental results}
\label{piresults}

\begin{figure}[!h]
  \center\scalebox{0.18}[0.18]{\includegraphics{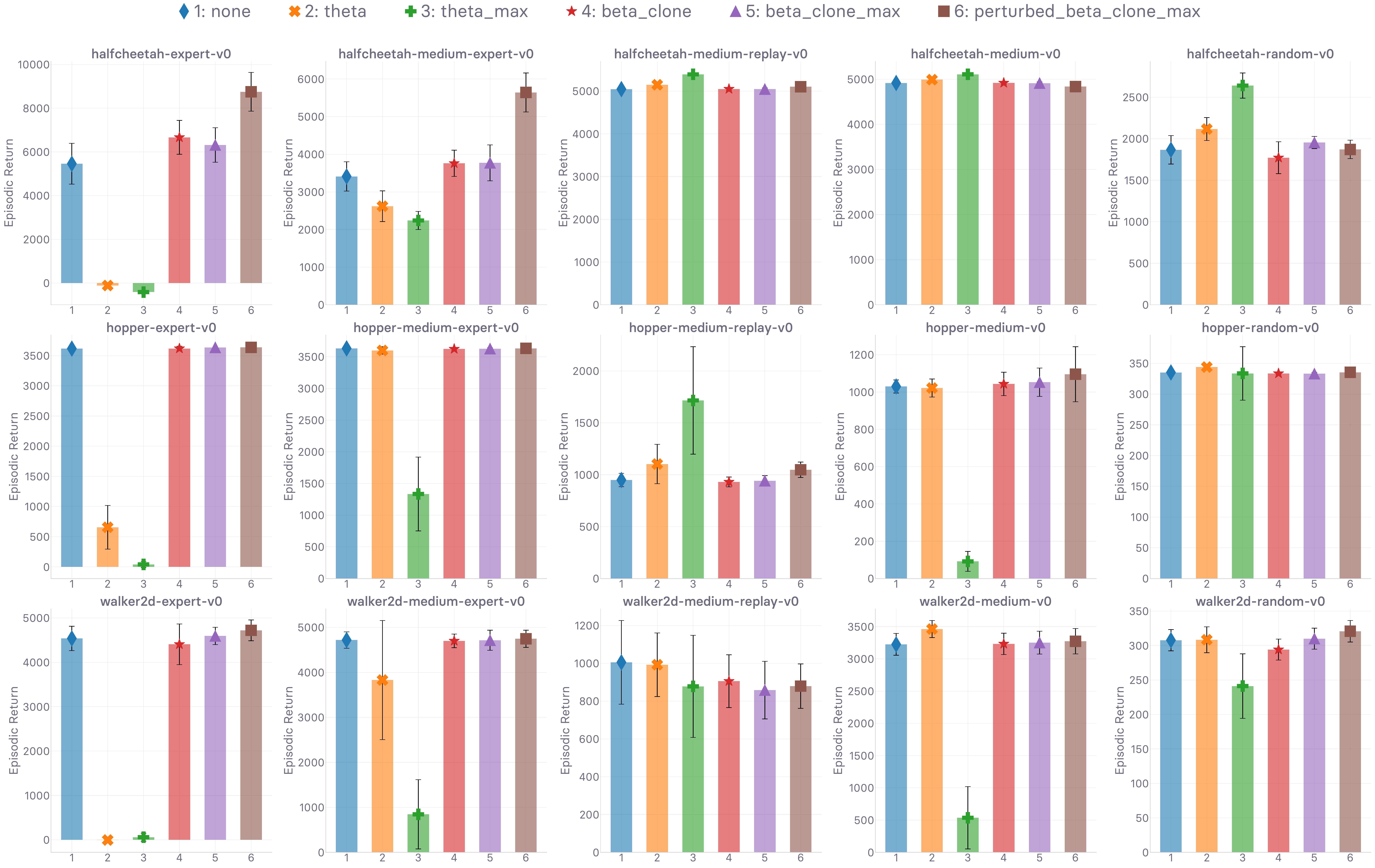}}
  \caption{Final performance of GIWR (\textit{cf.}~\textsc{Algorithm}~\ref{algopi})
  with the policy improvement carried out
  under the different proposal distributions that we introduced in \textsc{Section}~\ref{proposalsimplex}.
  Everything except the proposal policy $\zeta$ in use is identical.
  We use $\kappa=0.2$ as scaling coefficient for the contribution of $\zeta$
  in \textsc{eq}~\ref{giwrloss}.
  Runtime is 12 hours. Best seen in color.}
  \label{giwr02:barplot}
\end{figure}

Again, we rely on the experimental setting thoroughly described in \textsc{Appendix}~\ref{experimentalsetting}
to carry out the empirical investigation laid out here.
We remind the reader that, as specified at the beginning of \textsc{Section}~\ref{actorupdate},
we adopt the proposal distribution strategy \ref{theta} for policy \emph{evaluation} in the empirical investigations
performed in this section tackling policy improvement.
We first report the empirical evaluation of the experimental scenarios assembled and
laid out at the end of \textsc{Section}~\ref{multisteams}
in \textsc{Figure}~\ref{giwr02:barplot}.
For every experiment reported in \textsc{Figure}~\ref{giwr02:barplot}, we set the scaling coefficient
$\kappa$ to $0.2$, and we report the counterpart performances for $\kappa \in \{0.1, 0.5\}$ in
\textsc{Appendix}~\ref{giwrsweep}, \textsc{Figures}~\ref{giwr01:barplot} and \ref{giwr05:barplot} respectively.

\textsc{Figure}~\ref{giwr02:barplot} tells a story that echoes the one we laid out in \textsc{Section}~\ref{peresults}.
Similarly, the proposal distributions \ref{theta} and \ref{thetamax} perform very inconsistently across the
considered benchmark of datasets, overall being the two worst choices of proposal policices $\zeta$
--- in the setting we set out to work with in \textsc{Section}~\ref{multisteams},
\textit{i.e.}~$|Z| \leq 2$ and $\beta \in Z$.
We remind the reader that we gave thorough interpretations for every proposal distributions introduced
in \textsc{Section}~\ref{proposalsimplex} for policy evaluation,
as we re-purposed them in the context of policy \emph{improvement},
and refer the reader to these discussions.
In short, we observe in \textsc{Figure}~\ref{giwr02:barplot} that,
by encouraging the actor's policy $\pi_\theta$
to stay close to $\mathcal{T}_\textsc{Eval}[\pi_{\theta^\text{old}}]$ or
$\mathcal{T}_\textsc{Max}^{\omega, m}[\pi_{\theta^\text{old}}]$ in reverse KL divergence
(\textit{cf.}~\textsc{eq}~\ref{piineq} in \textsc{Section}~\ref{projectionoptions}),
in addition to remaining close to $\beta$ (\textit{cf.}~\ref{betasarsa}) in reverse KL since $\beta \in Z$
as per our experimental design choices, \ref{theta} and \ref{thetamax} (respectively)
yield terrible returns in most scenarios, while improving upon the baseline only in
rare isolated cases.
Note, we do not enforce these constraints \emph{explicitly}, as we optimize the \emph{reduction} that we derived
in \textsc{Section}~\ref{projectionoptions} and extended in \textsc{Section}~\ref{multisteams}.
Indeed, despite being essentially equivalent to AWAC \cite{Nair2020-gd} or CRR \cite{Wang2020-sr}
(\textit{blue} color in \textsc{Figure}~\ref{giwr02:barplot})
in which an extra \textit{MPO}-like trust-region constraint \cite{Abdolmaleki2018-sp} is plugged in,
seem not to be as conservative and \textit{safe} in terms of how the actor's policy navigates the simplex
depicted in \textsc{Figure}~\ref{policysimplex}
as one might expect.
The value of $\kappa$ might be the culprit here, and a deeper hyper-parameter sweep for $\kappa$ could be key
to strike the right trade-off of safety against destructive policy updates.
This hypothesis is to a certain extent corroborated by \textsc{Figure}~\ref{giwr01:barplot}
in \textsc{Appendix}~\ref{giwrsweep}, where the use of \ref{theta} in GIWR
seem to yield considerably better returns in environments where performance seemed to be disastrous
(\textit{e.g.} in the expert-grade datasets).
Lower values of $\kappa$ \emph{can} mitigate the dips in performance caused by destructively big updates
in parameter space, based on how much the returns drop for the \ref{theta} and \ref{thetamax}
proposal distributions in \textsc{Appendix}~\ref{giwrsweep} \textsc{Figure}~\ref{giwr05:barplot}
($\kappa=0.1$),
compared to in \textsc{Appendix}~\ref{giwrsweep} \textsc{Figure}~\ref{giwr01:barplot}
($\kappa=0.5$).
These two heuristics seem not to be worth introducing from an offline RL practitioner's standpoint,
judging by how sensitive (stiff, as we characterized earlier in \textsc{Section}~\ref{convenience})
these methods are \textit{w.r.t.} the value of $\kappa$,
especially in the expert-grade datasets.
Still, it was shocking to observe just how well \ref{thetamax} performs in the random dataset of the
\texttt{halfcheetah} environment within the tackled benchmark (top-right in \textsc{Figures}~\ref{giwr02:barplot},
\ref{giwr01:barplot}, and \ref{giwr05:barplot}).

We report the performance of GIWR when $\zeta$ is picked from the \textit{SPI} group
in \textsc{Figure}~\ref{giwr02appendix:barplot}
and discuss the these in \textsc{Appendix}~\ref{spipiresults}.
In short, as we have observed and reported earlier in \textsc{Section}~\ref{peresults},
the involvement of \ref{thetamax} in every distribution of the \textit{SPI} group
(\textit{cf.}~\textsc{Appendix}~\ref{spiproposalsimplex}) is too detrimental for
the safety these offer to be beneficial.

Finally, \textsc{Figures}~\ref{giwr02:barplot}, \ref{giwr01:barplot}, and \ref{giwr05:barplot}
show that the proposal policies in the \textit{clone} group
(\ref{betaclone}, \ref{betaclonemax}, and \ref{perturbedbetaclonemax})
are positively assisting the baseline methods in \emph{every} environment where it struggled in the first place
against the other state-of-the-art offline RL methods
(\textit{cf.}~\textsc{Section}~\ref{experimentalassessment}).
Importantly, these add-ons do \emph{not} harm the baseline while enhancing it in the environments where
it was lagging behind.
While \ref{betaclone}, \ref{betaclonemax} do not improve upon the baseline by a significant margin,
\ref{perturbedbetaclonemax} widens said margin to a greater extent across the benchmark,
and especially in the dataset-environment couples in which the baseline showed signs of struggle in
\textsc{Section}~\ref{experimentalassessment}.
\textit{In fine}, involving the perturbation model $\xi$ on top of a clone $\beta$
of the behavioral distribution $\beta_\textsc{c}$
\emph{and} leveraging the $\mathcal{T}_\textsc{Max}^{\omega', m}$ operator
to build the proposal distribution described in \textsc{eq}~\ref{xieq}
achieves the best results, by a large margin relative
to the other proposal strategies in the \textit{clone} family in the dataset and environments where
the baseline needs it most.
Note, this proposal distribution coincides with the actor learned in BCQ \cite{Fujimoto2018-mj}
--- yet, $Q_\omega$ is decoupled, \textit{cf.}~\textsc{Section}~\ref{proposalsimplex},
and \emph{not} learned as in BCQ \cite{Fujimoto2018-mj}, which would correspond to performing
policy evaluation with the proposal distribution \ref{perturbedbetaclonemax}.
We observe identical results in \textsc{Figures}~\ref{giwr01:barplot} and \ref{giwr05:barplot}
in \textsc{Appendix}~\ref{giwrsweep}.

\begin{figure}[!h]
  \center\scalebox{0.18}[0.18]{\includegraphics{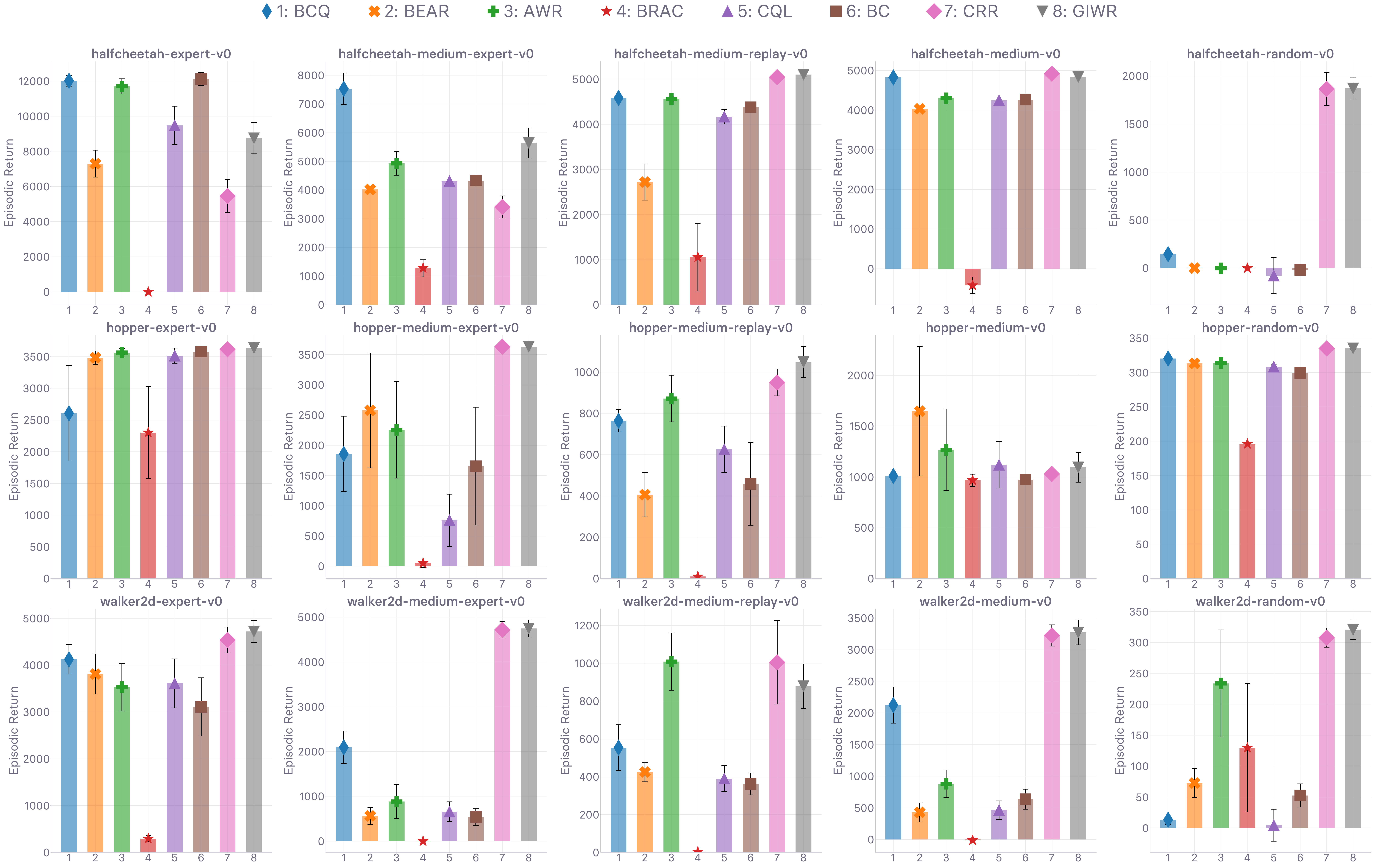}}
  \caption{Final performance comparison between the best performing studied variant of GIWR
  (\textit{cf.}~\textsc{Algorithm}~\ref{algopi}),
  the one using \ref{perturbedbetaclonemax} for $\zeta$,
  and the other state-of-the-art offline RL methods, introduced in \textsc{Section}~\ref{baselines},
  and compared empirically in \textsc{Section}~\ref{experimentalassessment}.
  We use $\kappa=0.2$ as scaling coefficient for the contribution of $\zeta$
  in \textsc{eq}~\ref{giwrloss}.
  Runtime is 12 hours. Best seen in color.}
  \label{giwrbaselines:barplot}
\end{figure}

We place the best performing studied variant of GIWR, the one using \ref{perturbedbetaclonemax} for $\zeta$,
among the other state-of-the-art offline RL baselines introduced in \textsc{Section}~\ref{baselines}
and compared empirically in \textsc{Section}~\ref{experimentalassessment},
with $\kappa = 0.2$, in \textsc{Figure}~\ref{giwrbaselines:barplot}.
We omit SAC \cite{Haarnoja2018-bm}
and our version of D4PG \cite{Barth-Maron2018-ot}
judging by how poorly they performed in the analysis we carried out
and laid out in \textsc{Section}~\ref{experimentalassessment}.
While \textsc{Figure}~\ref{giwrbaselines:barplot} does not provide new information \textit{per se},
it puts things in perspective as for how GIWR enables us to close the gap between
the chosen baseline (\textit{cf.}~\textsc{Algorithm}~\ref{algobase})
and its competition in the environments in which it lagged behind.
For instance, in the second plot of the grid in \textsc{Figure}~\ref{giwrbaselines:barplot}
(first row, second column), CRR displays the eighth highest return,
while GIWR achieves the second highest return --- behind BCQ \cite{Fujimoto2018-mj}
which underperforms both CRR and our GIWR instance in \emph{13 out of the 15} datasets of the suite.
Note, we could fall back to the next-in-line best performing model, AWR \cite{Peng2019-hu},
simply by appropriating $Q_\omega$ via Monte-Carlo estimation instead of temporal-difference learning,
for the $\beta \in Z$ component of the GIWR loss (\textit{cf.}~\textsc{eq}~\ref{giwrloss}),
or for every distribution of the proposal family $Z = (\beta,\zeta)$.
The obtained results confirm our intuition, laid out in \textsc{Section}~\ref{multisteams} and illustrated in
the simplices of
\textsc{Figure}~\ref{valuesimplex} and \textsc{Figure}~\ref{policysimplex}:
it \emph{can} be highly beneficial to directly urge $\pi_\theta$ to approach $\pi^*$, and by using \ref{theta}
as proposal distribution in policy evaluation as we do in this section,
to indirectly urge $Q_\omega$ to $Q^*$ --- essentially skewing the SARSA update
\cite{Rummery1994-qp, Thrun1995-sz, Sutton1996-ky, Van_Seijen2009-yw}
consisting in using \ref{theta} as proposal distribution in policy evaluation
into a Q-learning update \cite{Watkins1989-ir, Watkins1992-gl},
\emph{while} dealing with the hindrance of distributional shift in offline RL.

By far the most crucially appealing feature of our framework is that the practitioner \emph{need not} take any
risky decisions when it comes down to the design of the policy improvement rule, as we have shown that we can
increase the performance of the state-of-the-art offline RL baseline in specific datasets
(\textit{e.g.} in the expert ones)
without hurting its performance in the remainder of the suite.
This best-of-both-worlds trade-off has not been struck by any other method preceding GIWR,
as we have shown in \textsc{Section}~\ref{inductivebiases}.
\textbf{By leveraging implicit I-projections
while being heavily modular, GIWR enables the practitioner to
built policy improvement update rules that can fit their use cases while being safely shielded from
spurious inductive biases most baselines aggressively inject in their agents.}

\paragraph{N.B.}

The GPI scheme revisited in this work for offline RL (which we illustrated in
\textsc{Figure}~\ref{gpidiag})
entangles policy evaluation steps and policy improvement steps in an alternating,
iterative process:
one seemingly small disparity between evaluation (\textit{cf.}~\ref{pe})
or improvement (\textit{cf.}~\ref{pi})
update rules can cause considerable \textbf{ripple effects} on the whole compound procedure.
In this work, we have undertaken an in-depth and in-breadth revisitation of GPI
in the offline regime, which
further burdens GPI with additional points of failure such as distributional shift,
caused by the inability for the
agent to collect more training data as it tries to learn
an optimal policy (that is interactive by nature).

\paragraph{Guidelines}

We can therefore articulate the following guidelines for the choice of proposal distribution
in the policy improvement objective of GIWR (\textit{cf.}~\textsc{Algorithm}~\ref{algopi}).
\textit{1)} If the dataset grade is \emph{known}, use \ref{perturbedbetaclonemax} for near-expert-grade datasets,
and none otherwise (which reduces to the update rule of CRR and AWAC).
Aside from GIWR, one could also use BCQ or BC directly for near-expert-grade datasets.
\textit{2)} If the dataset grade is \emph{unknown}, the agnostic best choice is to
use GIWR with \ref{perturbedbetaclonemax} as proposal in the policy improvement objective,
As is substantially improve upon CRR/AWAC, the previously most versatile method,
across the spectrum.

\section{Conclusion}

We conlude this work by following the list of contributions laid out in \textsc{Section}~\ref{contriblist}.
Our \textbf{first contribution (\#1)}
consisted in the re-implementation of the main state-of-the-art offline RL
baselines under a fair, unified, highly factorized, and open-sourced framework and accompanying codebase.
We attribute the success and failure of these baselines over the spectrum of considered datasets and environments
to how \emph{biased} the agent is made (by the offline method)
towards positing the optimality of the policy underlying the provided
offline dataset for the given task.
Approaches that perform well on one end of the spectrum (\textit{e.g.} with expert-grade datasets)
typically achieve deterringly low returns on the other end of the spectrum (random-grade datasets),
and \textit{vice versa}.
Understandably, the hyper-parameters that control the bias injection are the hardest to tune,
across the entire range of methods.
We looked for the method that achieved the best over the spectrum of considered dataset qualities,
and therefore took a advantage-weighted regression template as base.
We first studied how this method --- well-behaved on the \textit{low}-quality end of the spectrum,
subpar relative to the competing baselines on the \textit{high}-quality end ---
reacted to the purposeful injection of optimality inductive bias, and how it impacted final performance.
Via a toy extension of the base method, we have shown just how brutally detrimental
the usual direct injection of bias can be on the achieved levels of return when the offline dataset
is \textit{sub}-optimal.
This empirical evidence constitutes the \textbf{second contribution (\#2)} of this work.
As our \textbf{third contribution (\#3)},
we have proposed generalizations of the policy evaluation and improvement steps for offline RL,
involving several proposal distributions over actions.
Through these generalizations, we effectively revisit the generalized policy iteration
scheme for the offline regime, setting out to understand how to design an offline RL method that
enables the agent to \textbf{close in on optimality, while remaining shielded from the
distributional shift hindering offline methods}.
Notably, in policy evaluation, we showed that
even a method as simple
as \textit{SARSA} with respect to the offline distribution yields surprisingly good and
robust results.
In policy improvement, we have proposed the GIWR framework which enables the practitioner to craft the objective that
suits the desired level of awareness about the quality of the dataset.
The closer to optimality, the more bias should be injected.
\textbf{Contrary to previous works (re-implemented and empirically compared as our first contribution)
and the toy extension studied purposely through the lens of inductive bias injection (second contribution),
we \emph{can} get gains on one end of the spectrum without hurting performance on the other end.}
We consistently highlight which proposals seem to perform best in evaluation and improvement
respectively, and advocate for their usage in practical scenarios since they enable
\textbf{improvements without compromise, despite not being aware of how sub-optimal
the offline distribution actually is}.

The availibility of privileged information about the quality of the dataset
telling the agent about how close to being optimal the underlying policy is,
is paramount to design the best suited offline learning method for the RL agent.
The involvement of \textbf{data labelers} in the offline learning stack
can bridge the gap and make such information available to the agent, and have a considerable impact
in safety-critical systems such as autonomous driving and healthcare.

\bibliography{bibliography}

\clearpage

\appendix

\section{Comprehensive exposition of the experimental setting}
\label{experimentalsetting}

\subsection{Offline RL baselines: re-implementation details
and fairness-conscious design choices}
\label{baselinesdetails}

The SAC \cite{Haarnoja2018-bm} and
D4PG \cite{Barth-Maron2018-ot}
algorithms were originally introduced as online RL algorithms, and we simply repurposed them as offline RL ones
by \textit{a)} removing the agent's ability to interact, collect, and store \textit{new} data, and
\textit{b)} leaving the training loop unaltered, yet replacing the replay buffer $\mathcal{R}$
with the offline dataset $\mathcal{D}$.
This candid transition from the online to the offline regime
for SAC \cite{Haarnoja2018-bm} and
D4PG \cite{Barth-Maron2018-ot}
has proved disastrous numerous times, as reported in
some of the candidate baselines above
(SAC \cite{Haarnoja2018-bm} has been reported the most;
D4PG only rarely, despite the promising potential displayed by distributional values in \cite{Agarwal2020-eu}).
Behavioral Cloning (BC) \cite{Pomerleau1989-nh, Pomerleau1990-lm, Ratliff2007-fc, Bagnell2015-ni}
is the only imitation learning \cite{Bagnell2015-ni}
method of the above listing, and only uses the state-action pair $(s,a)$ of transitions pulled from the
offline dataset $\mathcal{D}$ to learn $\pi_\theta$ as a supervised learning regressor
(states $s$ are the inputs; actions $a$ are the real-values outputs).
Such method is not equiped to leverage the reward information communicated through the offline dataset $\mathcal{D}$,
or equivalently via the observable, fictitious MDP $\mathbb{M}^\textsc{off}$.
State-of-the-art imitation learning techniques that \emph{are} able to do so (\textit{e.g.} adversarial approaches
like GAIL \cite{Ho2016-bv}, SAM \cite{Blonde2019-vc}, and DAC \cite{Kostrikov2019-jo})
require the agent to interact with $\mathbb{M}$ to collect more data to update their \textit{surrogate}
reward proxy. Since only $\mathbb{M}^\textsc{off}$ is accessible at training time,
in the offline RL regime, such methods can not be used here.
In addition, since AWAC \cite{Nair2020-gd}
was released concurrently with CRR \cite{Wang2020-sr}
and are essentially equivalent (\textit{cf.}~statement from the authors in \cite{Nair2020-gd}),
we use the \textit{``CRR''} notation to denote either indifferently.
In line with the results reported in \cite{Wang2020-sr} and \cite{Nair2020-gd},
showing that ABM \cite{Siegel2020-lo}
is consistently outperformed by their respective approaches (CRR and AWAC respectively),
we save valuable resources by not including ABM in the list of baselines.
Besides, \textit{a)} ABM's optimization shares its derivation with AWAC and CRR
(\textit{cf.}~\textsc{Section}~\ref{pi}),
\textit{b)} the generalized framework we propose in \textsc{Section}~\ref{pi} subsumes ABM.
We encourage the reader to directly jump to that section
for more details about how their (and our) objectives are derived.
Finally, we omit approaches solely relying on ensemble learning (\textit{e.g.} REM \cite{Agarwal2020-eu},
BEAR's
UCB-like ensemble-based extension \cite{Kumar2019-rw} based on Bootstrapped DQN \cite{Osband2016-ff}),
as we want to factor \emph{out} ensembling techniques from the equation
to figure out what are the \emph{core} aspects
of the studied approaches that are single-handedly responsible for the best performance.
Then, one can trivially involve ensembling to reduce the epistemic (parametric) uncertainty.

As mentioned above, we tried to use the hyper-parameter values suggested in the baselines' papers or codebases (provided
the latter is provided, and does not conflict with the companion report), unless the used values are clearly
giving an unfair advantage to one method over the others, while not being part of the claimed reasons
why said method outperforms its competitors. For instance, we aligned the number of layer,
number of hidden units per layer,
and output heads in the neural function approximator of CRR \cite{Wang2020-sr}
with the one used in SAC \cite{Haarnoja2018-bm},
BEAR \cite{Kumar2019-rw},
CQL \cite{Kumar2020-zb},
AWAC \cite{Nair2020-gd} (equivalent approach), among others.
Consequently, we used a 2-layer MLP with 256 hidden units in each layer --- for both the policy (actor) and the critic,
with a single Gaussian head (the policy network returns a single mean $\mu$ and standard deviation $\sigma$ pair).
\cite{Wang2020-sr} uses \emph{mixtures} of Gaussian heads, by contrast.
We make the option available via our re-implementation codebase,
yet do not report any result involving mixtures of Gaussian heads.

We only consider the variant of CRR \cite{Wang2020-sr}
where the function $f$ used to wrap the advantage weighting the likelihood is the exponential function
$x \mapsto \exp(x / \tau)$,
along with the estimation of the advantage where the state-value $V$ is estimated as an empirical average.
The other variants (defining $V$ in the advantage with a maximum operator instead of an expectation,
Heaviside step function for $f$)
were performing poorly across the board, for the suite of tasks and datasets
we selected for this present study.
As such, like AWAC \cite{Nair2020-gd},
the objective used for policy improvement in CRR \cite{Wang2020-sr}
under the considered setting
can be derived exactly in line with the analysis laid out in \textsc{Section}~\ref{pi}.
Plus, by simply replacing the method used to estimate the advantage in AWR \cite{Peng2019-hu}
from a Monte-Carlo one to a temporal-difference one, we end up collapsing onto the objective optimized by
CRR \cite{Wang2020-sr}
and AWAC \cite{Nair2020-gd}.
In line with what was first reported in the \textsc{Appendix} C of AWR \cite{Peng2019-hu},
and in later CRR \cite{Wang2020-sr},
among others,
we clamp the advantage-based exponential weights
--- re-weighting the actor's likelihood in the policy improvement objective
--- to remain below a maximum value of $20$,
for all the tackled methods sharing the same derivation
(AWR \cite{Peng2019-hu},
AWAC \cite{Nair2020-gd},
and CRR \cite{Wang2020-sr},
\textit{cf.}~\textsc{Section}~\ref{pi}).

As for the temperature $\tau$ used in the advantage-based exponential weights objective of
AWR \cite{Peng2019-hu}
and CRR \cite{Wang2020-sr},
we use the values recommended in the respective reports.
As such, we use $\tau = 1$ for CRR, and $\tau = 0.05$ for AWR.
In an effort to uniformize the temperature across the two methods, we investigated how AWR would perform
if the temperature $\tau$ was raised to $1$, and report the associated auxiliary experiment in
\textsc{Appendix}~\ref{awrsweep}, \textsc{Figure}~\ref{awrsweep:barplot}.
Judging by these auxiliary results, which show that neither temperature value
(neither $\tau = 0.05$, nor $\tau = 1$)
clearly outperforms the other,
we do not have reason enough to stray from the original suggested $\tau$ value.
We thus use $\tau = 0.05$ in AWR.

A fair portion of the baselines listed out above involve a \emph{warm-up} period consisting in using
a behavioral cloning loss in the objective optimized by the actor's policy $\pi_\theta$. This is done
either by adding it as an extra piece of the main policy improvement loss, or by replacing the main loss
with the cloning one until a certain iteration is reached.
The thresholds used vary vastly from one baseline to another (values reported in the companion codebase or report).
We use these original values per baseline,
and otherwise do not use any warm-up at all
(\textit{e.g.} CRR \cite{Wang2020-sr} does \emph{not}).

Among the tackled baselines, AWR \cite{Peng2019-hu}
is the only one that approximates the action-value $Q^{\pi_\theta}$ or the actor's policy $\pi_\theta$
with $Q_\omega$ learned via Monte-Carlo (MC) estimation, while \emph{all} the others estimate it via
temporal-difference (TD) learning \cite{Sutton1984-ce, Sutton1988-to, Sutton1999-ii}.
As such, AWR needs the offline dataset $\mathcal{D}$ to be sequentially organized in connex trajectories, in
order to be able to compute the Monte-Carlo returns of each state-action pairs in $\mathcal{D}$,
and use them as targets for $Q_\omega$.
If such information about the $\mathcal{D}$'s sequentiality is \emph{not} available, then AWR is \emph{not} usable.
Yet, if it is, one could then say that AWR has access to privileged information over the other baselines, and
therefore benefits from a putative, unfair advantage.
Either way, this reliance on full trajectories to estimate $Q_\omega$ is a clear drawback.
On the flip side, by not involving bootstrapping over potentially out-of-distribution actions
burdening $Q_\omega$'s stability in the offline regime, AWR is naturally shielded from such instabilities,
or at the very least hindered by them to a lesser extent.
This is a clear advantage for AWR.
Nonetheless, we will see momentarily (\textit{cf.}~\textsc{Section}~\ref{experimentalassessment})
that, despite possessing and leveraging privileged information about the sequentiality
of the offline dataset $\mathcal{D}$,
AWR is consistently outperformed by another baseline in every dataset of the tackled suite.
Since it uses a critic $Q_\omega$ learned via MC-based policy evaluation, and is the only one to do so here,
the methods that outperform it are TD-based.

It is rather surprising that this should happen in the offline RL regime, since the approximation of
the action-value via temporal-difference learning is so much more sensitive to distributional shift (more so than
in the online setting where the shift already takes a toll).
Despite being more prone to $Q_\omega$-related instabilities, it seems that
temporal-difference learning is still
the strongest contender relative to Monte-Carlo estimation in policy evaluation
(\textit{cf.}~\textsc{Section}~\ref{experimentalassessment}).
To prevent $Q_\omega$ misestimation caused by out-of-distribution actions injected in Bellman's equation
(TD learning),
we use the same mechanism that is usually used in the online regime
to counteract the \emph{overestimation} bias \cite{Thrun1993-or}
suffered by $Q_\omega$.
Said mechanism is Double Q-learning \cite{Van_Hasselt2010-qk},
or in particular the extension of Double Q-learning
to modern deep neural models, Double DQN \cite{Van_Hasselt2015-uc}.
Considering our continuous control setting, we use the direct counterpart of Double DQN
for actor-critic architectures presented in \cite{Fujimoto2018-pe}.
As such, by default, every approach evaluated empirically
involves a second Q-value estimate (called \textit{twin} critic
in actor-critics, \textit{cf.}~\cite{Fujimoto2018-pe}).
In the same vein as in \textsc{Section}~\ref{baselines}, we do not study the extent to which more prolific ensembles
(more than two action values) impact the agent's ability to fend off the overestimation bias,
something that has shown limited success so far (\textit{e.g.} in \cite{Kumar2019-rw}).

When estimating $Q_\omega$ via temporal-difference learning, we use target networks,
in line with all the value-based and actor-critic deep RL works that came after
DQN \cite{Mnih2013-rb, Mnih2015-iy}
that originated the stabilization trick.
When porting the various techniques and tricks introduced in DQN to the DPG algorithm \cite{Silver2014-dk},
DDPG \cite{Lillicrap2016-xa}
opted for a slight variation in how the target network trick was executed.
Instead of replacing the frozen parameters of the target networks
\emph{periodically} with a snapshot of the latest iterate
(for both actor and critic, respectively), \cite{Lillicrap2016-xa}
makes the target networks slowly track the main networks by applying \emph{every iteration} an update rule
akin to Polyak's averaging technique \cite{Polyak1992-ci}.

\subsection{Evaluation suite: environments and datasets}
\label{evalsuite}

We carry out all our experiments in the D4RL suite of environment-dataset couples.
We refer the reader to its companion paper \cite{Fu2020-ic}
where the authors report in great detail
\textit{a)} the proficiencies an agent must possess to achieve high performance
in the various physics-based simulated robotics tasks, and
\textit{b)} how the spectrum of datasets associated with each task were collected
--- ranging from data collected from an expert-grade agent (which we sometimes refer to as \textit{high}-quality data)
to purely random data (\textit{low}-quality data).
We focus on the subset of environment-dataset couples from D4RL that are based on the
fast and scalable \textsc{MuJoCo} \cite{Todorov2012-gc} physics engine
and interfaced via OpenAI's \textsc{Gym} \cite{Brockman2016-un} API.
That represents a total of 15 datasets per experiment: 3 distinct environments (corresponding to
tasks involving a distinct \textsc{MuJoCo}-based simulated robot), and 5 distinct datasets for each environment
(of different quality grades, yet the same 5 grades for all 3 environments).
Such consistency enables us to draw more generalizable conclusions from our findings about how different
algorithms perform when provided with dataset of various qualities from the available spectrum.
\textbf{Importantly, such reasoning consistency can only be ensured for the \textsc{MuJoCo}-based tasks of D4RL,
since it is the only benchmark for which the spectrum of dataset qualities ranges
(in 5 increments)
from \textit{low} to \textit{high}
for every single task.}

Throughout this work, we consistently organize the results under this categorization
by arranging the 15 plots in a grid where the \textit{rows} correspond to the distinct five
environments (or equivalently, tasks),
while the \textit{columns} describe the spectrum of the five dataset qualities
--- from \textit{expert} for the left-most column, to \textit{random} for the right-most column.
Intermediate grades include \textit{medium} (collected from a partially-trained agent),
or even \textit{replay} (contents of the replay buffer \cite{Lin1993-qe}
kept from the training procedure of an agent), \textit{cf.} \cite{Fu2020-ic}
for finer details about the spectrum.
As such, in the plots laying out the empirical results,
looking at a \textit{row} gives the respective performances in a given environment in five different datasets
organized left-to-right from expert- to random-grade, while looking at a \textit{column} informs the reader about
how proficient the agent is at accumulating reward and achieving high return
compared to its competitors for a given dataset quality,
across 3 different \textsc{MuJoCo}-based locomotion tasks
(\textit{cf.}~\textsc{Figure}~\ref{baselines:barplot} for an arbitrary example).

\subsection{Evaluation protocol: how do we evaluate our agents?}
\label{evalprotocol}

We run each experiment for 0.5M iterations (the value most often used in the different baselines),
or for a threshold total duration of 12 hours, whichever occurs first.
The stopping condition based on the runtime duration has the obvious advantage of penalizing
the methods that take longer per iteration than their competitors, hence
favoring the ones that are computationally cheaper to run in terms of flops.

In every reported experiment, each agent uses its own modern high-end GPU.
Note, the codebase allows for more than one GPU to be used by an agent, and load-balances
the distributed workers across the available ones automatically.
The distributed paradigm used for D4PG is the following:
\textit{a)} at the start, each worker is assigned a distinct rank $k \geq 0$;
\textit{b)} at the start, each worker sets its random seed as the output of the same deterministic function
that only depends on the rank $k$, making them in effect sample different minibatches
from the offline dataset $\mathcal{D}$;
\textit{c)} at each iteration, each worker with rank $k>0$ computes the gradient of its actor's loss,
then sends it to the worker with rank $k=0$, who aggregates all the $k$ received gradients
(including its own) by computing their empirical average, and finally sends the mean gradients to the
$k-1$ workers with $k>0$ to replace their own gradients with. We decided not to distribute the baselines
(except for D4PG whose distributed aspect is the very core of the method)
primarily to save on computational budget, but also to prevent the gradient averaging scheme to
conceal numerical instabilities some baselines might suffer from more than others.

Additionally, every experiment is repeated over a fixed set of $4$ random seeds, given to the agent beforehand.
Every single plot reported in this work averages the statistics across these random seeds.
Solid lines correspond to the \emph{mean} $\mu$ over the seeds.
Shaded areas correspond to trust regions around $\mu$ whose width are equal to $0.95 \, \sigma$,
where $\sigma$ is the standard deviation of the studies recorded statistic
(\textit{e.g.} the return) over the fixed set of random seeds.

We monitored every experiment with the Weights \& Biases \cite{Biewald2020-wb}
tracking and visualization tool.

We evaluate a given offline RL agent by setting it loose in an \emph{online} instance of the
environment in which the \emph{offline} dataset $\mathcal{D}$ it was trained with was collected.
In other words, our agents are trained in the fictitious MDP $\mathbb{M}^\textsc{off}$,
and evaluated the real MDP $\mathbb{M}$ (\textit{cf.}~\textsc{Section}~\ref{bg}).
Concretely, we evaluate the agent every $5000$ training iterations across all experiments.
Evaluating agents offline, and consequently \textit{a fortiori} off-policy,
is a tedious and challenging feat to carry out properly, as attested by the myriad of works on
off-policy evaluation (OPE) in recent years. We refer the reader to \cite{Voloshin2019-oi}
for a comprehensive overview of the current OPE landscape.
Questions related to opting for off-policy evaluation in offline RL have been raised very recently in
\cite{Le_Paine2020-sb},
and tackled there to an extent.
Overall, how to best evaluate the agent \emph{purely} offline,
\textit{i.e.} without any iteration with $\mathbb{M}$
at any point in the lifetime of the agent, remains an open question.

In terms of metrics used to gauge the quality and proficiency of the learned agent,
we assess to what extent the agent satisfies its performance objective
by observing how high of a return it can accumulate over the course of an episode.
We determine which agent is the best by comparing these average episodic returns ---
the higher, the better. Reaching the top performance \emph{faster} also constitutes a valuable asset for
an agent.

As an agent goes through parameter updates and gets better at interacting with $\mathbb{M}$,
it will survive longer, leading to evaluation trials that also last longer.
These extended survival periods due to the agent's own proficiency at tackling the task at hand have a direct effect
on its total learning process, entangling alternatively training and evaluation phases.
This increase in duration per online evaluation trial will in effect cause the agent (\textit{e.g.} the
BCQ \cite{Fujimoto2018-mj} agent
in the top-left sub-plot of \textsc{Figure}~\ref{baselines:barplot})
to hit the \textit{timeout}
before reaching the 0.5M iterations mark.
Such preliminary termination in terms of iterations thus effectively does not impact how we rank the method,
since there seems to be nothing left for the agent to learn then.
By contrast with prolonged \emph{evaluation} trials,
the performance traces might appear truncated in \textsc{Figure}~\ref{baselines:barplot}
(depicting that the agent has hit the runtime timeout before satisfying the
\textit{``number-of-iterations''} stopping criterion)
due to considerably longer \emph{training} durations (\textit{e.g.} the
CQL \cite{Kumar2020-zb} agent
in the top-left sub-plot of \textsc{Figure}~\ref{baselines:barplot}).
In our experimental setting,
a longer training duration per iteration can only be caused by a higher computational complexity
(allocated computational resources are identical across agents).
While the cause underlying an increase in evaluation time is nonissue since the agent must already be proficient
at the task for such an inflation to even occur,
an extended training duration is more often than not an issue, since it does \emph{not} depend on how well
the agent performs.
By displaying significantly longer training times, an agent might reach the timeout while still performing poorly
and having much left to learn from $\mathcal{D}$.
\textbf{Limiting the allowed time for an agent to solve the task (like we do here purposely) is therefore
\emph{penalizing}
agents whose complexity (and by extension, computational cost) exceed the complexity of its competitors
by too large of a margin.}
Note, we see in \textsc{Figure}~\ref{baselines:barplot} that the used runtime timeout is virtually always long enough
to enable agents to reach the 0.5M iterations mark.
Based on this observation \emph{and} our compute budget, we did not deem necessary to increase said timeout period.
Besides, it seems fair to punish methods that fail to achieve their final performance within the allowed
runtime while so many manage to do so.

\clearpage

\section{Safe Policy Improvement}

\subsection{Conditional operators}
\label{spioperators}
In addition to the operators we have created in \textsc{Section}~\ref{operators},
we also introduce the operators $\mathcal{T}_\textsc{Cond-Eval}^{\omega, \theta, m, \Gamma^\theta_\delta}$
and $\mathcal{T}_\textsc{Cond-Max}^{\omega, \theta, m, \Gamma^\theta_\delta}$,
both from $\mathcal{P}(\mathcal{A})^\mathcal{S}$ to $\mathcal{P}(\mathcal{A})^\mathcal{S}$
like $\mathcal{T}_\textsc{Eval}$ and $\mathcal{T}_\textsc{Max}^{\omega, m}$, which we define
as follows:
\begin{align}
  \big(\forall \pi \in \mathcal{P}(\mathcal{A})^\mathcal{S}\big)
  (\forall s \in \mathcal{S})
  \qquad
  a \sim \mathcal{T}_\textsc{Cond-Eval}^{\omega, \theta, m, \Gamma^\theta_\delta}[\pi](\cdot | s)
  &\iff
  a = \Gamma^\theta_\delta(s) \, \tilde{a}_\textsc{Eval} \: + \: \big(1 - \Gamma^\theta_\delta(s)\big) \,
  \tilde{a}^\theta_\textsc{Max} \label{spievaleq} \\
  &\text{with} \quad
  \tilde{a}_\textsc{Eval} \sim \mathcal{T}_\textsc{Eval}[\pi](\cdot | s)
  \;\, \text{and} \;\,
  \tilde{a}^\theta_\textsc{Max} \sim \mathcal{T}_\textsc{Max}^{\omega, m}[\pi_\theta](\cdot | s)
\end{align}
and
\begin{align}
  \big(\forall \pi \in \mathcal{P}(\mathcal{A})^\mathcal{S}\big)
  (\forall s \in \mathcal{S})
  \qquad
  a \sim \mathcal{T}_\textsc{Cond-Max}^{\omega, \theta, m, \Gamma^\theta_\delta}[\pi](\cdot | s)
  &\iff
  a = \Gamma^\theta_\delta(s) \, \tilde{a}_\textsc{Max} \: + \: \big(1 - \Gamma^\theta_\delta(s)\big) \,
  \tilde{a}^\theta_\textsc{Max} \label{spimaxeq} \\
  &\text{with} \quad
  \tilde{a}_\textsc{Max} \sim \mathcal{T}_\textsc{Max}^{\omega, m}[\pi](\cdot | s)
  \;\, \text{and} \;\,
  \tilde{a}^\theta_\textsc{Max} \sim \mathcal{T}_\textsc{Max}^{\omega, m}[\pi_\theta](\cdot | s)
\end{align}
where $\Gamma^\theta_\delta(s) \coloneqq
\mathds{1}\big[\rho \big(s, \tilde{a}^\theta_\textsc{Max}\big) \geq \delta \big]$,
and where $\rho$ is a potential function taking non-negative real values over the product space
$\mathcal{S} \times \mathcal{A}$.
Since $\Gamma^\theta_\delta$ takes values in the binary set $\{0,1\}$,
$a = \tilde{a}_\textsc{Eval}$ or $\tilde{a}_\textsc{Max}$
if $\rho \big(s, \tilde{a}^\theta_\textsc{Max}\big) \geq \delta$
--- for $\mathcal{T}_\textsc{Cond-Eval}^{\omega, \theta, m, \Gamma^\theta_\delta}$
and $\mathcal{T}_\textsc{Cond-Max}^{\omega, \theta, m, \Gamma^\theta_\delta}$ respectively,
and $a = \tilde{a}^\theta_\textsc{Max}$ if $\rho \big(s, \tilde{a}^\theta_\textsc{Max}\big) < \delta$.
In practice, typical good candidates for $\rho$ are density, novelty, or uncertainty estimates over
$\mathcal{S} \times \mathcal{A}$, which can be obtained, among other techniques,
via random network distillation (RND) \cite{Burda2018-vl},
or by estimating the epistemic uncertainty via an ensemble \cite{Osband2016-ff}.
Note, any signal over $\mathcal{S} \times \mathcal{A}$ that has shown promises when distilled into a reward function
or even inspire the design of one is usually a suitable candidate for $\rho$.
--- \textit{e.g.} signals derived from
psychology and animal learning,
typically categorized under the \textit{intrinsic motivation} \cite{Barto2013-jl, Niekum2010-yk, Schmidhuber2010-bg}
class of incentives to guide the artificial agent's exploration.
We define the
$\mathcal{T}_\textsc{Cond-Eval}^{\omega, \theta, m, \Gamma^\theta_\delta}$ and
$\mathcal{T}_\textsc{Cond-Max}^{\omega, \theta, m, \Gamma^\theta_\delta}$
operators to later introduce proposal policies inspired from \textit{safe policy improvement} (SPI)
\cite{Petrik2016-yc}.
Specifically SPIBB \cite{Laroche2019-ar} relies on an estimate of pseudo-counts
$\tilde{N}_\mathcal{D}(s,a)$ \cite{Bellemare2016-ab, Tang2016-ys, Ostrovski2017-ww},
themselves inspired from the counts involved in the design of upper confidence bounds
in the multi-armed bandit literature
building on the principle of OFUL (\textit{cf.}~\textsc{Section}~\ref{relatedwork}).
The framework we introduce in this work allows us to replicate SPIBB \cite{Laroche2019-ar}
by getting the actions used to bootstrap $Q_\omega$ from a proposal action selection method
built with the operators
$\mathcal{T}_\textsc{Cond-Eval}^{\omega, \theta, m, \Gamma^\theta_\delta}$ and
$\mathcal{T}_\textsc{Cond-Max}^{\omega, \theta, m, \Gamma^\theta_\delta}$,
where $\rho$ would align with the pseudo-count estimator $\tilde{N}_\mathcal{D}$.
In this work, we define $\rho(s,a) > 0$ to be a score aligned with the propensity of the pair
$(s,a) \in \mathcal{S \times \mathcal{A}}$ to be generated by the policy $\beta$ underlying the
offline dataset $\mathcal{D}$.
We achieve this by learning a novelty score over the offline dataset $\mathcal{D}$
(at the same time as the other networks)
via RND \cite{Burda2018-vl}.
This score, which we denote by $\eta(s,a)$, is defined as the prediction error between the outputs of
a neural function approximator frozen after initialization and a non-frozen copy that is updated to predict the
arbitrary frozen outputs of the first network.
While aligned with a novelty signal in terms of variations,
prediction errors (especially from quadratic losses) do not have appropriate scales
to behave well as score surrogates.
As such, we maintain an online rolling estimate of the standard deviation
$\sigma^\eta_\textsc{online}$ of these prediction errors
--- as suggested originally in \cite{Burda2018-vl} ---
and use a normalized novelty score instead, $\bar{\eta}(s,a) \coloneqq \eta(s,a) / \sigma^\eta_\textsc{online}$.
\textit{In fine},
we can now define the potential function $\rho$ that we will use in all the reported empirical results thereafter:
\begin{align}
  (\forall s \in \mathcal{S})(\forall a \in \mathcal{A})
  \qquad
  \rho(s,a) \coloneqq 1 - e^{-\displaystyle\bar{\eta}(s,a)/\tau}
  \label{rhodef}
\end{align}
where a sweep performed in preliminary searches lead us to choose the temperature $\tau=0.06$ in every
subsequent empirical studies reported in this work.
Note, RND's temperature is akin to the bandwidth in kernel density estimation.
The higher the bandwidth (or equivalently, the temperature), the smoother the density estimator.
These sweeps also helped us pick a suitable value for the threshold variable $\delta$,
for which we assign the value $\delta=0.6$.
In terms of range, $\rho$ takes values in $[0,1)$ since $e^{-\displaystyle\bar{\eta}(s,a)/\tau}$
takes values in $(0,1]$.
If the pair $(s,a)$ is deemed novel by $\eta$ (\textit{i.e.} $\bar{\eta}(s,a)$ has high value),
then $\rho(s,a)$ is close to $1$. Conversely, if $(s,a)$ is not considered as novel,
then $\rho(s,a)$ is close to $0$. As we will show shortly, the assembled score $\rho$ can therefore be
instrumental in the design of \textit{safe} action selection methods, which ultimately motivated the
introduction of the operators $\mathcal{T}_\textsc{Cond-Eval}^{\omega, \theta, m, \Gamma^\theta_\delta}$
and $\mathcal{T}_\textsc{Cond-Max}^{\omega, \theta, m, \Gamma^\theta_\delta}$.
These will be used to design \textit{safe} proposal policies, which in the context of offline RL
corresponds to presenting a \textit{low risk} of injecting out-of-distribution action
into function approximators (most critically, into the learned action-value $Q_\omega$
approximator at training time).

\subsection{Adaptively safe proposal distributions}
\label{spiproposalsimplex}
We here introduce proposal policies
inspired from Safe Policy Improvement (SPI) \cite{Petrik2016-yc}
(we give an account of SPI in \textsc{Section}~\ref{relatedwork}).
Instead of focusing only on a single edge among the two top edges of the simplex in \textsc{Figure}~\ref{valuesimplex}
(the edge linking $Q^\beta$ to $Q^*$, and the one linking $Q^\beta$ to $Q^{\pi_\theta}$)
these proposal policies would be located somewhere in between if they were depicted on the simplex of
\textsc{Figure}~\ref{valuesimplex} (not done for legibility reasons).
We define these proposal policies, leveraging the operators
$\mathcal{T}_\textsc{Cond-Eval}^{\omega, \theta, m, \Gamma^\theta_\delta}$
and $\mathcal{T}_\textsc{Cond-Max}^{\omega, \theta, m, \Gamma^\theta_\delta}$
we introduced earlier in
\textsc{eq}~\ref{spievaleq} and \textsc{eq}~\ref{spimaxeq}
respectively, as follows:
\begin{align}
  \zeta \coloneqq \mathcal{T}_\textsc{Cond-Eval}^{\omega', \theta, m, \Gamma^\theta_\delta}
  \big[\beta_\textsc{c}\big]
  \quad \implies \quad
  a' &\sim \mathcal{T}_\textsc{Cond-Eval}^{\omega', \theta, m, \Gamma^\theta_\delta}
  \big[\beta_\textsc{c}\big](\cdot | s')
  \\ \tag*{\textsc{``spi beta clone''}}
  \label{spibetaclone} \\
  \zeta \coloneqq \mathcal{T}_\textsc{Cond-Max}^{\omega', \theta, m, \Gamma^\theta_\delta}
  \big[\beta_\textsc{c}\big]
  \quad \implies \quad
  a' &\sim \mathcal{T}_\textsc{Cond-Max}^{\omega', \theta, m, \Gamma^\theta_\delta}
  \big[\beta_\textsc{c}\big](\cdot | s')
  \\ \tag*{\textsc{``spi beta clone max''}}
  \label{spibetaclonemax} \\
  \zeta \coloneqq \mathcal{T}_\textsc{Cond-Max}^{\omega', \theta, m, \Gamma^\theta_\delta}
  \big[\beta_\textsc{c}^\xi\big]
  \quad \implies \quad
  a' &\sim \mathcal{T}_\textsc{Cond-Max}^{\omega', \theta, m, \Gamma^\theta_\delta}
  \big[\beta_\textsc{c}^\xi\big](\cdot | s')
  \\ \tag*{\textsc{``spi perturbed beta clone max''}}
  \label{spiperturbedbetaclonemax}
\end{align}
Since these are inherently adaptive, data-dependent convex combinations of previously introduced and discussed
operators (\textit{cf.}~\textsc{Section}~\ref{operators})
we can expect the action value $Q_\omega$ learned with $\ell_\omega$ and these hybrid proposal policies
to be in the convex hull of the three corners of the simplex depicted in \textsc{Figure}~\ref{valuesimplex},
$Q^*$, $Q^\beta$, and $Q^{\pi_\theta}$.
Note, the condition $\Gamma^\theta_\delta$ involves the potential function $\rho$ over
$\mathcal{S} \times \mathcal{A}$ defined in \textsc{eq}~\ref{rhodef}, and, perhaps more critically,
depends on $\tilde{a}^\theta_\textsc{Max}
\sim \mathcal{T}_\textsc{Max}^{\omega', m}[\pi_\theta](\cdot | s')$.
Concretely, the \textit{``safe''} proposal policy will act according to
$\mathcal{T}_\textsc{Max}^{\omega', m}[\pi_\theta]$ when $\tilde{a}^\theta_\textsc{Max}$ is close to
being distributed as $\beta$ (\textit{i.e.}~$\pi_\theta$ is close to $\beta$),
but will act according to $\tilde{\zeta} \in \{
\mathcal{T}_\textsc{Eval}[\beta_\textsc{c}],
\mathcal{T}_\textsc{Max}^{\omega',m}[\beta_\textsc{c}],
\mathcal{T}_\textsc{Max}^{\omega',m}[\beta_\textsc{c}^\xi]
\}$ when $\tilde{a}^\theta_\textsc{Max}$ does not seem to have been sampled from $\beta$
(\textit{i.e.}~$\pi_\theta$ is far from $\beta$).
In other words,
$a' \sim \mathcal{T}_\textsc{Max}^{\omega', m}[\pi_\theta]$
if $\rho$ believes $\tilde{a}^\theta_\textsc{Max} \sim \beta$,
and $a' \sim \tilde{\zeta} \in \{
\mathcal{T}_\textsc{Eval}[\beta_\textsc{c}],
\mathcal{T}_\textsc{Max}^{\omega',m}[\beta_\textsc{c}],
\mathcal{T}_\textsc{Max}^{\omega',m}[\beta_\textsc{c}^\xi]
\}$, depending on the chosen strategy, when $\rho$ believes $\tilde{a}^\theta_\textsc{Max} \nsim \beta$.
These three \textit{safe} proposal policies have a direct grasp on whether the actor's policy is about to
predict out-of-distribution actions, and can act on it by instead opting for a \textit{next} action more
likely to be in-distribution, by sampling from an alternate, \textit{safer} proposal
distribution derived from a estimated clone of the offline policy $\beta$.
Making sure $\rho$'s beliefs should be trusted is of independent interest, and its design
comes with its own set of challenges.
In the experiments reported in this work, we stick to the implementation of $\rho$ reported in
\textsc{Section}~\ref{operators}.
The role of $\rho$ could be filled by a myriad of density, novelty, or uncertainty estimators.
Yet, their effectiveness and impact on learning dynamics and final performance
is left out of the scope of this work.

\subsection{SPI policy evaluation}
\label{spiperesults}

We report the performance of \textsc{Base} (\textit{cf.}~\textsc{Algorithm}~\ref{algobase})
for the proposal policies $\zeta$ in the \textit{SPI} group (\textit{cf.}~\textsc{Appendix}~\ref{spiproposalsimplex})
in \textsc{Figure}~\ref{nextappendix:barplot} (we also include the
non-SPI $\zeta$'s reported in \textsc{Figure}~\ref{next:barplot} to make comparisons easier).

The proposal distributions of the \textit{SPI} group
(\ref{spibetaclone}, \ref{spibetaclonemax}, \ref{spiperturbedbetaclonemax};
\textit{cf.}~\textsc{Appendix}~\ref{spiproposalsimplex})
are to a certain extent
\textit{hybrids} between \textit{a)} \ref{thetamax} and \textit{b)}
\ref{betaclone}, \ref{betaclonemax}, and \ref{perturbedbetaclonemax}, respectively.
The quality of these methods depends not only on the clone $\beta_\textsc{c}$ or
the perturbed clone $\beta_\textsc{c}^\xi$ (adding respectively \emph{one} and \emph{two} extra
function approximators to the global neural architecture), but also on the quality of the density
(or novelty, uncertainty, \textit{cf.}~\textsc{Section}~\ref{operators}) estimator $\rho$,
which determines from which policy the next action will be sampled.
While the proposal strategies for $\zeta$ belonging to the \textit{SPI} group perform better than
\ref{thetamax} on the \textit{``expert''} datasets (left-most column of plots
in the grid of \textsc{Figure}~\ref{nextappendix:barplot}) while performing worse than the \textit{clone} methods,
this pattern is not maintained across every dataset.
In fact, a given strategy from the \textit{SPI} group often underperforms both of the strategies it \emph{mixes}
\textit{i.e.}~\ref{thetamax} and either one of the options from the \textit{clone} group
(listed out just above in \textit{b)})
depending on the used variant.

Even if these theoretically-safer strategies
\emph{can} outperform the others in some environment-dataset scenarios
(\textit{e.g.} bottom-right corner in \textsc{Figure}~\ref{nextappendix:barplot})
it appeared not to be worth spending the extra resources to dedicate a large chunk of our budget for them
(including the time it takes to tune the extra moving pieces and knobs,
\textit{e.g.} the density estimator which can be hard to tune per dataset).
In our search, the results were too inconsistent to justify otherwise.

\subsection{SPI policy improvement}
\label{spipiresults}

\begin{figure}[h!]
 \begin{subfigure}{\textwidth}
    \caption{Return of \textsc{Base} (\textit{cf.}~\textsc{Algorithm}~\ref{algobase})
    for every proposal distribution (including the SPI ones)}
    \label{nextappendix:barplot}
    \center\scalebox{0.16}[0.16]{\includegraphics{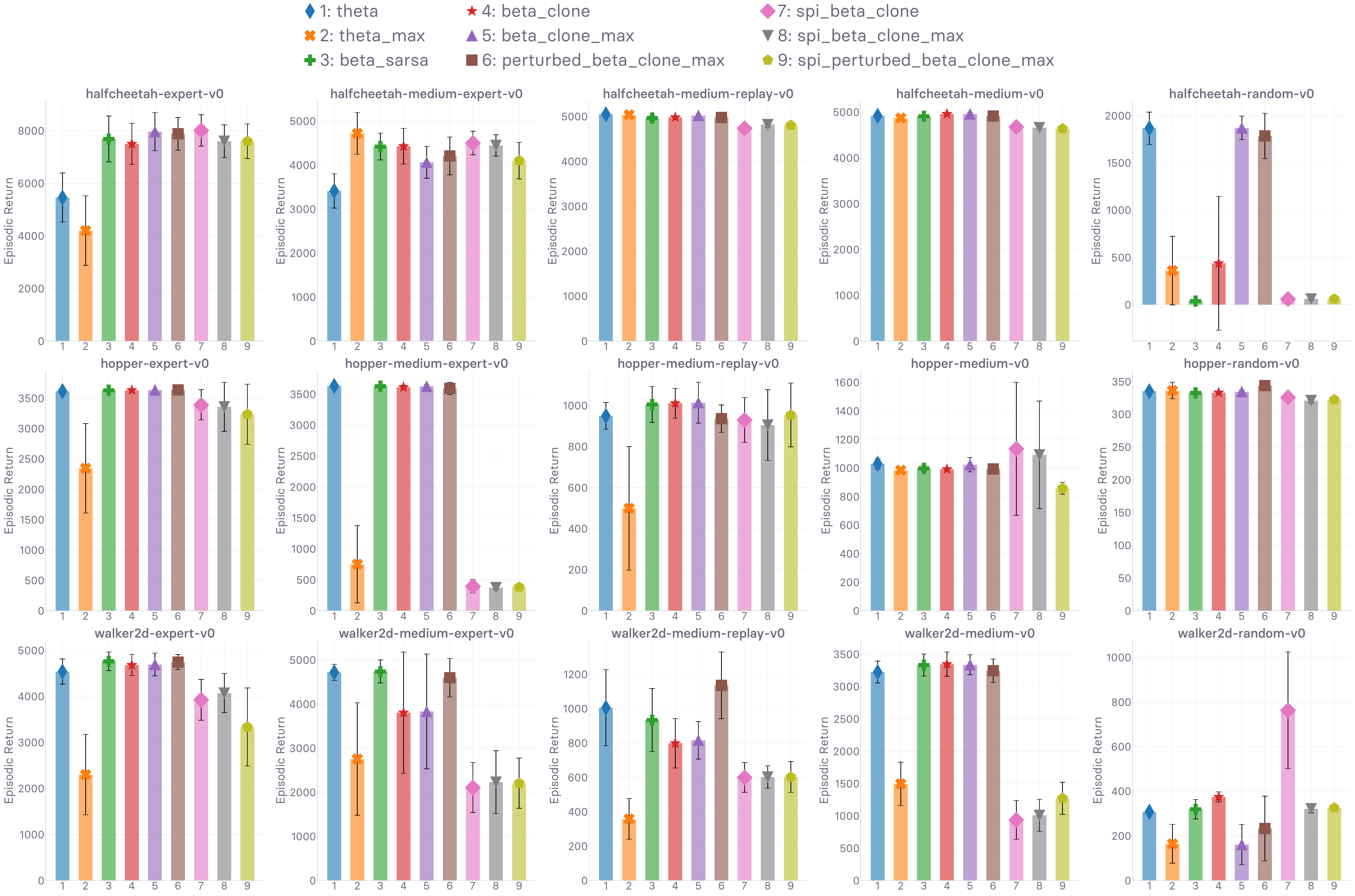}}
  \end{subfigure}
  \begin{subfigure}{\textwidth}
    \caption{Return of GIWR (\textit{cf.}~\textsc{Algorithm}~\ref{algopi})
    for every proposal distribution (including the SPI ones)}
    \label{giwr02appendix:barplot}
    \center\scalebox{0.16}[0.16]{\includegraphics{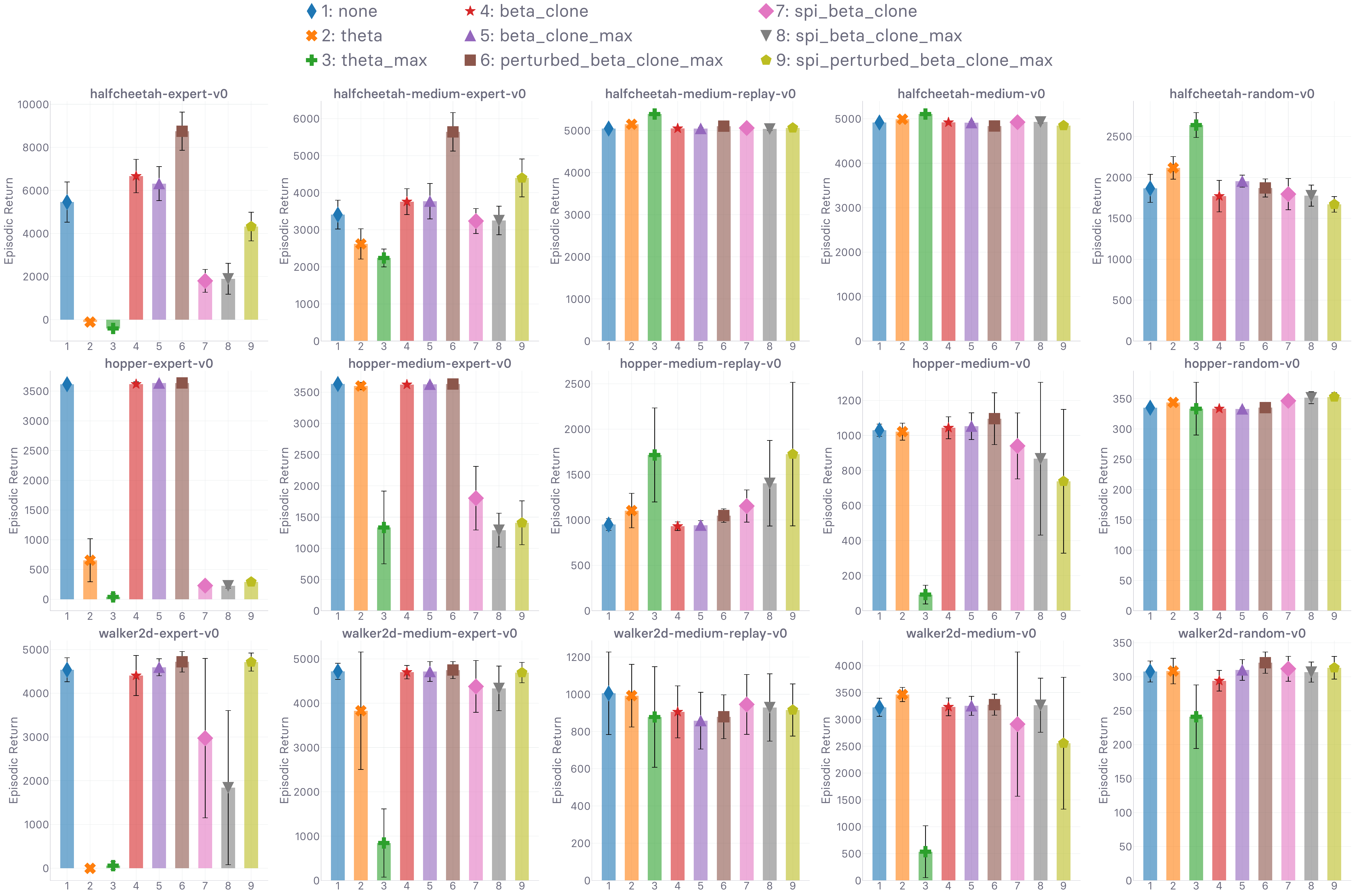}}
  \end{subfigure}
  \caption{(a) Final performance of \textsc{Base} (\textit{cf.}~\textsc{Algorithm}~\ref{algobase})
  with the policy evaluation carried out
  under the different proposal distributions that we introduced in \textsc{Section}~\ref{proposalsimplex}
  and \textsc{Appendix}~\ref{spiproposalsimplex}.
  Everything except the proposal policy $\zeta$ used to sample the \textit{next} action from is identical.
  (b) Final performance of GIWR (\textit{cf.}~\textsc{Algorithm}~\ref{algopi})
  with the policy improvement carried out
  under the different proposal distributions that we introduced in \textsc{Section}~\ref{proposalsimplex}
  and \textsc{Appendix}~\ref{spiproposalsimplex}.
  Everything except the proposal policy $\zeta$ in use is identical.
  We use $\kappa=0.2$ as scaling coefficient for the contribution of $\zeta$
  in \textsc{eq}~\ref{giwrloss}.
  Runtime is 12 hours. Best seen in color.}
\end{figure}

We report the performance of GIWR (\textit{cf.}~\textsc{Algorithm}~\ref{algopi})
for the proposal policies $\zeta$ in the \textit{SPI} group (\textit{cf.}~\textsc{Appendix}~\ref{spiproposalsimplex})
in \textsc{Figure}~\ref{giwr02appendix:barplot} (we also include the
non-SPI $\zeta$'s reported in \textsc{Figure}~\ref{giwr02:barplot} to make comparisons easier).

As we observed earlier in \textsc{Appendix}~\ref{spiperesults},
and adopting the terminology introduced then,
the proposal distributions from the \textit{SPI} group are \emph{often} severely hindered
(yet not \emph{always})
by the weak performance of the $\mathcal{T}_\textsc{Max}^{\omega', m}$ operator in \ref{thetamax}
which they coincide with by design when
$\rho \big(s, \tilde{a}^\theta_\textsc{Max}\big) \geq \delta$
(\textit{cf.}~\textsc{eq}~\ref{spimaxeq} for the definition of the SPI operator template,
and \textsc{eq}~\ref{rhodef} for the definition and surrounding discussion on the design choices related to $rho$).
Despite achieving higher returns overall than their \ref{thetamax} component,
the proposal distributions from the \textit{SPI} group are overall outperformed by their counterparts
proposal policies in the \textit{clone} group, with a gap in performance seemingly stemming from how \emph{mediocre}
\ref{thetamax} is in the considered dataset.
As in \textsc{Figure}~\ref{giwr02appendix:barplot}, this behavior is observed when $\kappa \in \{0.1, 0.5\}$ too,
as exhibited in
\textsc{Figures}~\ref{giwr01:barplot} and \ref{giwr05:barplot},
reported in \textsc{Appendix}~\ref{giwrsweep}.
As such, the methods withing the \textit{SPI} group achieve performance \emph{consistently} ranked in between
\ref{thetamax} and their counterparts in the \textit{clone} group, but only rarely reach the return
accumulated by the best of the two methods between which they are attempting to strike a trade-off.
Since such balance is fully determined by $\Gamma^\theta_\delta(s) \coloneqq
\mathds{1}\big[\rho \big(s, \tilde{a}^\theta_\textsc{Max}\big) \geq \delta \big]$,
one might be able to strike a better trade-off (achieve \emph{``best of both worlds''} results)
by fine-tuning the threshold $\delta$ for the given dataset-environment couple,
and exploring a wider variety of designs for the potential function $\rho$ over $\mathcal{S} \times \mathcal{A}$
(\textit{cf.}~\textsc{eq}~\ref{rhodef}).
Nevertheless, \textsc{Figures}~\ref{giwr02appendix:barplot}, \ref{giwr01:barplot}, and \ref{giwr05:barplot} show that
in most of the considered datasets and environments,
our design choices enable the proposal distribution in the \textit{SPI} group to make good
and \emph{safe} (\textit{cf.}~\textsc{Section}~\ref{relatedwork})
compromises.

In line with these findings, the strategy that displays the highest performance among the \textit{SPI} group
is the one whose counterpart in the \textit{clone} group is \ref{perturbedbetaclonemax}
--- which outperforms every other method
as we have observed in \textsc{Figure}~\ref{giwr02appendix:barplot}.
Such an observation is not surprising but attests to the consistency and robustness of the proposal
heuristics we have put into place.
In the same vein, we also observe that the GIWR framework we here introduce is \emph{not} stiff
(\textit{cf.}~\textsc{Section}~\ref{convenience}) with respect to the choice of $\kappa$, as depicted in
\textsc{Figures}~\ref{giwr01:barplot} and \ref{giwr05:barplot}
from \textsc{Appendix}~\ref{giwrsweep}
where the ranking of methods is essentially
identical to the one observed in \textsc{Figure}~\ref{giwr02appendix:barplot}.
While being robust \textit{w.r.t.} $\kappa$, we see from these plots describing the performed sweep that
increasing the value of $\kappa$ increases the return of the best performing methods further for
expert datasets, while not having neither unexpectedly positive nor unexpectedly negative effect
in the non-expert datasets.
All in all, the GIWR framework is robust in that respect.

\clearpage

\section{Baird's advantage-learning investigation}
\label{bairdal}

\begin{figure}[H]
  \center\scalebox{0.16}[0.16]{\includegraphics{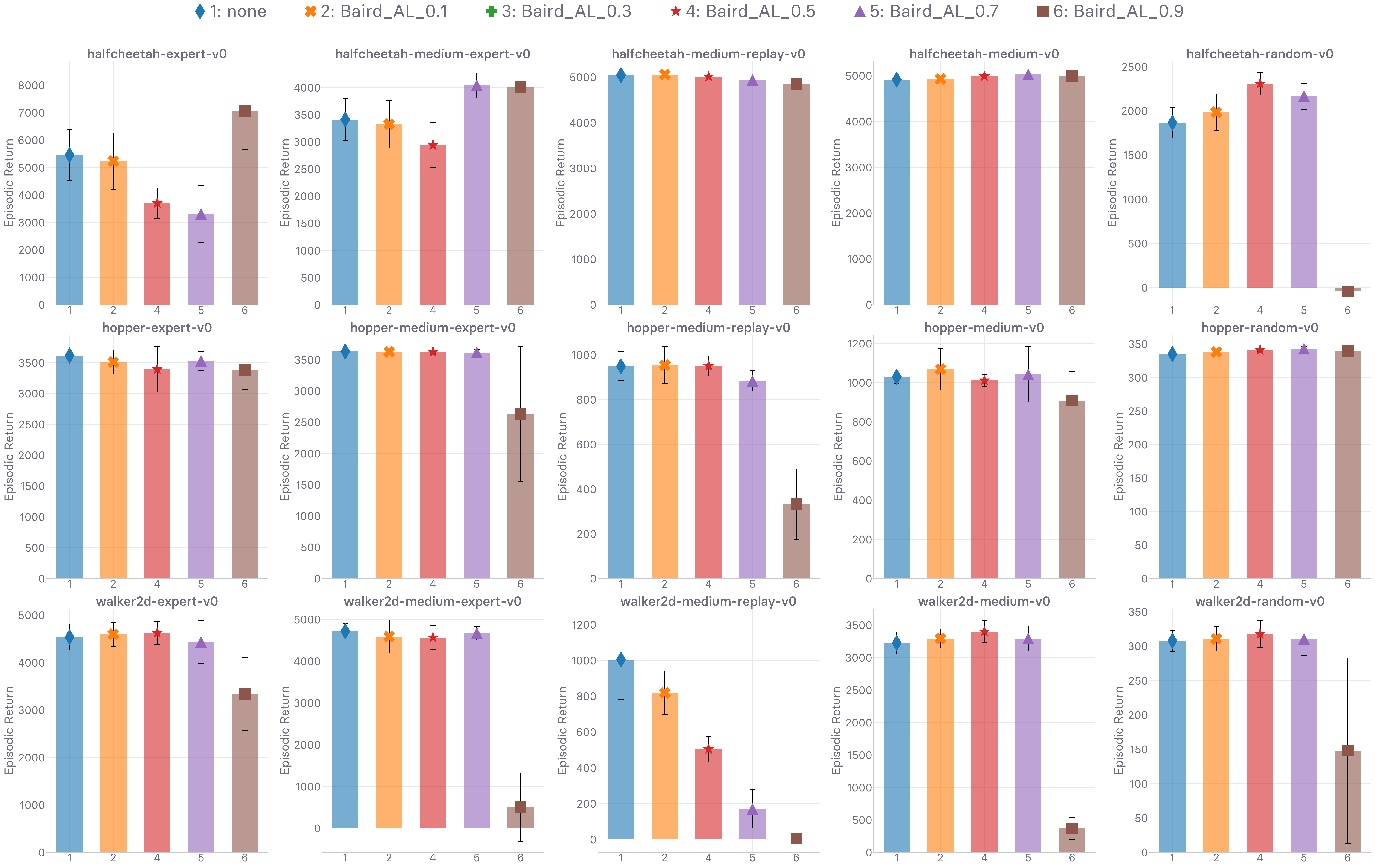}}
  \caption{Empirical evaluation of the use of Baird's advantage-learning bonus (\textit{cf.}~\ref{criticlossal}),
  and sweep over the associated scaling coefficient $\alpha$.
  Runtime is 12 hours. Best seen in color.}
  \label{bairdal:barplot}
\end{figure}

\clearpage

\section{Proposal involving $\mathcal{T}_\textsc{Max}^{\omega, m}$ in policy evaluation}
\label{tmaxops}

\begin{figure}[H]
  \center\scalebox{0.16}[0.16]{\includegraphics{Plots/next_max_m_sweep/%
  beta_clone_max/plots_main_eval_env_ret_barplot.pdf}}
  \caption{Sweep over the number of samples $m$ used in the operator
  $\mathcal{T}_\textsc{Max}^{\omega', m}\big[\beta_\textsc{c}\big]$
  (\textit{cf.}~\textsc{Section}~\ref{operators}, \ref{betaclonemax}).
  Runtime is 12 hours. Best seen in color.}
  \label{betaclonemaxm:barplot}
\end{figure}
\begin{figure}[H]
  \center\scalebox{0.16}[0.16]{\includegraphics{Plots/next_max_m_sweep/%
  perturbed_beta_clone_max/plots_main_eval_env_ret_barplot.pdf}}
  \caption{Sweep over the number of samples $m$ used in the operator
  $\mathcal{T}_\textsc{Max}^{\omega', m}\big[\beta_\textsc{c}^\xi\big]$
  (\textit{cf.}~\textsc{Section}~\ref{operators}, \ref{perturbedbetaclonemax}).
  Runtime is 12 hours. Best seen in color.}
  \label{perturbedbetaclonemaxm:barplot}
\end{figure}
\begin{figure}[H]
  \center\scalebox{0.16}[0.16]{\includegraphics{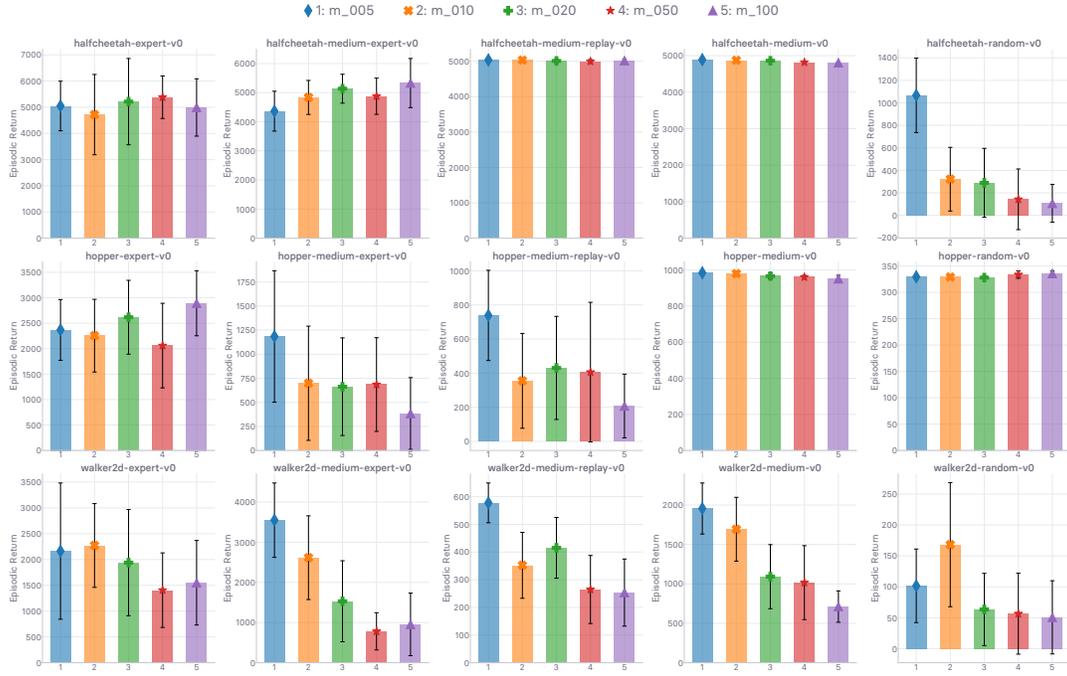}}
  \caption{Sweep over the number of samples $m$ used in the operator
  $\mathcal{T}_\textsc{Max}^{\omega', m}[\pi_\theta]$
  (\textit{cf.}~\textsc{Section}~\ref{operators}, \ref{thetamax}).
  Runtime is 12 hours. Best seen in color.}
  \label{thetamaxm:barplot}
\end{figure}

\clearpage

\section{Policy improvement objective derivation}
\label{agentobjectivedetail}

We begin with the \emph{forward} KL: we unpack the measure and the expectations into integral form,
and inject \textsc{eq}~\ref{pistar}:
\begin{align}
  \mathbb{E}_{s \sim \rho^\beta(\cdot)}
  \Big[
  \Delta\big(\pi_\theta(\cdot | s), \zeta_\textsc{iw}(\cdot | s)\big)
  \Big]
  &\coloneqq
  \mathbb{E}_{s \sim \rho^\beta(\cdot)}
  \Big[
  D^{\zeta_\textsc{iw}}_{\overrightarrow{\textsc{kl}}}[\pi_\theta](s)
  \Big]
  \\
  &= \int_{s \in \mathcal{S}} \rho^\beta(s) \int_{a \in \mathcal{A}} \zeta_\textsc{iw}(a | s)
  \big(\log \zeta_\textsc{iw}(a | s) - \log \pi_\theta(a | s)\big) \, da \, ds
  \\
  &= \int_{s \in \mathcal{S}} \rho^\beta(s) \int_{a \in \mathcal{A}} \zeta_\textsc{iw}(a | s)
  \log \zeta_\textsc{iw}(a | s) \, da \, ds \nonumber \\
  & \qquad -
  \int_{s \in \mathcal{S}} \rho^\beta(s) \int_{a \in \mathcal{A}} \zeta_\textsc{iw}(a | s)
  \log \pi_\theta(a | s) \, da \, ds
\end{align}
\begin{align}
  \implies \quad
  \theta
  &\in
  \argmin_{\theta \in \Theta} \;\,
  \mathbb{E}_{s \sim \rho^\beta(\cdot)}
  \Big[
  \Delta\big(\pi_\theta(\cdot | s), \zeta_\textsc{iw}(\cdot | s)\big)
  \Big]
  \\
  &= \argmin_{\theta \in \Theta} \;\,
  - \int_{s \in \mathcal{S}} \rho^\beta(s) \int_{a \in \mathcal{A}}
  \zeta_\textsc{iw}(a | s)
  \log \pi_\theta(a | s) \, da \, ds \\
  &= \argmin_{\theta \in \Theta} \;\,
  - \int_{s \in \mathcal{S}} \rho^\beta(s) \int_{a \in \mathcal{A}}
  \zeta(a | s) \exp (\frac{1}{\lambda_\textsc{kl}} A^{\pi_\theta}_\omega(s,a))
  \log \pi_\theta(a | s) \, da \, ds \\
  &= \argmax_{\theta \in \Theta} \;\,
  \mathbb{E}_{s \sim \rho^\beta(\cdot), a \sim \zeta(\cdot | s)}
  \bigg[
  \exp (\frac{1}{\lambda_\textsc{kl}} A^{\pi_\theta}_\omega(s,a)) \log \pi_\theta(a | s)
  \bigg]
\end{align}
Conversely, by opting for the \emph{reverse} KL instead, the problem in \textsc{eq}~\ref{pistar}
reduces to the following problem:
\begin{align}
  \mathbb{E}_{s \sim \rho^\beta(\cdot)}
  \Big[
  \Delta\big(\pi_\theta(\cdot | s), \zeta_\textsc{iw}(\cdot | s)\big)
  \Big]
  &\coloneqq
  \mathbb{E}_{s \sim \rho^\beta(\cdot)}
  \Big[
  D^{\zeta_\textsc{iw}}_{\overleftarrow{\textsc{kl}}}[\pi_\theta](s)
  \Big]
  \\
  &= \int_{s \in \mathcal{S}} \rho^\beta(s) \int_{a \in \mathcal{A}} \pi_\theta(a | s)
  \big(\log \zeta_\textsc{iw}(a | s) - \log \pi_\theta(a | s)\big) \, da \, ds
  \\
  &= \int_{s \in \mathcal{S}} \rho^\beta(s) \int_{a \in \mathcal{A}} \pi_\theta(a | s)
  \log \zeta_\textsc{iw}(a | s) \, da \, ds \nonumber \\
  & \qquad -
  \int_{s \in \mathcal{S}} \rho^\beta(s) \int_{a \in \mathcal{A}} \pi_\theta(a | s)
  \log \pi_\theta(a | s) \, da \, ds
\end{align}
\begin{align}
  \implies \quad
  \theta
  &\in
  \argmin_{\theta \in \Theta} \;\,
  \mathbb{E}_{s \sim \rho^\beta(\cdot)}
  \Big[
  \Delta\big(\pi_\theta(\cdot | s), \zeta_\textsc{iw}(\cdot | s)\big)
  \Big]
  \\
  &= \argmin_{\theta \in \Theta} \;\,
  \int_{s \in \mathcal{S}} \rho^\beta(s) \int_{a \in \mathcal{A}}
  \pi_\theta(a | s)
  \log \bigg(\zeta(a | s) \exp (\frac{1}{\lambda_\textsc{kl}} A^{\pi_\theta}_\omega(s,a))\bigg) \, da \, ds
  \nonumber \\
  & \qquad -
  \int_{s \in \mathcal{S}} \rho^\beta(s) \int_{a \in \mathcal{A}} \pi_\theta(a | s)
  \log \pi_\theta(a | s) \, da \, ds
  \\
  &= \argmin_{\theta \in \Theta} \;\,
  \mathbb{E}_{s \sim \rho^\beta(\cdot), a \sim \pi_\theta(\cdot | s)}
  \bigg[
  \log \zeta(a | s) + \frac{1}{\lambda_\textsc{kl}} A^{\pi_\theta}_\omega(s,a)
  \bigg]
  + \mathbb{E}_{s \sim \rho^\beta(\cdot)} \big[H\big(\pi_\theta(\cdot | s)\big)\big]
\end{align}
where $H\big(\pi_\theta(\cdot | s)\big)$ denotes the entropy of $\pi_\theta$ for a given state $s$.

\clearpage

\section{Generalized Importance-Weighted Regression sweep}
\label{giwrsweep}

\begin{figure}[H]
  \begin{subfigure}{\textwidth}
    \caption{Using $\kappa=0.1$
    as scaling coefficient for the contribution of $\zeta$
    in \textsc{eq}~\ref{giwrloss}}
    \label{giwr01:barplot}
    \center\scalebox{0.16}[0.16]{\includegraphics{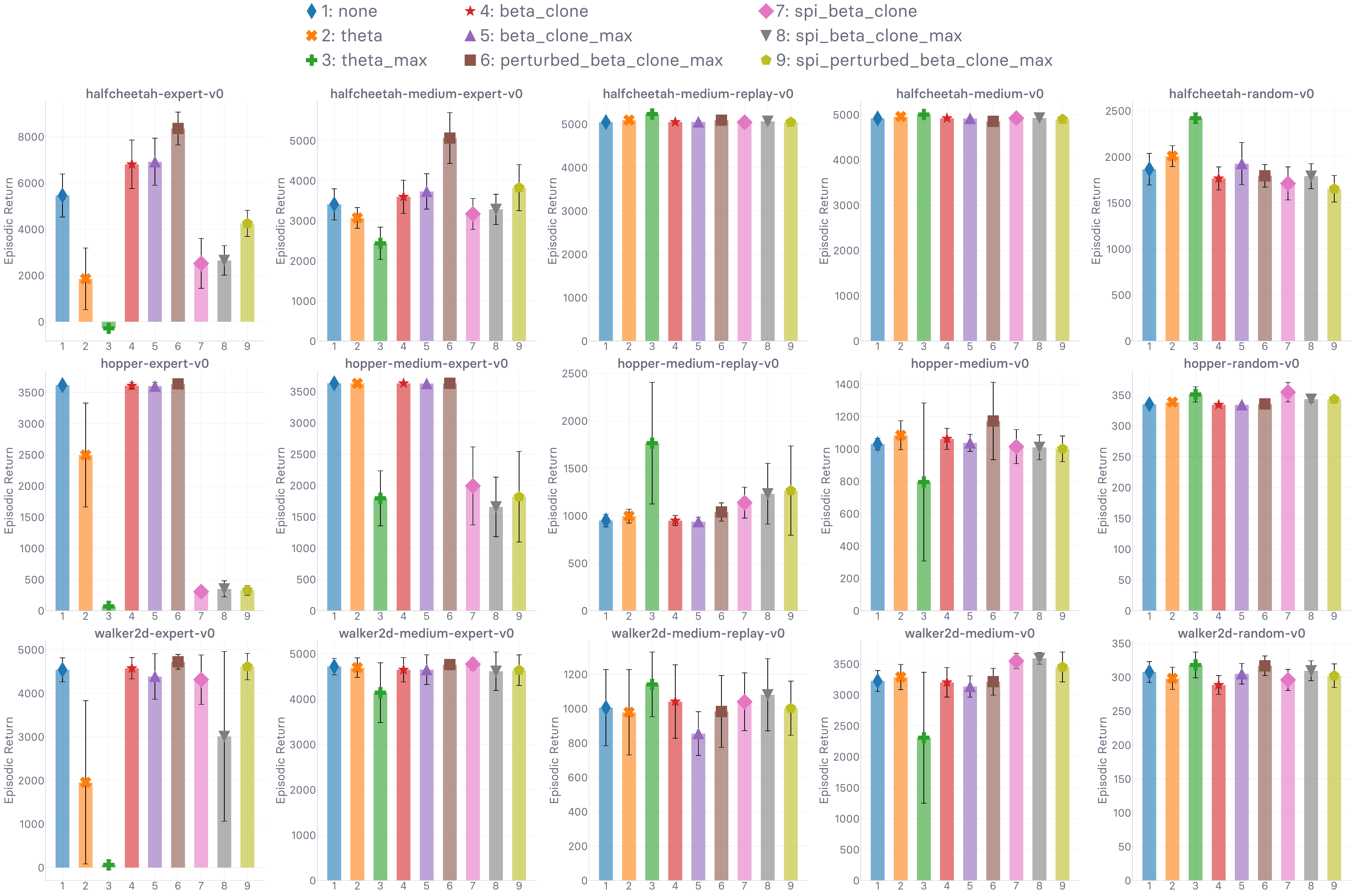}}
  \end{subfigure}
  \begin{subfigure}{\textwidth}
    \caption{Using $\kappa=0.5$
    as scaling coefficient for the contribution of $\zeta$
    in \textsc{eq}~\ref{giwrloss}}
    \label{giwr05:barplot}
    \center\scalebox{0.16}[0.16]{\includegraphics{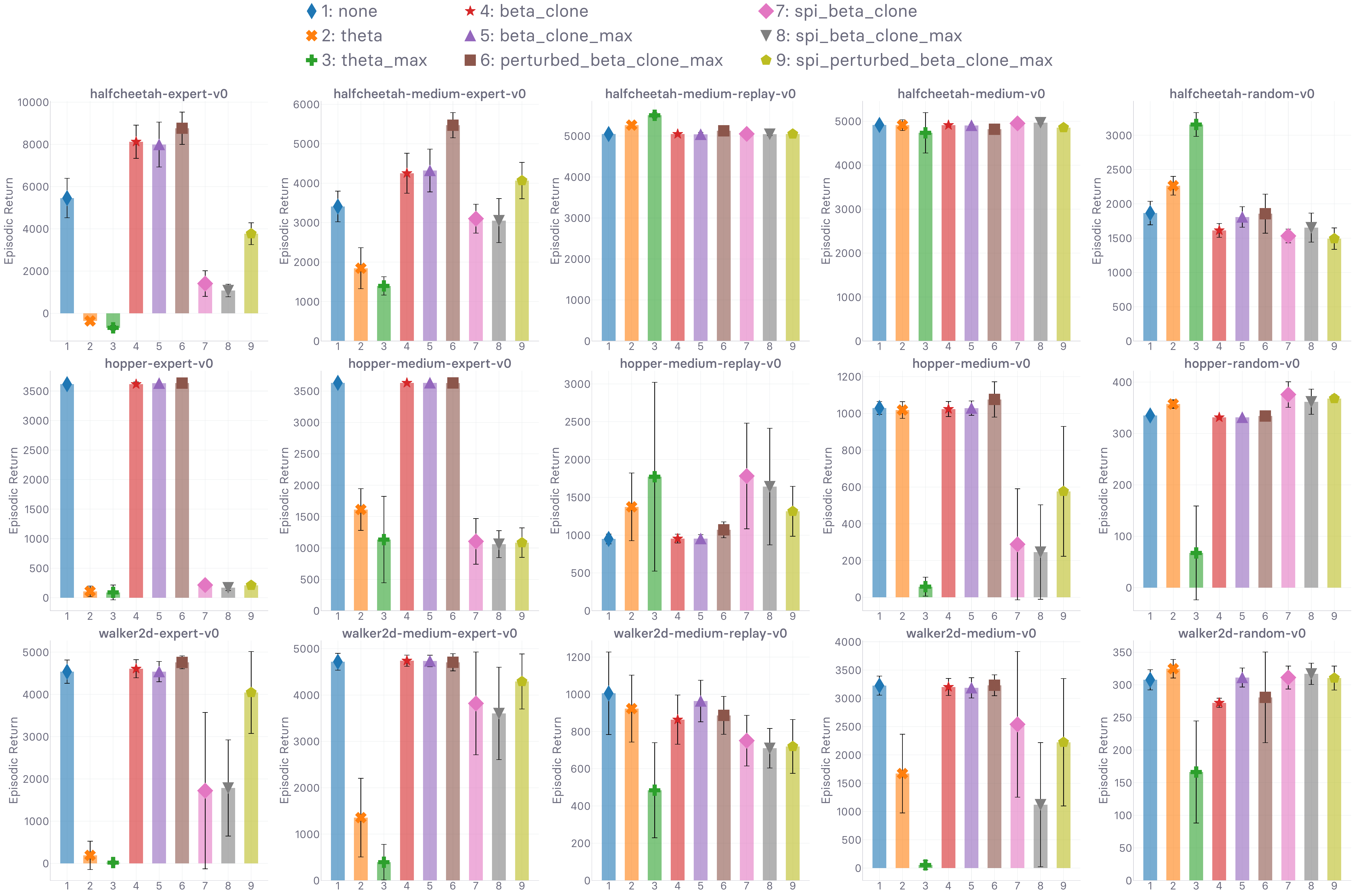}}
  \end{subfigure}
  \caption{Final performance of GIWR (\textit{cf.}~\textsc{Algorithm}~\ref{algopi})
  with the policy improvement carried out
  under the different proposal distributions that we introduced in \textsc{Section}~\ref{proposalsimplex}
  and \textsc{Appendix}~\ref{spiproposalsimplex}.
  Everything except the proposal policy $\zeta$ in use is identical.
  We use (a) $\kappa=0.1$ and (b) $\kappa=0.5$ as scaling coefficient for the contribution of $\zeta$
  in \textsc{eq}~\ref{giwrloss}.
  Runtime is 12 hours. Best seen in color.}
\end{figure}

\clearpage

\section{Temperature sweep in AWR}
\label{awrsweep}

\begin{figure}[H]
  \center\scalebox{0.16}[0.16]{\includegraphics{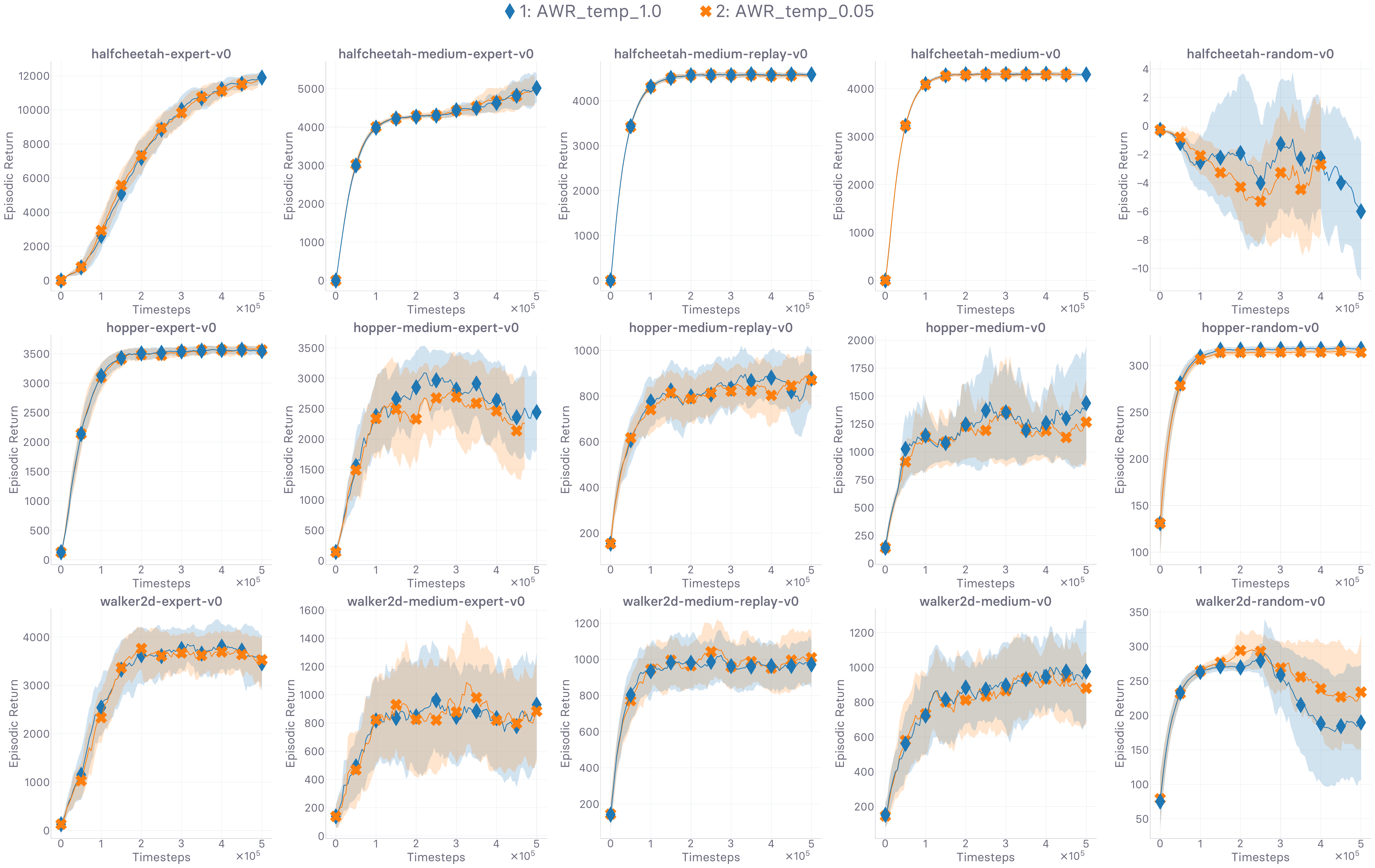}}
  \caption{Sweep over the temperature $\tau$ used in the advantage-based exponential weights objective of AWR.
  Note some sets of runs (\textit{e.g.} top-right sub-plot) terminated early due to an issue on
  our computational infrastructure.
  Since the results were conveying the message we wanted to communicate (the temperature has little to no
  impact on performance), we did not deem it necessary to re-run these experiments.
  Runtime is 12 hours. Best seen in color.}
  \label{awrsweep:barplot}
\end{figure}

\end{document}